\documentclass[12pt]{cmuthesis}
\pdfoutput=1
\usepackage{libertine}
\usepackage[T1]{fontenc}
\usepackage{fullpage}

\usepackage[numbers,sort]{natbib}
\usepackage{array}
\usepackage{multirow}
\usepackage{titlesec}
\usepackage{tabto}
\usepackage{amsmath}
\usepackage{lscape} 
\usepackage{xcolor}
\usepackage{hhline}
\usepackage{multirow, booktabs}
\usepackage{arydshln}
\usepackage{siunitx}
\usepackage{footmisc}
\usepackage{pifont}
\usepackage{comment}
\usepackage{blindtext}

\usepackage{url}
\usepackage[normalem]{ulem}
\usepackage[titletoc,title]{appendix}
\usepackage{graphicx}
\usepackage{wrapfig,lipsum}
\usepackage{makecell}
\usepackage{enumitem}
\usepackage{caption}
\usepackage{amsmath,amsfonts,amssymb,bbm,epsfig,bm,amsthm}
\usepackage{latexsym}
\usepackage{tikz}
\usepackage{color}
\usepackage{listings}
\usepackage{scalerel}
\usepackage{multirow}
\usepackage{pbox}
\usepackage{todonotes}
\usepackage[export]{adjustbox}
\usepackage{enumitem}
\usepackage{bibentry}
\nobibliography*

\usepackage[ruled,lined,algo2e]{algorithm2e}
\usepackage{diagbox}
\usepackage{subfigure}

\usepackage[backref,pageanchor=true,plainpages=false, pdfpagelabels, bookmarks,bookmarksnumbered,
]{hyperref}

\newcommand*{\fullref}[1]{\hyperref[{#1}]{\autoref*{#1} \nameref*{#1}}} 

\usepackage[nameinlink,capitalize]{cleveref}

\usetikzlibrary{tikzmark}
\usetikzlibrary{positioning}

\hypersetup{
    colorlinks=true,
    linkcolor=blue,
    anchorcolor=black,
    linktocpage=false,
    citecolor=blue,
    urlcolor=blue,
}

\usepackage{xcolor, soul}
\sethlcolor{yellow}

\usepackage{kky}
\usepackage{algorithm}
\usepackage[noend]{algpseudocode}

\newcommand{\real}{\mathbb{R}}

\newcommand{\context}{\mathcal{H}}
\DeclareMathAlphabet{\mathsfit}{\encodingdefault}{\sfdefault}{m}{sl}
\SetMathAlphabet{\mathsfit}{bold}{\encodingdefault}{\sfdefault}{bx}{n}

\usepackage[letterpaper,twoside,vscale=.8,hscale=.75,nomarginpar,hmarginratio=1:1]{geometry}

\begin{document} 
\frontmatter

\pagestyle{empty}

\title{ 
{\bf From Recognition to Prediction: \\
 Analysis of Human Action and Trajectory Prediction in Video
}}

\author{Junwei Liang}
\date{CMU-LTI-21-008}
\Year{2021}
\trnumber{}

\committee{
\begin{center}
\begin{tabular}{rl}
Alexander Hauptmann (Chair) & Carnegie Mellon University \\
  Alan W. Black    & Carnegie Mellon University \\
    Kris Kitani   & Carnegie Mellon University \\
    Lu Jiang   & Google Research 
\end{tabular}\end{center}
}
\support{}
\disclaimer{}



\maketitle

\pagestyle{plain} 

\begin{abstract}
With the advancement in computer vision deep learning, systems now are able to analyze an unprecedented amount of rich visual information from videos to enable applications such as autonomous driving, socially-aware robot assistant and public safety monitoring.
Deciphering human behaviors to predict their future paths/trajectories and what they would do from videos is important in these applications. 
However, human trajectory prediction still remains a challenging task, as scene semantics and human intent are difficult to model.
Many systems do not provide high-level semantic attributes to reason about pedestrian future.
This design hinders prediction performance in video data from diverse domains and unseen scenarios.
To enable optimal future human behavioral forecasting, it is crucial for the system to be able to detect and analyze human activities as well as scene semantics, passing informative features to the subsequent prediction module for context understanding.

In this thesis, 
we conduct human action analysis and develop robust algorithms and models for human trajectory prediction in urban traffic scenes.
This thesis consists of three parts. 
The first part analyzes human actions. 
We aim to develop an efficient object detection and tracking system similar to the perception system used in self-driving, and tackle the action recognition problem under weakly-supervised learning settings.
We propose a method to learn viewpoint invariant representations for video action recognition and detection with better generalization.
In the second part, we tackle the problem of trajectory forecasting with scene semantic understanding.
We study multi-modal future trajectory prediction using scene semantics and exploit 3D simulation for robust learning.
Finally, in the third part, we explore using both scene semantics and action analysis for the prediction of human trajectories.
We show our model efficacy on a new challenging long-term trajectory prediction benchmark with multi-view camera data in traffic scenes.
\end{abstract}

 \begin{dedication}
Thank you to my academic adviser, Alex, the best teacher in the world, who has guided me both in research and in life for the past six years.
Thank you to the committee who kept me on track and co-authors along the way that provided me with great help and advice.
Thank you to Lu, who mentored me and led me to everything I know about computer science research.
And finally, thank you to Xinxin, my lovely wife, who traveled through Cambodia for a month to get to the U.S., to accompany and support me during the COVID-19 pandemic, and to be the pillar of my life.
\end{dedication}

{
  \hypersetup{linkcolor=black}
  \tableofcontents
}

\mainmatter








\chapter{Introduction}

\section{Motivation of Research}
With the advancement in deep learning and computer vision, systems now are able to analyze an unprecedented amount of rich visual information from videos to enable many AI applications.
Researchers have achieved great successes in a wide range of computer vision tasks like image classification~\cite{he2016deep}, object detection and instance segmentation~\cite{he2017mask}, object tracking~\cite{sadeghian2017tracking,wu2015object} and scene semantic segmentation~\cite{deeplabv3plus2018}.
Computer vision engineers and scientists are able to put these models into production for applications like image/video content retrieval and in areas like retail and public safety.
One of the most exciting applications is autonomous driving, which may revolutionize how people get around places and how freight moves.
While many prototypes from companies like Waymo and Baidu Apollo have been built over the years, many challenges remain.
One of the main challenges is collision avoidance, which is crucial for self-driving systems co-mingling with humans.
This requires systems to anticipate human motions in the future.
This important analysis is called future person trajectory prediction.
Many existing systems do not provide high-level semantic attribute detection like current human activities and intention recognition to reason about pedestrian future.
In this thesis, our goal is to build a robust trajectory prediction system with in-depth semantic context understanding.
We utilize joint analysis of human actions and enhanced contextual cues from the environment to achieve intent-aware future trajectory prediction.
The future trajectory prediction task has received a lot of attention in the research community~\cite{kitani2012activity,alahi2016social,gupta2018social,liang2019peeking,rhinehart2017first} (see also this comprehensive survey paper~\cite{rudenko2020human}).
It is regarded as an fundamental building block in video understanding because forecasting human behaviors is useful in many applications other than self-driving cars like socially-aware robots~\cite{luber2010people}, advanced surveillance systems, etc.

\begin{figure}[ht]
	\centering
		\includegraphics[width=0.96\textwidth]{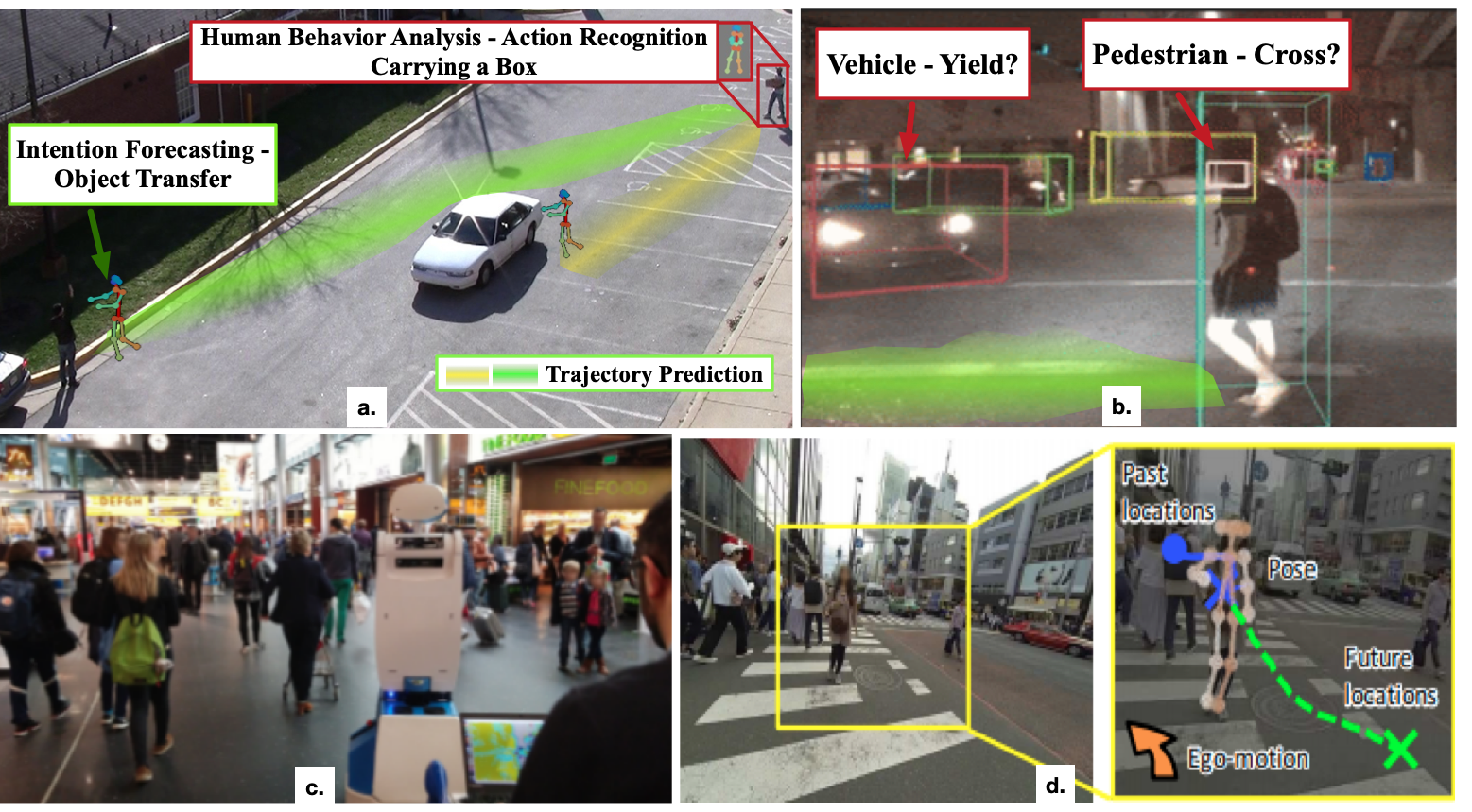} 
	\caption{Our goal is to jointly analyze scene semantics and human actions to predict their future trajectories. (a.) Joint analysis in stationary 45-degree view cameras. The green and yellow line show two possible future trajectories. By recognizing that the person is carrying a box, the system should be able to forecast the intention based on scene constraints and social interactions (the person at the bottom left is waving at the target person), and predict the correct future trajectory. (b.) Future trajectory prediction in self-driving system. (c.) Socially-aware robots from ~\cite{rudenko2020human}. (d.) First-person view trajectory prediction~\cite{yagi2018future}.}
	\label{fig:title}
\end{figure}

Humans navigate through public spaces often with specific purposes in mind, ranging from simple ones like entering a room to more complicated ones like putting things into a car.
Even though their intentions might be determined early, their future trajectories might also be altered by other constraints like social interactions and scene constraints (i.e., external stimuli, as noted in ~\cite{rudenko2020human}). 
This poses several challenges for the pedestrian future trajectory and action prediction task as follows: 

\textbf{(1) The scene constraints are complex and they are changing dynamically.} 
In urban environment, the scene can be diverse and unpredictable with multiple actors including vehicles and pedestrians performing different tasks.
For self-driving systems, as the cars are moving around, the prediction models have to adapt to the dynamic changes of the scene (as well as taking into account the ego-motions), which makes the forecasting problem even more complex.
We need prediction models that are robust to scene and viewpoint variants.

\textbf{(2) The future is often uncertain.} 
Given the same historical trajectory, a person may take different paths, depending on their  (latent) goals. 
Consider the example in Fig.~\ref{fig:title} a., the person (at the top-right corner) might take different paths depending on their intention, e.g., they might take the green path to~\emph{transfer object} or the yellow path to~\emph{load object into the car}. 
We have also demonstrated this nature in our proposed Forking Paths dataset in ~\autoref{chap:0201_multi}.
Thus recent work has started focusing on  \emph{multi-future trajectory prediction}~\cite{tang2019multiple,chai2019multipath,li2019way,makansi2019overcoming,thiede2019analyzing,lee2017desire}.

\textbf{(3) Training data is limited for rare scenarios.} 
For self-driving applications, safe operation is a priority. Data from incidents can help us better improve the system but they are often very rare and scene-dependent. 
Some traffic events are impossible or too dangerous to be acted out by actors for data collection purposes.
As large amounts of well-annotated video data for such rare scenarios are hard to get, in this thesis we investigate the usage of 3D simulator and weakly-supervised learning to alleviate this challenge.
We also propose a new multi-view trajectory prediction dataset, with rich activity annotations that could enable enhanced contextual cue models.

With the advancement of Convolutional Neural Networks (CNNs), Recurrent Neural Networks (RNNs) and graphical models ~\cite{velivckovic2017graph}, we will investigate how to design efficient and robust models suitable for joint scene semantic and action analysis for future trajectory predictions.
To tackle the aforementioned challenges, in \autoref{part:action_analysis}, we first explore and develop action analysis perception system that is efficient to capture behavioral cues and semantic attributes of pedestrians.
We also look into weakly-supervised learning by utilizing vastly available Internet data in \autoref{chap:0102_weakly} to get long-tail action training data.
We aim to build robust models that could produce viewpoint invariant representations in \autoref{chap:0103_viewpoint}.
In \autoref{part:future_prediction}, we propose multi-modal trajectory prediction model and study domain robustness.
For the second challenge, we develop multi-future trajectory prediction model and a new benchmark in \autoref{chap:0201_multi}.
For the third challenge, we utilize recent success of learning from 3D simulation~\cite{ruiz2018learning} and adversarial learning~\cite{tramer2017ensemble} to train a robust model in \autoref{chap:0202_3d}. 
In \autoref{part:joint}, we develop a prediction system that could utilize enhanced contextual cues in the scene and study the more challenging long-term trajectory prediction problem in traffic scenes.

\section{Applications}
Forecasting human behaviors in complex urban environment is useful in many applications such as autonomous driving, public/traffic safety monitoring, socially-aware robots, etc.

\subsection{Autonomous Driving}
Safe operation is a crucial prerequisite for self-driving cars to become ubiquitous in the future.
Forecasting human behavior including pedestrians and cyclists is an essential building block to achieve that~\cite{liang2019peeking,karasev2016intent}.
In Fig.~\ref{fig:title} b., it shows an example camera view from the self-driving cars. The human action analysis and prediction system is key to recognizing the pedestrian and whether they are going to continue to walk across the road, making sure that the driving planning system makes the right action for maximum safety.

\subsection{Socially-aware Robots}
Socially-aware smart robot assistants (Fig.~\ref{fig:title} c.) can be utilized in the future to help humans in performing everyday tasks, such as grocery shopping and delivery.
They are required to operate in a safe and socially-acceptable manner, since they would move closely among other people.
The ability to analyze and predict human behaviors from first-person camera view (Fig.~\ref{fig:title} d.) is essential for motion planning of the robots.

\subsection{Advanced Public/Traffic Safety Monitoring}
Human action analysis and prediction system can also be applied to stationary cameras for safety monitoring.
With trajectory prediction enabled, traffic safety system can issue warnings to pedestrians and vehicles before potential accidents happen.
Meanwhile, by looking into the predicted collided trajectories that do not actually happen, traffic safety researchers can study these cases as near-misses to identify risky intersections or roadways.
Furthermore, trajectory prediction can enable analysis and control of crowd flow in populated areas such as malls and airports by combining other crowd dynamics analysis tools like crowd counting~\cite{cheng2019learning} (see this news story~\footnote{\url{https://www.washingtonpost.com/investigations/interactive/2021/dc-police-records-capitol-riot/}} from the Washington Post how our crowd dynamics analysis has been used in real-world scenario).
In terms of public safety monitoring, the forecasting system can also be applied to predicting the path a suspect may flee after an robbery or other crime incidents, by using long-term prediction model that takes into account scene constraints and external knowledge.
This would be an important addition to the set of existing public safety tools like gunshot detection~\cite{liang2017temporal}, shooter localization~\cite{liang2019technical}, social media monitoring~\cite{huang2018multimodal} and event reconstruction~\cite{chen2016videos,liang2017event}.

\section{Thesis Organization}
This thesis aims to predict pedestrian future trajectories by jointly analyzing human behaviors and contextual cues from the environment. 
We first investigate human action analysis, including efficient object detection and viewpoint-invariant human action recognition \& detection in videos.
This provides important semantic attribute inference and robust representations that could help aid human intention recognition.
We then focus on developing models for future trajectory prediction with semantic scene understanding.
Finally, we utilize our findings from the first two parts and develop a robust model for challenging long-term trajectory prediction, on a newly annotated multi-view benchmark. 
A detailed overview of each part is as follows:

\noindent \paragraph{\fullref{part:action_analysis}}
In this part, we study the problem of human action analysis, including efficient object detection and tracking, and action analysis.
For object detection and tracking, our goal is to build an efficient perception system that utilizes robust image object detection model and fast traditional tracking method for extended videos (\autoref{chap:0101_object}).
Our focus in this part is on action recognition and detection. 
As human actions are diverse, it is impossible to collect training data for all action classes.
Therefore we first study weakly-supervised learning with massive video data with weak labels from Internet platforms like YouTube (\autoref{chap:0102_weakly}).
To enable better generalization of video action recognition models to all kinds of camera views, we then investigate viewpoint-invariant representation learning (\autoref{chap:0103_viewpoint}).

\noindent \paragraph{\fullref{part:future_prediction}} 
In this part,
we study how trajectory prediction can benefit from semantic context understanding of the scene.
Since the future is uncertain, we first introduce the \textit{Multiverse} model to tackle the multiple-future trajectory prediction problem (\autoref{chap:0201_multi}).
To alleviate the limited training data  challenge as mentioned in previous section, we propose a machine learning algorithm called \textit{SimAug}, to efficiently learn from 3D simulation data for robust trajectory prediction (\autoref{chap:0202_3d}).

\noindent \paragraph{\fullref{part:joint}} 
In the final part, we aim to build robust prediction model with enhanced contextual cues from scene semantics and human action representations through multi-task learning.
We first study jointly predicting short-term pedestrian trajectories and activities on common benchmarks (\autoref{chap:0301_joint}).
Since short-term future prediction is not enough to ensure safe operations in autonomous driving applications, we introduce a new, human-annotated long-term trajectory and action prediction benchmark with multi-view camera data (\autoref{chap:0302_longterm_data}) in urban traffic scenes.
Finally, we utilize our findings from the first two parts and develop a robust model for the aforementioned challenging long-term trajectory prediction task (\autoref{chap:0303_longterm_model}).

\section{Contributions}
In this section, we briefly discuss our contributions in each task considered in this thesis.

\subsection{\autoref{part:action_analysis}: Human Action Analysis}
As noted before, our goal is to build a robust trajectory prediction system with enhanced semantic context understanding.
In order to achieve that, in \autoref{part:action_analysis}, we first focus on machine perception, as it is important for prediction models to get enhanced contextual information from the scene and the target agent~\cite{rudenko2020human}.
We investigate efficient object detection and tracking models in \autoref{chap:0101_object}. 
It is not our intention to achieve state-of-the-art on common benchmarks like MSCOCO~\cite{lin2014microsoft}, but to create an efficient framework for video perception that is easy to swap in new models in the future.
In \autoref{chap:0102_weakly}, we are one of the early works that study how we could better utilize weakly-supervised video data from the Internet using self-paced curriculum learning~\cite{jiang2015self}.
In \autoref{chap:0103_viewpoint}, we explore viewpoint-invariant representation for action recognition and detection, which is one of the early works to tackle this problem in video.

\subsection{\autoref{part:future_prediction}: Pedestrian Trajectory Prediction with Scene Semantics}
In \autoref{part:future_prediction}, we study trajectory prediction models with scene semantic understanding.
Following the taxonomy proposed in ~\cite{rudenko2020human}, in \autoref{chap:0201_multi}, we study a pattern-based sequential model that considers scene semantics as contextual cues for multi-future trajectory prediction. In this work, we propose the first human-annotated benchmark for multi-future trajectory prediction.
We then explore adversarial learning in ~\autoref{chap:0202_3d} to build robust trajectory prediction models for different environment and camera views, which is an important under-explored problem in this research field~\cite{rudenko2020human}.

\subsection{\autoref{part:joint}: Joint Analysis of Human Actions and Trajectory Prediction}
In \autoref{part:joint}, we study trajectory prediction models with scene semantic understanding and action analysis.
In \autoref{chap:0301_joint}, we develop the first joint action and trajectory prediction model.
In this work, we are also among the very early works that look into how \textbf{noisy input} of imperfect object detection and tracking would affect trajectory prediction performance. 
Such robustness experiments have been noted in \cite{rudenko2020human} as very important to understand and measure the effectiveness of the prediction models.
In \autoref{chap:0302_longterm_data}, we propose a new human-annotated long-term trajectory prediction benchmark with multi-view video data in urban traffic scenes.
In \autoref{chap:0303_longterm_model}, we propose a new model that utilizes both scene understanding and viewpoint-invariant action representation (\autoref{chap:0103_viewpoint}) for the challenging long-term trajectory prediction problem in urban traffic scenes.

\section{Overall Impact}
\label{sec:overall_impact}
This thesis has potential impact on crowd dynamics analysis in general, in particular, on several directions of the pedestrian trajectory prediction research, including multi-modal trajectory prediction (\autoref{chap:0201_multi}), learning from 3D simulation (\autoref{chap:0202_3d}), learning from enhanced contextual cues (\autoref{chap:0301_joint}, \autoref{chap:0302_longterm_data}, \autoref{chap:0303_longterm_model}), to name a few.
This thesis also has potential impact on the computer vision perception direction, specifically on efficient object detection and tracking (\autoref{chap:0101_object}), weakly-supvervised learning (\autoref{chap:0102_weakly}), and action recognition (\autoref{chap:0103_viewpoint}).
Below, we highlight the impact of our work in academia and industry.

\noindent\textbf{Impact in Academia}
\begin{itemize}
\item Our work has been published at top conferences and journals including CVPR, TPAMI, ECCV, AAAI, IJCAI, etc. As of June 2021, our work has received a total of more than 200 citations from all over the world.
\item Our work on using enhanced contextual cues for trajectory prediction (CVPR'19, \autoref{chap:0301_joint}) has received 140+ citations and it is one of the top-cited paper at CVPR 2019 in this research field. Notably, many researchers have extended our work in terms of multi-task learning for trajectory prediction~\cite{rasouli2020pedestrian,bi2019joint}, action prediction~\cite{chen2019scr,kotseruba2021benchmark} and egocentric-view trajectory prediction~\cite{quan2021holistic,poibrenski2020m2p3,cai2021pedestrian}. 
Our work has also inspired ~\cite{wong2020bgm,wang2021graphtcn} to propose more efficient models that are suitable for deployment and \cite{chen2019scr,sun2020recursive,xue2020location} have developed a new improved version of our work with graph models for single-future trajectory prediction.
\item Our recent work on multimodal prediction and simulation dataset have also started to gain momentum in the research community. Our Multiverse model (CVPR'20, \autoref{chap:0201_multi}) has inspired follow-up work to extend the use of grid occupancy maps\cite{rasouli2020pedestrian,sadat2020perceive,pang2021trajectory,gilles2021home,katuwandeniya2021multi,casas2021mp3} and spatial-temporal graph attention~\cite{wang2021graphtcn,bae2021disentangled,bertugli2020ac} for multimodal trajectory prediction. Our dataset using 3D simulation (\autoref{chap:0201_multi} and \autoref{chap:0202_3d}) has enabled ~\cite{makansi2021exposing,amirian2020opentraj,priisalu2020semantic} to work on this new problem setting for trajectory prediction and \cite{qi2020learning,ma2020diverse} have used our dataset for multimodal evaluation.
\item Most of our research work has been open-sourced and our Github repositories have a total of 800+ stars and 300+ forks as of June 2021. Our \autoref{chap:0301_joint} work is \#1 Tensorflow-based code on PaperWithCode~\footnote{\url{https://paperswithcode.com/task/trajectory-prediction}} in trajectory prediction task.
\end{itemize}

\noindent\textbf{Impact in Industry}

\begin{itemize}
\item Our weakly-supervised work \autoref{chap:0102_weakly} has won the TRECVID Ad-hoc Video Search Challenge~\footnote{\url{https://trecvid.nist.gov/}}, and our efficient object detection and tracking work \autoref{chap:0101_object} has won the TRECVID Activities in Extended Videos Challenge.
Our pedestrian trajectory prediction work, which is related to traffic safety and public safety, has won the NIST ASAPS challenge~\footnote{\url{https://www.herox.com/ASAPS1/update/3483}} with a \$30,000 prize.
\item Our trajectory prediction work has been implemented for a COVID-19 related project called Social Distancing Early Forecasting System~\footnote{\url{https://github.com/JunweiLiang/social-distancing-prediction}} and we have received \$6200 research grant for this from Google Cloud.
\item Our efficient object detection and tracking code (\autoref{chap:0101_object}) has been used by many companies, including Paravision, iMotion Germany, Neosperience, etc.
\end{itemize}

\noindent\textbf{Media Coverage}
\begin{itemize}
\item Our work on trajectory prediction has been reported by two major Chinese Tech outlet and it received over 30,000 views in a week. Feb 14, 2019.
\item Our blog on pedestrian trajectory prediction has been mentioned and referenced by many bloggers~\footnote{\url{https://medium.com/@junweil/social-distancing-early-forecasting-system-60186baa67f5}}.
\end{itemize}

\section{Terminology}
In this paper, we use the term ``trajectory'' and ``path'' interchangeably to refer to the ground-level 2D positions of an agent (humans, vehicles, etc.) over a period of time. 
We refer to short-term and long-term prediction to characterize prediction horizons of 3-5 seconds and 12 seconds ahead, respectively.
The ``long-term'' definition is consistent with recent publications in the field~\cite{karasev2016intent,tran2021goal,rudenko2020human}.
We also use the term ``actions'' and ``activities'' interchangeably to refer to a predefined set of human actions of various duration from public datasets (VIRAT~\cite{oh2011large}, Kinetics~\cite{kay2017kinetics}, etc.).
We use ``intent'', ``intention'' and ``goal'' interchangeably to refer to a person's latent, near-future (same time horizon as the prediction) objective or purpose that associates with a physical destination and an action.

\part{Human Action Analysis}
\label{part:action_analysis}
In this part, we study the problem of human action analysis, including efficient object detection and tracking, and action recognition.
For object detection and tracking, our goal is to build an efficient perception system that utilizes robust image object detection model and fast traditional tracking method for extended videos (\autoref{chap:0101_object}).
Our focus in this part is on action recognition and detection. 
As human actions are diverse, it is impossible to collect training data for all action classes.
Therefore we first study weakly-supervised learning with massive video data with weak labels from Internet platforms like YouTube (\autoref{chap:0102_weakly}).
To enable better generalization of video action recognition models to all kinds of camera views, we then investigate viewpoint-invariant representation learning (\autoref{chap:0103_viewpoint}).

\chapter{Efficient Object Detection and Tracking in Extended Videos 
}  \label{chap:0101_object}

{\let\thefootnote\relax\footnote{Code and models are released at \url{https://github.com/JunweiLiang/Object_Detection_Tracking}}}

\noindent In this chapter, we study the problem of building an efficient perception system for extended videos, either from surveillance cameras or onboard cameras from self-driving cars.
The perception system, which includes object detection and tracking models, provides the basic inputs, i.e., person and object tracklets, to action recognition and future prediction models in later chapters. 
We study the object detection and tracking models for videos in the Activities in Extended Videos Prize Challenge (ActEV)~\cite{awad2020trecvid,liu2020argus}.
With the availability of large-scale video surveillance dataset such as VIRAT~\cite{oh2011large}, ActEV seeks to encourage the development of real-time robust automatic activity detection algorithms in surveillance scenarios. 
We aim to develop a robust and efficient object detection and tracking model trained on the MSCOCO dataset~\cite{lin2014microsoft} as the object annotations in the ActEV dataset are quite small compared to MSCOCO.
Many image object detection research works~\cite{he2017mask,dai2016r} have been done on the MSCOCO dataset~\cite{lin2014microsoft}, which includes 80 object classes.
Compared to other object detection datasets including PASCAL VOC dataset~\cite{everingham2010pascal}, MSCOCO has better annotations, with many more bounding boxes and classes (The YouTube-BoundBox dataset~\cite{real2017youtube} has more boxes than MSCOCO dataset but it is not exhaustively annotated).
We also study Tensorflow~\cite{tensorflow2015-whitepaper} and TensorRT to build a perception system that reaches faster-than-real-time relative processing speed with certain hardware requirements.
It is not our intention to develop State-of-the-Art (SOTA) object detection model but instead to construct a framework for extended videos that we could easily swap the model with the latest COCO-trained models (we have implemented Mask-RCNN~\cite{he2017mask} from 2017 and EfficientDet~\cite{tan2020efficientdet} from 2020.)

\section{Motivation}

\begin{figure}[ht]
	\centering
		\includegraphics[width=0.96\textwidth]{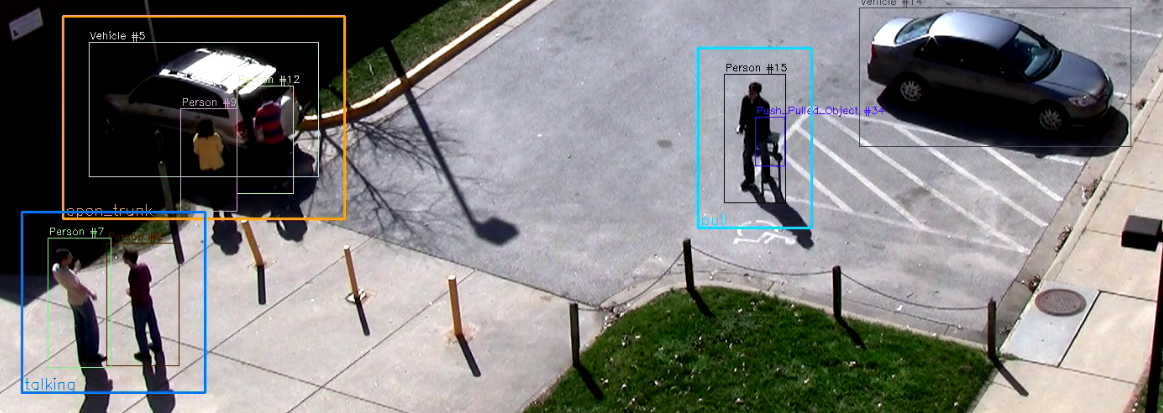} 
	\caption{Example of the ActEV dataset. The videos are annotated with pedestrian and vehicle tracklets with bounding boxes as well as the pedestrian's activities. For example, the two persons at the bottom-left are labeled with ``talking''.}
	\label{fig:0101_actev}
\end{figure}

In recent years, the volume of video data from widely deployed surveillance cameras has grown dramatically. However, camera network operators are overwhelmed with the data to be monitored, and usually cannot afford to view or analyze even a small fraction of their collections. 
To enable timely response for public safety and traffic safety events, there is thus strong incentive to develop fully-automated methods to identify and localize activities in extended video collections and provide the capability to alert and triage emergent videos. 
These methods will alleviate the current manual process of monitoring by human operators and scale up with the growth of sensor proliferation in the near future.
With the availability of largescale video surveillance dataset such as VIRAT~\cite{oh2011large}, the Activities in Extended Videos Prize Challenge (ActEV)~\cite{awad2020trecvid} seeks to encourage the development of real-time robust automatic activity detection algorithms in surveillance scenarios. 
Specifically, an activity is defined to be “one or more people (or vehicle) performing a specified movement or interacting with an object or group of objects”. 
Fig.~\ref{fig:0101_actev} illustrates the ``talking'', ``open\_trunk'' and ``pull'' activities. 
A few recent work applied a two-stage architecture for temporal action localization~\cite{xu2017r,dai2017temporal}, and demonstrated competitive performance. 
In particular, R-C3D network~\cite{xu2017r} closely follows the original Faster R-CNN~\cite{ren2015faster} architecture but in the temporal domain. 
There are also several previous works~\cite{gleason2019proposal,hou2017tube} focusing on online action detection or fine-grained action detection untrimmed Internet videos. 
However,these methods may not generalize to a more challenging spatial-temporal activity detection problem, which is central for extended video analysis in surveillance videos and self-driving cameras.
In this chapter, we describe our object detection and tracking models to generate action proposals of our ActEV system.

\section{Efficient Object Detection and Tracking}

The activities of concern in ActEV are summarized in Table~\ref{tab:0101_actev}. 
These activities involve either person or vehicle object, we use this prior knowledge to build the event proposal module starting from the object detection step. 
The output of this step is person and vehicle bounding box for each frame. The immediate natural next step is to associate detected object across frames, which is tracking. The output of this step is person tracklet and vehicle tracklet.

\begin{table}[h]
\centering
\begin{tabular}{|l|l|}
\hline 
Type           & Activities                                                                                                                                                \\ \hline
Person-Only    &  \makecell[l]{ Transport\_HeavyCarry, Riding, Talking, Activity\_carrying, \\ Specialized\_talking\_phone, Specialized\_texting\_phone, \\ Entering, Exiting, Closing, Opening} \\ \hline
Vehicle-Only   & Vehicle\_turning\_left, Vehicle\_turning\_right, Vehicle\_u\_turn                                                                                         \\ \hline
Person-Vehicle & Open\_Trunk, Loading, Closing\_trunk, Unloading \\ \hline                                                                                                         
\end{tabular}
\caption{Activity categorization on the ActEV dataset.}
\label{tab:0101_actev} 
\end{table}

\subsection{Object Detection}
\label{sec:od}
We utilize Mask-RCNN~\cite{he2017mask} with feature pyramid network~\cite{lin2017feature} on ResNet~\cite{he2016deep} as the backbone for object detection, in which RoIAlign~\cite{he2017mask} is used to extract features for Region-of-Interest. 
We also experiment with the latest state-of-the-art EfficientDet~\cite{tan2020efficientdet} model.
The object detection models are trained on either MSCOCO~\cite{lin2014microsoft} object detection training set or the VIRAT~\cite{oh2011large} training set.
We apply object detection on every k frame (we find 8 to be a good speed-and-accuracy-trade-off number) in the videos. 
Full resolution images (1920x1080) or HD resolution (1280x720) images are input to the model to generate object bounding boxes.

\noindent \textbf{Experimental Results.}
We first experiment with a number of variations of Mask-RCNN and EfficientDet models on the VIRAT dataset, a representative dataset of extended videos, and the AVA-Kinetics~\cite{li2020ava} dataset, a recent action detection dataset based on Internet videos.
In these experiments, we mainly evaluate the ``Person'' and/or ``Vehicle'' object classes, as they are the most dominant and useful classes for action detection and prediction.


\begin{table}[]
\centering
\begin{tabular}{|l|c|c|c|c|}
\hline
                                          & COCO-mAP & \makecell[c]{ VIRAT-Person \\ Val-AP} &  \makecell[c]{ VIRAT-Vehicle \\ Val-AP} &  \makecell[c]{ VIRAT-Bike \\ Val-AP} \\ \hline
MaskRCNN, R50-FPN                         & 0.389                        & 0.374		              & 0.943		              & 0.367           \\ \hline
MaskRCNN, R101-FPN                        & 0.407                        & 0.378	              & 0.947	            & 0.399             \\ \hline
EfficientDet-d2                           & 0.425                        &    	      0.371           &    	       0.949         &          0.293       \\ \hline
EfficientDet-d6                           & 0.513                        &  0.422	                 &      0.947	                &       0.355             \\ \hline
MaskRCNN, R101-FPN*             & -                            & 0.831               & 0.982                & 0.588             \\  \hline
\end{tabular}
\caption{Object detection evaluation on the VIRAT dataset. ``*'' is finetuned on the VIRAT training set.}
\label{tab:0101_virat_obj}
\end{table}

Table~\ref{tab:0101_virat_obj} shows the object detection results on the VIRAT dataset. As we see, for MaskRCNN, Resnet-50 backbone and Resnet-101 backbone perform similarly on detection persons and vehicles. Further finetuning helps tremendously on the VIRAT dataset. 
The EfficientDet models are significantly better than MaskRCNN overall on all 80 classes on the MS-COCO dataset. 
However, on the three objects we focus on in the VIRAT dataset, the improvement is not efficient: especially for EfficientDet-d6, the computation cost is more than 2x compared to MaskRCNN, R101-FPN.

Since our goal is to develop an efficient object detection and tracking framework for extended videos, we experiment with different infrastructure and different code bases. 
As shown in Table~\ref{tab:0101_speed_machine}, we experiment with 3 machines, with different GPUs, CPUs and I/O condition.
In Table~\ref{tab:0101_speed_run}, we show the experiments that we have run. We mainly test different versions of Tensorflow as well as TensorRT acceleration on the VIRAT validation set, with about 206k images.
We use full resolution (1920x1080) inputs.
The experiments are shown in Table~\ref{tab:0101_speed}. All these runs produce the same object detection results.
As we see, later version of Tensorflow with later version of CUDA improves speed significantly.
The frozen graph method in Tensorflow also helps.
The TensorRT acceleration does not show significant improvement with Tensorflow version 1.14.
In conclusion, it is enough to use frozen graph of Tensorflow models for object detection inferencing. Other ways to increase speed include using lower resolution input images and smaller backbone like Resnet-50, but these methods will lead to a decrease in detection average precision.

\begin{table}[]
\centering
\begin{tabular}{|l|l|}
\hline
\multicolumn{2}{|l|}{Machine Type}                   \\ \hline
Machine 1 & 2 GTX 1080 TI, i7, nvme             \\ \hline
Machine 2 & 3 GTX 1080 TI + 1 TITAN X, E5, nvme \\ \hline
Machine 3 & 4 RTX 2080 TI , i9-9900X, SSD       \\ \hline
\end{tabular}
\caption{Machine types that we use to test object detection model speed.}
\label{tab:0101_speed_machine}
\end{table}

\begin{table}[]
\centering
\begin{tabular}{|r|l|}
\hline
\multicolumn{2}{|l|}{Run}                                                          \\ \hline
1 & tf 1.10 (CUDA 9, cudnn 7.1), Variable Model                                        \\ \hline
2 & tf 1.13 (CUDA 10.0 cudnn 7.4), Variable Model                                      \\ \hline
3 & tf 1.13 (CUDA 10.0 cudnn 7.4), Frozen Graph (.pb)                                  \\ \hline
4 & tf 1.13 (CUDA 10.0 cudnn 7.4), Frozen Graph (.pb) -\textgreater TensorRT Optimized \\ \hline
5 & tf 1.14.0 (CUDA 10.0 cudnn 7.4), Variable Model                                    \\ \hline
6 & tf 1.14.0 (CUDA 10.0 cudnn 7.4), Frozen Graph (.pb)                                \\ \hline
7 & tf 1.14.0 (CUDA 10.0 cudnn 7.4), Frozen Graph (.pb) -\textgreater TRT FP32         \\ \hline
8 & tf 1.14.0 (CUDA 10.0 cudnn 7.4), Frozen Graph (.pb) -\textgreater TRT FP16         \\ \hline
\end{tabular}
\caption{Experiments that we run to test the model speed on VIRAT.}
\label{tab:0101_speed_run}
\end{table}

\begin{table}[]
\centering
\begin{tabular}{|c|c|c|c|c|}
\hline
Run & Machine & \# GPU Used & GPU Average Utilization & Per GPU FPS \\ \hline
1       & 1       & 2           & 65.3\%                  & 2.5         \\ \hline
2       & 1       & 2           & 62.0\%                  & 2.72        \\ \hline
2       & 2       & 4           & 33.8\%                  & 1.76        \\ \hline
3       & 2       & 4           & 23.5\%                  & 1.95        \\ \hline
3       & 2       & 2           & 38.0\%                  & 2.87        \\ \hline
2       & 2       & 1           & 41.4\%                  & 2.78        \\ \hline
3       & 2       & 1           & 54.8\%                  & 3.56        \\ \hline
2       & 2       & 1*4         & 46.2\%                  & 2.94        \\ \hline
3       & 2       & 1*4         & 52.3\%                  & 3.54        \\ \hline
5       & 2       & 1*4         & 53.2\%                  & 3.03        \\ \hline
6       & 2       & 1*4         & 61.7\%                  & 3.84        \\ \hline
7       & 2       & 1*4         & 61.0\%                  & 3.93        \\ \hline
8       & 2       & 1*4         & 52.6\%                  & 3.89        \\ \hline
2       & 3       & 1           & 61.2\%                  & 3.57        \\ \hline
3       & 3       & 1           & 61.2\%                  & 4.74        \\ \hline
2       & 3       & 1*4         & 62.6\%                  & 3.65        \\ \hline
3       & 3       & 1*4         & 65.2\%                  & 4.83        \\ \hline
\end{tabular}
\caption{Speed experiments on VIRAT with the MaskRCNN model. ``1*4'' means 4 inference processes are run concurrently, where one process utilizes one GPU.}
\label{tab:0101_speed}
\end{table}

\begin{table}[]
\centering
\begin{tabular}{|l|c|c|c|}
\hline
                              & COCO-mAP & \makecell[c]{AVA-Kinetics \\ Train-Person-AP} & \makecell[c]{AVA-Kinetics \\ Val-Person-AP} \\ \hline
MaskRCNN, R101-FPN            & 0.407                        & 0.664                        & 0.682                      \\ \hline
EfficientDet-d2               & 0.425                        & 0.650                        & 0.680                      \\ \hline
EfficientDet-d6               & 0.513                        & 0.623                        & 0.658                      \\ \hline
MaskRCNN, R101-FPN* & -                            & 0.735                        & 0.732                      \\ \hline
\end{tabular}
\caption{Person detection evaluation on the AVA-Kinetics dataset. ``*'' is finetuned on the AVA-Kinetics training set.}
\label{tab:0101_ava-kinetics_obj}
\end{table}

Table~\ref{tab:0101_ava-kinetics_obj} shows the person detection results on the AVA-Kinetics dataset.
As we see, even though the EfficientDet models perform better than MaskRCNN on COCO mean average precision over all 80 object classes, MaskRCNN is better on person detection on the AVA-Kinetics dataset.
Further finetuning improves further.
In later \autoref{chap:0103_viewpoint}, we will utilize the finetuned model for action detection experiments.

\subsection{Object Tracking}
We compare the performance of deep SORT~\cite{wojke2017simple} and kernelized correlation filter (KCF)~\cite{henriques2014high}. As shown in Table ~\ref{tab:0101_tracking}, deep SORT outperforms KCF for all the metrics except precision.
As the tracking module is used to generate tracklets which are proposal candidates, we expect a high recall and low ID switches with a comparable precision. 
The results are reported in Table ~\ref{tab:0101_tracking}.
In the final system, we utilize deep SORT~\cite{wojke2017simple} to generate tracklets by associating detected objects across frames. 
We follow a similar track handling and Kalman filtering framework ~\cite{henriques2014high}. 
We use bounding box center positions, aspect ratio, height and their respective velocities in image coordinates as Kalman states. We compute the Mahalanobis distance between predicted Kalman states and newly arrived measurement to incorporate motion information. 
For each bounding box detection, we use the feature obtained from object detection module as a appearance descriptor (either from the Resnet backbone or the EfficientNet backbone). 
We compute the cosine distance between tracks and detections in appearance space. 
To build the association problem, we combine both metrics using a weighted sum. An association is defined admissible if it is within the gating region of both metrics.

\begin{table}[]
\centering
\begin{tabular}{|l|c|c|c|c|c|}
\hline
Models    & Recall (\%) & Precision (\%) & ID switches & MOTA (\%) & MOTAL (\%) \\ \hline
KCF       & 93.5        & \textbf{97.1}           & 2519        & 91.3      & 90.5       \\ \hline
deep SORT & \textbf{95.2}        & 96.5           & \textbf{909}         & \textbf{91.7}      & \textbf{91.8}       \\ \hline
\end{tabular}
\caption{Results of multi-object tracking algorithms in the validation set of VIRAT dataset.}
\label{tab:0101_tracking}
\end{table}

\begin{figure}[ht]
	\centering
		\includegraphics[width=0.95\textwidth]{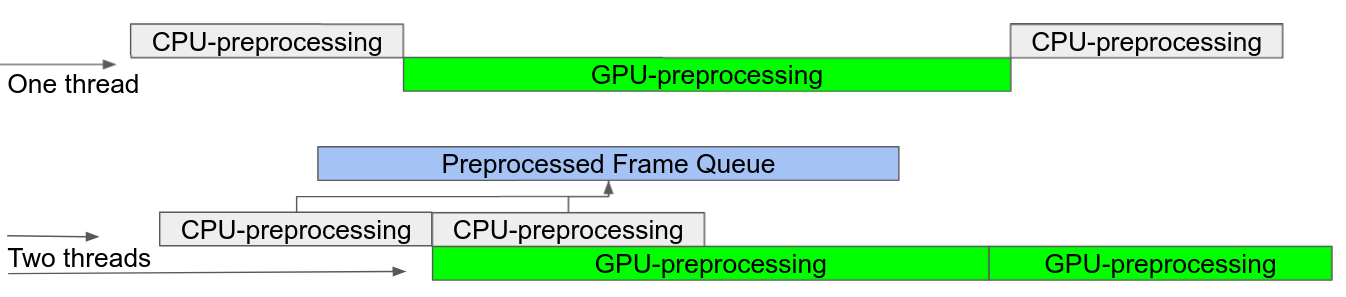} 
	\caption{Example of the efficient processing pipeline.}
	\label{fig:0101_efficient_od}
\end{figure}

\subsection{Efficient Processing}

Our system uses Python Queues to preprocess videos with CPUs while allowing GPUs to run inference on multiple images at the same time.
See Figure~\ref{fig:0101_efficient_od}. With parallel processing threads, we are able to leverage CPUs and GPUs efficiently. 

\begin{figure}[ht]
	\centering
		\includegraphics[width=0.95\textwidth]{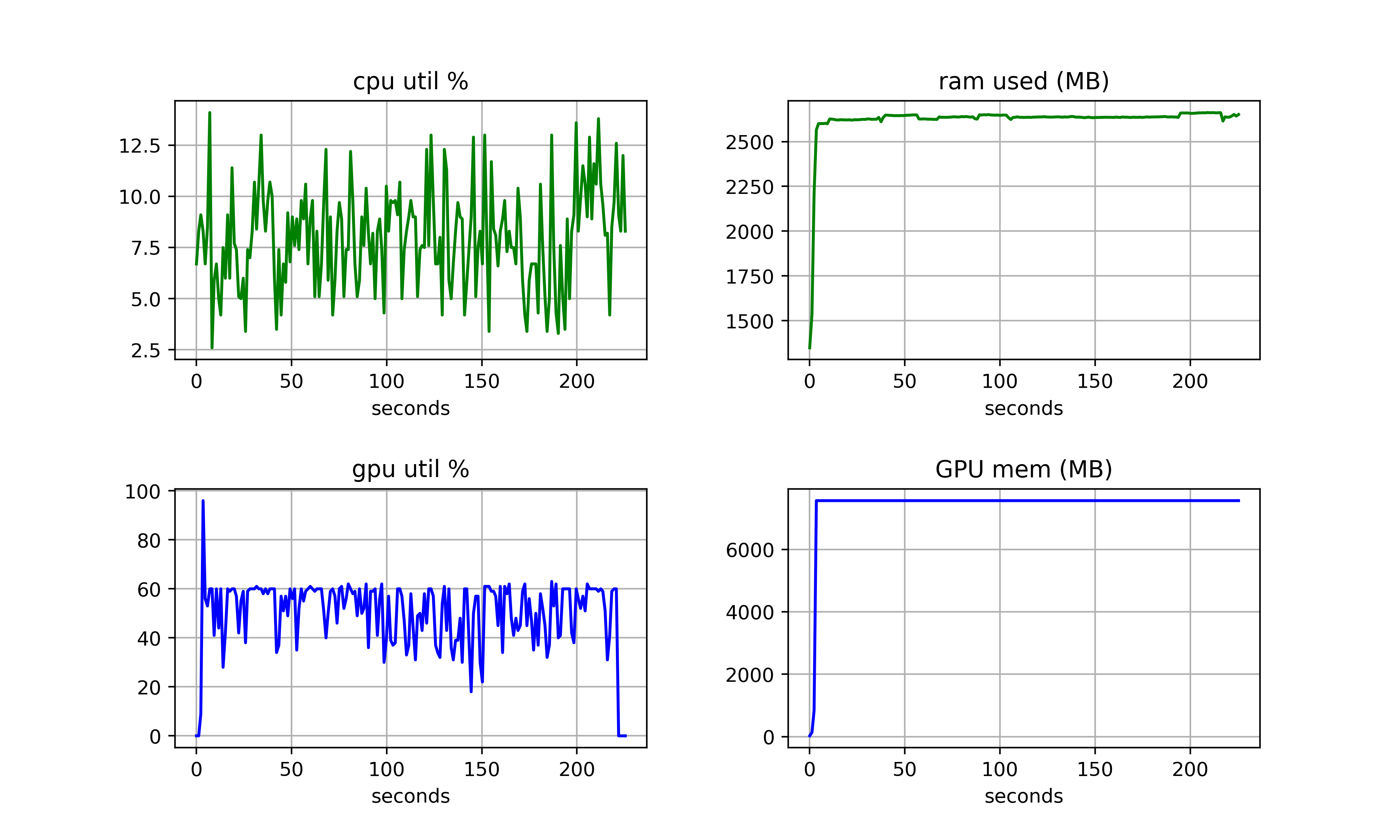} 
	\caption{CPU and GPU utilization graph for b=1 frozen,partial experiment. See text for details}
	\label{fig:0101_efficient_od_util1}
\end{figure}

\begin{figure}[ht]
	\centering
		\includegraphics[width=0.95\textwidth]{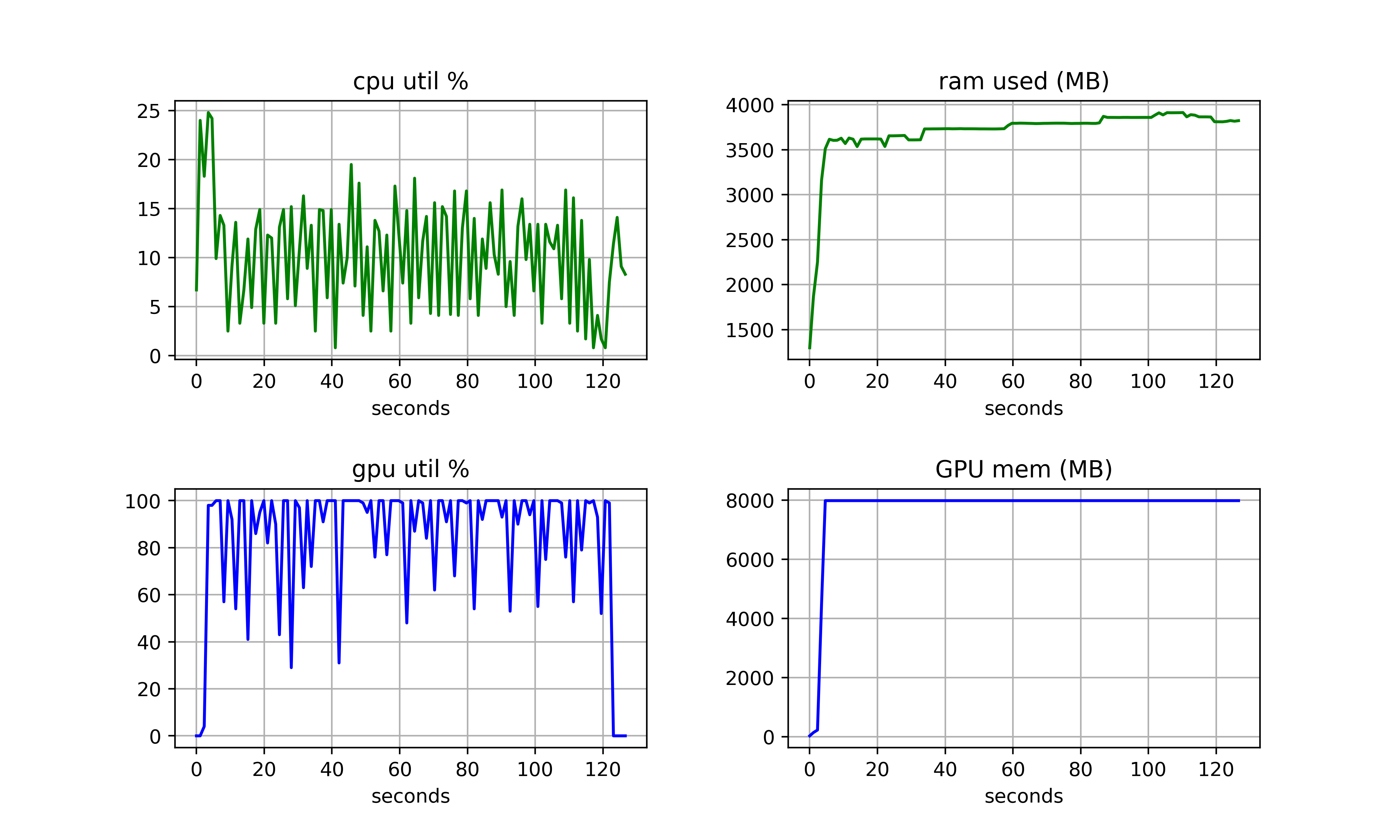} 
	\caption{CPU and GPU utilization graph for b=8 frozen,partial,m experiment. See text for details}
	\label{fig:0101_efficient_od_util8}
\end{figure}

\noindent\textbf{Experiment}. We conduct an experiment on a machine with a GTX 1070 TI GPU, i7-8700K CPU and SSDs. We use the FPN-ResNet50 object detection model. We run object detection and tracking on a 5-minutes video of 1280x720 resolution. Results are shown in Table~\ref{tab:0101_efficient_od}. ``var'', ``frozen'', ``partial'' means variable, frozen, partial (only output ``Person'' and ``Vehicle'' objects) object detection model, respectively. See section~\ref{sec:od} for more details on the different type of Tensorflow models. ``m'' means running the system with the aforementioned parallel processing method. As we see, the total processing time is reduced from 6 minutes to 2 minutes. In Figure~\ref{fig:0101_efficient_od_util1} and Figure~\ref{fig:0101_efficient_od_util8}, we show the CPU and GPU utilization graph for ``b=1 frozen,partial'' and ``b=8 frozen,partial,m'' experiments, respectively. As we see, the usage of the system is improved significantly.

\begin{table}[ht]
\centering
\begin{tabular}{|l|c|c|c|}
\hline
RunType & Time  & GPU Median Utilization & GPU Average Utilization \\ \hline
b=1 var                       & 06:21 & 53.00\%                & 54.24\%                 \\ \hline
b=1 frozen                    & 05:06 & 34.50\%                & 36.30\%                 \\ \hline
b=1 frozen,partial            & 03:43 & 57.00\%                & 49.55\%                 \\ \hline
b=8 var                       & 04:27 & 00.00\%                & 2.35\%                  \\ \hline
b=8 frozen                    & 03:12 & 62.00\%                & 53.37\%                 \\ \hline
b=8 frozen,partial            & 03:07 & 75.50\%                & 62.11\%                 \\ \hline
b=8 var,m                     & 03:29 & 100.00\%               & 73.45\%                 \\ \hline
b=8 frozen,m                  & 02:14 & 67.00\%                & 63.69\%                 \\ \hline
b=8 frozen,partial,m          & 02:08 & 99.00\%                & 83.83\%                 \\ \hline
\end{tabular}
\caption{Efficient object detection and tracking experiments.}
\label{tab:0101_efficient_od}
\end{table}


\section{Contributions}
Our object detection and tracking system is a fundamental component for the rest of the thesis. 
This system also supports multi-view person/vehicle long-term tracking with re-identification models.
This system is used in our submission to the Activities in Extended Videos Prize Challenge (ActEV)~\cite{awad2020trecvid,liu2020argus}, where we achieves \textbf{best performance} on the leaderboard.
This part of the thesis has been open-sourced at \url{https://github.com/JunweiLiang/Object_Detection_Tracking}. As of June 2021, this repository receives \textbf{240} stars and 82 forks, with a weekly traffic of \textbf{200} viewers and a dozen downloads. 

\chapter{Weakly-supervised Action Event Recognition 
}  \label{chap:0102_weakly}

Learning detectors that can recognize concepts, such as people actions in video content is an interesting but challenging problem for human behavioral analysis.
However, as human actions are diverse and combination of atomic actions can lead to an exponential amount of action classes, it is often difficult to have sufficient human-annotated training data for action recognition and detection.
In this chapter, we present our work on webly-labeled learning with multimodal video data~\cite{liang2017leveraging,liang2016learning,liang2017webly} to tackle the challenge of not enough manual annotations for action recognition.
In the next chapter (\autoref{chap:0103_viewpoint}), we propose viewpoint-invariant feature representation learning for action recognition and detection that could be generalized to multiple action datasets.

\section{Motivation}
Millions of videos are being uploaded to the Internet every day. These videos capture different aspects of multimedia content about our daily lives. Automatically categorizing videos into concepts, such as people actions, events, etc., is an important topic. Recently many studies have been proposed to tackle the problem of concept learning \cite{deng2009imagenet,liang2015towards,karpathy2014large,jiang2015fast,abu2016youtube,liang2014semantic}. 

Many datasets acquire the clean concept labels via annotators. These datasets include ImageNet~\cite{deng2009imagenet}, TRECVID MED~\cite{over2014trecvid} and FCVID~\cite{jiang2015exploiting}. Collecting such datasets requires significant human effort, which is particularly expensive for video data. As a result, the labeled video collection is usually much smaller than the image collection. For example, FCVID~\cite{jiang2015exploiting}, only contains about 0.09 million labels on 239 concept classes, much less than the 14 million labels on over 20,000 classes in the image collection ImageNet~\cite{deng2009imagenet}. On the other hand, state-of-the-art concept models utilize deep neural networks~\cite{karpathy2014large}, which need more data to train. However, training only on manually labeled clean data seem insufficient for large-scale concept learning. 

\begin{figure}
	\centering
		\includegraphics[width=0.7\textwidth]{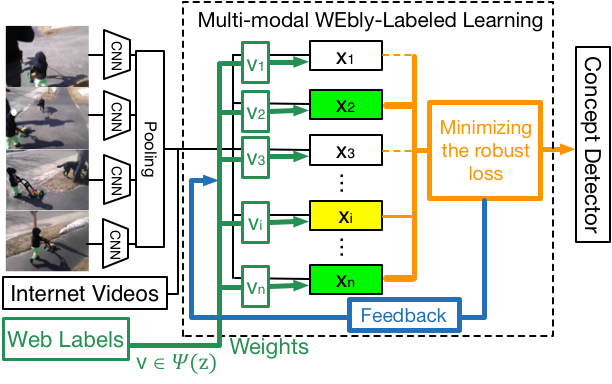}
	\caption{Overview of Multi-modal WEbly-Labeled Learning (WELL-MM). The algorithm jointly models the prior knowledge extracted from web labels and the current learned model at each iteration to overcome the noise labels. $\{\mathbf{x}_{i}\}_{i=1}^{n}$ are input samples and their current weights are determined by $\{\mathbf{v}_{i}\}_{i=1}^{n}$. Colored samples are the samples with nonzero weights at the current iteration. The blue line indicates the feedback from the previous objective function value.}
	\label{fig:well-cnn}
\end{figure}

\begin{figure}[!t]
	\centering
		\includegraphics[width=0.7\textwidth]{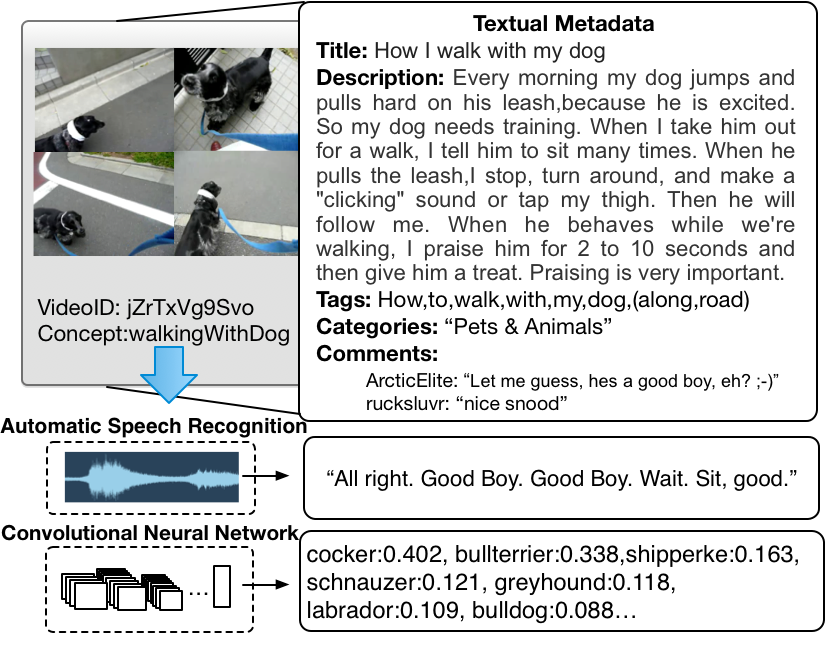}
	\caption{Multi-modal prior knowledge from web video.} 
	\label{fig:cur-example}
\end{figure}

Images or videos on the web often contain rich contextual information, such as their titles or descriptions. We can infer the label by the metadata. Fig.~\ref{fig:cur-example} shows an example of a video with inferred concept label ``walking a dog''. In this chapter, we call the samples with inferred labels \emph{weakly labeled} or \emph{webly labeled}. The webly-labeled data are easy to collect and thus usually orders-of-magnitude larger than manually-labeled data. However, the web labels are very noisy
and have both low accuracy and low recall.

Concept learning over weakly labeled data becomes popular as it allows for large-scale learning on big data. However, these methods have only focused on utilizing a single text modality to model the noisy labels~\cite{liang2016learning,chen2015webly}. For example, in Fig.~\ref{fig:cur-example}, the textual metadata is useful but also contain lots of noises. In fact, the video is of multiple modalities and our intuition is that the inference obtained from multiple modalities is more reliable than that from a single text modality. For example, we can more confident to say this video is about ``walk a dog'' if we spot the text in the title, hear the words ``good boy'' in the speech, and see a dog in some key frames. To this end, we can leverage the prior knowledge in automatically extracted multi-modal features from the video content such as pre-trained still image detectors, automatic speech recognition and optical character recognition. In some cases when videos have little textual metadata, multi-modal knowledge become the only useful clues in concept learning.

Recent studies on weakly labeled concept learning show promising results. However, since existing approaches only focuses on a single modality, two important questions have yet: 1) what are the important multi-modal prior knowledge, except textual metadata, for modeling noisy labels? 2) how to integrate the multiple modalities into concept learning in a theoretically sound manner? 

In this chapter, to utilize multi-modal prior knowledge for concept learning, we propose a learning framework called \textbf{Multi-modal WEbly-Labeled Learning (WELL-MM)}. The learning framework is motivated by human learning, in which the learner starts from learning easier aspects of a concept, and then gradually take more complex examples into the learning process\cite{bengio2009curriculum,kumar2010self,jiang2015self}. Specifically. WELL-MM learns a concept detector iteratively from first using a few samples with more confident labels, then gradually incorporate more samples with noisier labels.
Fig.~\ref{fig:well-cnn} shows the overview of the proposed framework. The algorithm integrates multi-modal prior knowledge, which is derived from the multi-modal video and image features, into the dynamic learning procedure. The idea of curriculum and self-paced learning paradigm has been proved to be efficient to deal with noise and outliers~\cite{chen2015webly,kumar2010self}. Our proposed method is the first to generalize the learning paradigm to leverage multi-modal prior knowledge into concept learning. Experimental results show that multi-modal prior knowledge is important in concept learning over noisy data. The proposed WELL-MM outperforms other weakly labeled learning methods on three real-world large-scale datasets, and obtains the state-of-the-art results with recent deep learning models.

The contribution of this chapter is threefold. First, we propose a novel solution to address the problem of weakly labeled data learning through a general framework that considers multi-modal prior knowledge. We show that the proposed WELL-MM not only outperforms state-of-the-art learning methods on noisy labels, but also, notably, achieves comparable results with models trained using manual annotation on one of the  video dataset. Second, we provide valuable insights by empirically investigating different multi-modal prior knowledge for modeling noisy labels. Experiments validate that by incorporating multi-modal information, our method is robust against certain levels of noisiness. Finally, the efficacy and the scalability have been demonstrated on three public large-scale benchmarks, which include datasets on both Internet videos and images. The promising results suggest that detectors trained on sufficient weakly labeled videos may outperform detectors trained on existing manually labeled datasets. 

\section{Multi-modal WEbly-Labeled Learning (WELL-MM)}
\subsection{Problem Description}
In this chapter, following~\cite{varadarajan2015efficient}, we consider a concept detector as a classifier and our goal is to train concept detectors from webly-labeled video data without any manually annotated labels. Given a collection of training samples with noisy labels, we do not make any assumption over the underlying noise distribution. 
Formally, we represent the training set as $\mathcal{D} = \{(\mathbf{x}_i,\mathbf{z}_i,\mathbf{\tilde{y}}_i)\}_{i=1}^n$, 
where $\{\mathbf{x}_1,\cdots,\mathbf{x}_n \}$ are the d-dimensional features of the training set,
and $\{\mathbf{z}_1,\cdots,\mathbf{z}_n \}$ represent each sample's corresponding noisy web labels. We assume that the noisy labels are given. The noisy web labels are often automatically inferred using the sample's textual metadata provided by its uploader, or from other modalities such as pre-trained convolutional neural network over still images\cite{chatfield2014return}, Automatic Speech Recognition \cite{povey2011kaldi}, or Optical character recognition \cite{smith2007overview}. For example, for instance, a video might have a noisy label ``cat'' as its title and speech both contain the word cat. The $\mathbf{\tilde{y}}_i \subset \mathcal{Y}$ is the inferred concept label set for the $i^{th}$ observed sample based on its noisy web label, and $\mathcal{Y}$ denotes the full set of target concepts. In our experiment, to simplify the problem, we employ one-versus-all strategy for multi-class classification, and discuss our method in the context of binary classification over the noisy web labels.

\subsection{Model}

\subsubsection{Objective Function}
In this section, we propose a model called Multi-modal WEbly-Labeled Learning (WELL-MM) to leverage multi-modal prior knowledge for weakly labeled data. 
Formally, given the training set $\mathcal{D}$ mentioned previously,  Let $L(\tilde{y}_i,g(\mathbf{x}_i,\mathbf{w}))$,
denote the loss function which calculates the cost between the inferred label $\tilde{y}_i$ and the estimated label given by the decision function $g(\mathbf{x}_i,\mathbf{w})$. Here $\mathbf{w}$ represents the model parameters. 
Our objective function is to jointly learn the model parameter $\mathbf{w}$ and the latent weight variable $\mathbf{v}= [v_1,\cdots,v_n]^T$ by:
\begin{equation}    \label{eq:spcl_obj}
\begin{split}
\min_{\mathbf{w},\mathbf{v}\in \lbrack 0,1]^{n}}  \mathbb{E}(\mathbf{w}, \mathbf{v} ;\lambda, \Psi ) = 
 \sum_{i=1}^n v_i L(\tilde{y}_i,g(\mathbf{x}_i,\mathbf{w})) + f(\mathbf{v}; \lambda), \\ 
\text{  subject to } \mathbf{v} \in \Psi
\end{split}
\end{equation}
where the latent weight variable $\mathbf{v=[}v_{1},\cdots ,v_{n}\mathbf{]}^{T}$ represents the inferred labels' confidence, and thus reflects the learning sequence of samples.
In order to learn concept detectors in noisy data, we utilize the self-paced regularizer $f$~\cite{jiang2015self} to control the learning process, where $f$ is expect to assign greater weights to samples with confident labels. For simplicity, we consider the linear regularizer Eq.~\eqref{eq:linear_scheme} proposed in~\cite{jiang2015self}:
\begin{equation}
\label{eq:linear_scheme}
f(\mathbf{v};\lambda) =  \frac{1}{2} \lambda \sum_{i=1}^n  ( v_i^2 - 2v_i ),
\end{equation}
$\lambda \in (0,1)$ is a hyper-parameter that controls the pace of model training, which resembles the "age" of the model. We set $\lambda$ to be small at the beginning and only samples of with small loss will be considered in training. As $\lambda$ grows, more samples with larger loss will be gradually included. As stated in related studies~\cite{liang2016learning,meng2015objective}, the self-paced in Eq.~\eqref{eq:linear_scheme} corresponds to a robust loss function. The robust loss in our problem tends to depress samples with noisy labels
or outliers and thus may be instrumental in avoiding bad local minima.

In order to utilize the rich contextual information in the noisy data, we embed the multi-modal prior knowledge derived from the web labels $\mathbf{z}$ into a convex curriculum region $\Psi$ for the latent weight variables. The shape of the region weakly implies the learning sequence, where favored samples have larger expected values. Generally, $\Psi$ can be represented by $\Psi = \{\mathbf{v} \,|\, c(\mathbf{v},\mathbf{a})\le b\}$, where $\mathbf{a} = [a_1,\cdots,a_n]$ is the parameters of the region. In this chapter, we use a linear constraint to form the curriculum region \cite{jiang2015self}:
\begin{equation}
\Psi = \{\mathbf{v}\, | \,\sum_{i=1}^n a_i v_i \le b\}
\end{equation}
 
The curriculum region is introduced to leverage the prior knowledge about the noisy labels and, as demonstrated in our experiments, is a crucial factor in weakly labeled data learning. 
We use multi-modal information to derive the probabilities of samples being positive of a class and if the probabilities are below a threshold (in our experiments it is set at zero) the samples will be consider as negatives. We assign value to $a_i$ in correlated to samples 's probabilities being in the class, and $b$ is set to 1. We use curriculum as a warm start in training, and set $\mu$ to zero after the first iteration. Since we empirically observed that curriculum constraints mostly benefit the first few iterations.
We will discuss how to derive the multi-modal curriculum in details in the following section.

Eq.~\eqref{eq:spcl_obj} is difficult to minimize over big data due to the constraints. In this work, we propose to relax the constraints by introducing a Lagrange multiplier $\mu$. The objective function then becomes:
\begin{equation}    
\begin{split}
\label{eq:spcl_obj1}
&\min_{\mathbf{w},\mathbf{v}\in \lbrack 0,1]^{n}}  \mathbb{E}(\mathbf{w}, \mathbf{v} ;\lambda, \mathbf{a},b, \mu) = \\
 &\sum_{i=1}^n v_i L(\tilde{y}_i,g(\mathbf{x}_i,\mathbf{w})) +  \frac{1}{2} \lambda \sum_{i=1}^n  ( v_i^2 - 2v_i ) + \mu(\sum_{i=1}^n a_i v_i - b), \\ 
& \quad \quad \quad \quad \quad \quad \quad \quad \quad \quad \quad \quad \quad \quad \quad \quad \quad \text{  subject to } \mu \ge 0 
\end{split}
\end{equation}

The proposed Eq.~\eqref{eq:spcl_obj1} has two benefits over Eq.~\eqref{eq:spcl_obj}. First it enables the large-scale training on noisy data. This is important because as our experiments show that training on noisy data can outperform training on manually labeled data only when the noisy data are orders-of-magnitude larger. Second, it may tolerate the noise introduced in the curriculum region.


\subsubsection{Multi-modal Curriculum}
In this section we discuss the details on how to construct the curriculum region $\Psi$. $\Psi$ is a feasible region that embeds the multi-modal prior knowledge extracted from the webly-labeled data as shown in Figure.~\ref{fig:cur-example}. 
It geometrically corresponds to a convex feasible space for the latent weight variable. 
Given a set of training samples $\mathbf{X}=\{\mathbf{x}_{i}\}_{i=1}^{n}$ with corresponding noisy labels $\mathbf{Z}=\{\mathbf{z}_{i}\}_{i=1}^{n}$, we want to extract the learning curriculum based on how related the training samples are to the target classes, which is modeled by the probability of the samples being the inferred class label (since we don't have the actual label in webly learning). 
The training samples with a greater value of probability mean that they are more confident to belong to the true class and should be learned earlier. 
Similar to Information Retrieval theory~\cite{manning2008introduction}, here we use random variable $\mathbf{z}$ to represent the noisy web labels, $\mathbf{y}$ to represent the label classes, and the curriculum for a sample is then determined by:

\begin{equation}
\label{m_pp}
\mathbf{P}(\mathbf{z}\,|\,\mathbf{y}) = \mathbf{P}(\mathbf{y}\,|\,\mathbf{z})
\mathbf{P}(\mathbf{z})/\mathbf{P}(\mathbf{y})
\end{equation}
Since $\mathbf{P}(\mathbf{y})$ is the same for all samples, it can be regarded as a constant. The prior probability of a web video $\mathbf{P}(\mathbf{z})$ can be implemented with the duration, the view count and comments about the video. In this work we treat the prior as uniform so it can be ignored as well. Therefore, we calculate the curriculum simply based on $\mathbf{P}(\mathbf{y}\,|\,\mathbf{z})$, the probability of the sample being class $\tilde{y_i}$ given the noisy label. Since we want to incorporate the multi-modal prior information, we calculate the curriculum from:

\begin{equation}
\label{m_p}
\mathbf{P}(\mathbf{y}\,|\,\mathbf{z}) \propto \sum_{m} \theta_m \mathbf{P}(\mathbf{y}\,|\,\mathbf{z_m})
\end{equation}
We use random variable $z_m$ to represent the $m$-th modality of the noisy labels for a sample and $\theta_m$ is the predetermined weight for modality $m$. 
In this work, other than the textual metadata, we also utilize other modalities such as Automatic Speech Recognition (ASR) \cite{povey2011kaldi}, Optical Character Recognition (OCR) \cite{smith2007overview} and basic image detector pre-trained on still images~\cite{ILSVRC15} (in this work we use VGG net~\cite{simonyan2014very}, extract keyframe-level image classification results and average them to get video-level results). Therefore the total number of the modalities is 4. We compare common ways to extract curriculum from web data for concept learning to the proposed novel method that utilizes state-of-the-art topic modeling techniques in natural language processing.

In the following methods (Word Hard Matching and Latent Topic with Word Embedding), we first extract bag-of-words features from different modalities for each video and then match them using specific matching methods to the concept words to get the probabilities in Eq.~\eqref{m_p}.

\textbf{Word Hard Matching} We build curriculum directly using exact word matching or stemmed word matching between the textual metadata of the noisy videos to the targeted concept names. This is the same method as stated in Webly Labeled Learning~\cite{liang2016learning}. Noted that this method only utilizes one modality.

\textbf{YouTubeTopicAPI} The YouTube topic API is utilized to search for videos that are related to the concept words. The topic API uses textual information of the uploaded videos to obtain related topics of the videos from Freebase. This is the method used in \cite{varadarajan2015efficient}.

\textbf{SearchEngine} The curriculum is built using the search result from a text-based search engine~\cite{mccandless2010lucene}. It is similar to related web-search based methods.

\textbf{Ours} We build the curriculum based on the latent topic we learned from the noisy label. We incorporate Latent Dirichlet Allocation (LDA)~\cite{blei2003latent} to determine how each noisy labeled video is related to each target concept. 
The intuition is that each web video consists of mixtures of topics (concepts), and each topic is characterized by a distribution of words. 
We impose asymmetric priors over the word distribution so that each learned topic will be seeded with particular words in our target concept. For example, a topic will be seeded with words "make, phone, cases" for the target concept "MakingPhoneCases". 
we use the online variational inference algorithm from \cite{hoffman2010online}. We then match noisy labels from each modality $\mathbf{z_{im}}$ to the latent topic word distribution using word embedding soft matching~\cite{mikolov2013distributed}. The word embedding is pre-trained using Google News data.

Our method can utilize information from different modalities while common methods like search engine only consider textual information.
We compare the performance of different ways of curriculum design by training detectors directly in Section 4.

\subsection{Algorithm}
As proven in recent studies~\cite{meng2015objective,kumar2010self}, Eq.~\eqref{eq:spcl_obj} is a biconvex optimization problem. We utilize the alternative convex search algorithm (ACS)~\cite{bazaraa2013nonlinear} to optimize Eq.~\eqref{eq:spcl_obj} following~\cite{kumar2010self,jiang2015self}.
Algorithm~\ref{alg:well} takes the input of the training set, an instantiated self-paced regularizer and the curriculum constraint function; it outputs an optimal model parameter $\mathbf{w}$. it derives the curriculum region from multi-modal noisy labels $\mathbf{Z} \in \mathbb{R}^{m \times n}$ and forms the curriculum constraint function. 
Then, it initializes the latent weight variables in the feasible region. 
In the while loop, the algorithm alternates between two steps until it finally converges: 
In step 4 given the most recent $\mathbf{v}^*$, the algorithm learns the optimal model parameters; 
In step 5, we fix the $\mathbf{w}^*$ and the algorithm learns the optimal weights $\mathbf{v}^*$ for each sample.
Starting in the beginning, the model grows from learning with easy (less noisy) samples with a small model "age". The model "age" is gradually increased so that the model can incorporate more noisy samples in the training and become more robust over time.
Step 4 can be implemented by existing off-the-shelf supervised learning methods such as the Support Vector Machine or back propagation. 
Gradient-based methods can be used to solve the convex optimization problem in Step 5. According to~\cite{gorski2007biconvex}, the alternative search in Algorithm~\ref{alg:well} converges as the objective function is monotonically decreasing and is bounded from below.

\begin{algorithm}
\SetKwData{Left}{left}\SetKwData{This}{this}\SetKwData{Up}{up}
\SetKwInOut{Input}{input}\SetKwInOut{Output}{output}
\LinesNumbered
\Input{Input dataset $\mathcal{D} = \{\mathbf{X},\mathbf{Z},\mathbf{\tilde{Y}} \}$, self-paced function $f$ and a curriculum constraint function $c$}
\Output{Model parameter $\mathbf{w}$}
\BlankLine
Derive curriculum region from $\mathbf{Z} \in \mathbb{R}^{m \times n}$ into $\mathbf{a},b$\; 

Initialize $\mathbf{v}^*$, $\lambda$ in the curriculum region\;

\While{not converged} {
Update $\mathbf{w}^* = \arg\min_{\mathbf{w}} \mathbb{E}(\mathbf{w},\mathbf{v}^*;\lambda, \mathbf{a},b)$\;

Update $\mathbf{v}^* = \arg\min_{\mathbf{v}} \mathbb{E}(\mathbf{w}^*,\mathbf{v}; \lambda, \mathbf{a},b)$\;

\lIf{$\lambda$ is small}{increase $\lambda$ by the step size}
}
\Return $\mathbf{w}^*$
\caption{Multi-modal WEbly-Labeled Learning}
\label{alg:well}
\end{algorithm}

\section{Experimental Results}
In this section, we evaluate our method WELL-MM for learning video detectors on noisy labeled data. We first conduct our method on noisy learning in image domain. The efficacy of our methods are mainly verified on two major public benchmarks: FCVID and YFCC100M, where FCVID is by far one of the biggest manually annotated video dataset \cite{jiang2015exploiting}, and the YFCC100M dataset is the largest multimedia benchmark \cite{thomee2015yfcc100m}.

\subsection{Experimental Setup}
\textbf{Datasets, Features and Evaluation Metrics}
    Previous studies on noisy learning in image domain have been focusing on noise estimation~\cite{sukhbaatar2014training,xiao2015learning}. We compare our method with them on the synthesized noisy dataset CIFAR-10 generated using code from ~\cite{xiao2015learning}. We report accuracy on each setting along with the results reported in papers~\cite{sukhbaatar2014training,xiao2015learning} experimented on the same dataset.
    
    Fudan-columbia Video Dataset (FCVID) contains 91,223 YouTube videos (4,232 hours) from 239 categories. It covers a wide range of concepts like activities, objects, scenes, sports, DIY, etc. Detailed descriptions of the benchmark can be found in~\cite{jiang2015exploiting}. Each video is manually labeled to one or more categories. In our experiments, we do not use the manual labels in training, but instead we automatically generate the web labels according to the concept name appearance in the video metadata. The manual labels are used only in testing to evaluate our and the baseline methods. Following~\cite{jiang2015exploiting}, the standard train/test split is used.
    The second set is YFCC100M \cite{thomee2015yfcc100m} which contains about 800,000 videos on Yahoo! Flickr with metadata such as the title, tags, the uploader, etc. There are no manual labels on this set and we automatically generate the curriculum from the metadata in a similar way. Since there are no annotations, we train the concept detectors on the most 101 frequent latent topics found in the metadata. 
    There are totally 47,397 webly labeled videos on the 101 concepts for training. 
    
    On FCVID, as the manual labels are available, the performance is evaluated in terms of the precision of the top 5 and 10 ranked videos (P@5 and P@10) and mean Average Precision (mAP) of 239 concepts. On YFCC100M, since there are no manual labels, for evaluation, we apply the detectors to a third public video collection called TRECVID MED which includes 32,000 Internet videos~\cite{over2014trecvid}. We apply the detectors trained on YFCC100M to the TRECVID videos and manually annotate the top 10 detected videos returned by each method for 101 concepts.

\textbf{Implementation Details}
    We build our method on top of a pre-trained convolutional neural network as the low-level features (VGG network~\cite{simonyan2014very}, except in the image experiment we use AlexNet \cite{krizhevsky2012imagenet} as in \cite{xiao2015learning}). We extract the key-frame level features and create a video feature by the average pooling. The same features are used across different methods on each dataset. The concept detectors are trained based on a hinge loss cost function by SVM. Algorithm~\ref{alg:well} is used to train the concept models iteratively and the $\lambda$ stops increasing after 100 iterations. 
    At each iteration, we apply a dropout of 0.5 when sampling negative samples.
    We automatically generate curriculum labels based on the video metadata, ASR, OCR and VGG net 1,000 classification results using latent topic modeling with word embedding matching as shown in Section 3.

\textbf{Baselines in video domain experiment}
    The proposed method is compared against the following five baseline methods which cover both the classical and the recent representative learning algorithms on webly-labeled data. 
    \textit{BatchTrain} trains a single SVM model using all samples in the multi-modal curriculum built with our method as described in section 3.2.2.
    \textit{Self-Paced Learning (SPL)} is a classical method where the curriculum is generated by the learner itself~\cite{kumar2010self}. 
    \textit{BabyLearning} is a recent method that simulates baby learning by starting with few training samples and fine-tuning using more weakly labeled videos crawled from the search engine \cite{liang2015towards}. 
    \textit{GoogleHNM} is a hard negative mining method proposed by Google \cite{varadarajan2015efficient}. It utilizes hard negative mining to train a second order mixture of experts model according to the video's YouTube topics. \textit{FastImage} \cite{han2015fast} is a video retrieval method that utilizes web images from search engine to match to the video with re-ranking. 
    \textit{WELL-MM} is the proposed method. The hyper-parameters of all methods including the baseline methods are tuned on the same validation set. On FCVID, the set is a standard development set with manual labels randomly selected from 10\% of the training set  (No training was done using ground truth labels) whereas on YFCC100M it is also a 10\% proportion of noisy training set.

\subsection{Experiments on FCVID}
    \textbf{Curriculum Comparison} As discussed in Section 3.2.2, we compare different ways to build curriculum for noisy label learning. 
    Here we also compare their effectiveness by training concept detectors directly using the curriculum labels. The batch train model is used for all generated curriculum labels. 
    In Table \ref{exp-cur} we show the batch trained models' precision at 5, 10 and mean average precision on the test set of FCVID. 
    For WELL-MM, we extract curriculum from different modalities as shown in Section 3.2.2, and combine them using linear weights. The weights are hyper-parameters that are tuned on the validation set, and the optimal weights for textual metadata, ASR, image classification and OCR results are 1.0, 0.5, 0.5 and 0.05, respectively. 
    This attempt to combining curriculum from different modalities serves as a pilot study. However, experiments show that such simple linear weighting is already effective with WELL-MM. Further research in this direction is left for future work.
    We also compare WELL-MM with using only latent topic modeling and word embedding soft matching. Results show that the curriculum generated by combining latent topic modeling and word embedding using multi-modal prior knowledge is the most accurate, which indicates our claim of exploiting multi-modal information is beneficial.

\begin{table}[]
\centering
\caption{Comparison of different curriculum using the BatchTrain learning method.}
\label{exp-cur}
\begin{tabular}{|l||c|c|c|}
\hline
Method            & P@5    & P@10   & mAP    \\ \hline \hline
WordHardMatching              & 0.782 & 0.763 & 0.469 \\ 
YouTubeTopicAPI &     0.587   &    0.563    &    0.315    \\ 
SearchEngine      &    0.723    &   0.713     & 0.413 \\ 
WordEmbedding          &     0.790   &     0.774   &     0.462   \\ 
LatentTopic          &   0.731    &     0.716   &    0.409   \\ 
\textbf{WELL-MM}               &   \textbf{0.838 }    &   \textbf{ 0.820 }   &     \textbf{ 0.486 } \\ \hline
\end{tabular}
\end{table}

    \textbf{Baseline Comparison} Table~\ref{exps-basline} compares the precision and mAP of different methods where the best results are highlighted. As we see, the proposed WELL-MM significantly outperforms all baseline methods, with statistically significant difference at $p$-level of 0.05.
    Comparing SPL with BatchTrain, it shows that the self-paced learning model over-fits to the noise without prior knowledge and performs worse than the simple BatchTrain model. Comparing WELL-MM with SPL and BatchTrain, the effect of incorporating multi-modal curriculum makes a significant difference in terms of performance, which suggests the importance of prior knowledge and preventing over-fitting in webly learning. The promising experimental results substantiate the efficacy of the proposed method.

\begin{table}[ht]
\centering

\caption{Baseline comparison on FCVID}
\label{exps-basline}

\begin{tabular}{|l||c|c|c|c|c|c|}
\hline
	   Method      & P@5 & P@10  & mAP \\ \hline \hline
	BatchTrain             & 0.838 & 0.820 & 0.486  \\
	FastImage~\cite{han2015fast} &- &- & 0.284\\
	SPL~\cite{kumar2011learning}           &  0.793 & 0.754  & 0.414    \\
    GoogleHNM~\cite{varadarajan2015efficient}      &  0.781 &  0.757 & 0.472  \\ 
	BabyLearning~\cite{liang2015towards}   & 0.834  & 0.817  & 0.496  \\ 
    \textbf{WELL-MM}      &\textbf{0.918}&\textbf{0.906} & \textbf{0.615}  \\ \hline
	\end{tabular}
\end{table}

  \textbf{Robustness to Noise Comparison}
  In this comparison we manually control the noise level of the curriculum in order to systematically verify how our methods would perform with respect to the noise level within the web data. To this end, we randomly select video samples with ground truth labels for each concept, so that the noise level of the curriculum labels are set at 20\%, 40\%, 60\%, 80\%  and we fix the recall of all the labels. We then train WELL-MM using such curriculum and test them on the FCVID testing set. We also compare WELL-MM to three other methods with the same curriculum, among them \textit{GoogleHNM} is a recent method to train video concept detector with large-scale data. We exclude \textit{BabyLearning}, which relies on the returned results by the search engine, since in this experiment the curriculum is fixed. As shown in Table \ref{exp-noise}, as the noise level of the curriculum grows, WELL-MM maintains its performance while other methods drop significantly. Specifically, when the noise level of curriculum increased from 60\% to 80\%, other methods' mAP drops 46.5\% on average while WELL-MM's mAP only drops 19.1\% relatively. It shows that WELL-MM is robust against different level of noise, which shows great potential in larger scale webly-labeled learning as the dataset gets bigger, the noisier it may become.

\begin{table}[ht]
\centering
\caption{WELL-MM performance with curriculum consisting of multiple artificial noise levels.}
\label{exp-noise}

\begin{tabular}{|l||c|c|c|c|}
\hline
   \backslashbox{Method}{Noise Level}   & 20\% & 40\% & 60\% & 80\% \\ \hline \hline
BatchTrain  &   0.592&    0.538&   0.463&     0.232               \\    
SPL           &    0.586&    0.515&     0.396&      0.184     \\ 
GoogleHNM             &    0.602&     0.552&   0.477& 0.304   \\ 
\textbf{WELL-MM}           &    \textbf{0.673}&    \textbf{0.646}&      \textbf{0.613}&    \textbf{0.496}   \\ \hline
\end{tabular}
\end{table}

    \textbf{Noisy Dataset Size Comparison} To investigate the potential of concept learning on webly-labeled video data, we apply the methods on different sizes of subsets of the data. Specifically, we randomly split the FCVID training set into several subsets of 200, 500, 1,000, and 2,000 hours of videos, and train the models on each subset without using manual annotations. The models are then tested on the same test set. Table~\ref{exps-small} lists the average results of each type of subsets. As we see, the accuracy of WELL-MM on webly-labeled data increases along with the growth of the size of noisy data while other webly learning methods' performance tend to be saturated.
    
    Comparing to the methods trained using ground truth, In Table~\ref{exps-small}, WELL-MM trained using the whole dataset (2000 hours) outperforms Static CNN (trained using manual labels) using around 1400 hours of data and rDNN-F (trained using manual labels with three features) trained using around 450h of data. And since the incremental performance increase of WELL-MM is close to linear, we conclude that with sufficient webly-labeled videos (which are not hard to obtain) WELL-MM will be able to outperform the rDNN-F trained using 2000h of data, which is currently the largest manual labeled dataset.

\begin{table}[ht]
\centering
\caption{MAP comparison of models trained using web labels and ground-truth labels on different subsets of FCVID. The methods marked by * are trained using human annotated labels.}
\label{exps-small}
\setlength{\tabcolsep}{4pt} 
\renewcommand{\arraystretch}{1.1} 
\begin{tabular}{|l||c|c|c|c|}
\hline
\backslashbox{Method}{Dataset Size}  & 200h & 500h & 1000h & 2000h \\ \hline
BatchTrain     & 0.364    & 0.422     & 0.452    & 0.486    \\ 
	SPL~\cite{kumar2011learning}           & 0.327 &0.379 &0.403 &0.414 \\
    GoogleHNM~\cite{varadarajan2015efficient}     & 0.361   & 0.421  &0.451 &0.472 \\ 
	BabyLearning~\cite{liang2015towards}   &0.390   & 0.447    & 0.481    & 0.496 \\ 
WELL-MM     &  0.487  &  0.554    & 0.595  & 0.615   \\ \hline
Static CNN\cite{jiang2015exploiting}*        &     0.485     &   0.561        &    0.604      & 0.638 \\ 
rDNN-F\cite{jiang2015exploiting}* &   0.550    &   0.620       &      0.650    &  0.754\\ \hline 
\end{tabular}
\end{table}

\subsection{Experiments on CIFAR-10}
    Following ~\cite{xiao2015learning}, we generate synthesized noisy training data with a noise level of 30\%, 40\% and 50\% on CIFAR-10 dataset. The models are trained on noisy data and tested on clean data. Classification Accuracy is reported. Our method doesn't assume any kind of noise distribution, while Noisy-CNN~\cite{sukhbaatar2014training} assumes the noise distribution depends on classes and Massive-Learning~\cite{xiao2015learning} assumes it also depends on the image content. 
    We show the experimental results in Table~\ref{image-exp}. The results show that 
    WELL-MM outperforms the other methods at all noise levels. More interestingly, as the noise level rises from 30\% to 50\%, the performance of Massive-Learning~\cite{xiao2015learning} drops about 9.8\%, while WELL-MM only drops 3.8\%. It shows that WELL-MM can also effectively learn robust concept detectors in image domain.

\begin{table}[]
\centering
\caption{Experimental results on CIFAR-10}
\label{image-exp}
\begin{tabular}{|l||c|c|c|}
\hline
\backslashbox{Methods}{Noise Level}        & 30\%    & 40\%    & 50\%    \\ \hline \hline
Noisy-CNN~\cite{sukhbaatar2014training}     & 0.697 & 0.667 & 0.634 \\ 
Massive-Learning~\cite{xiao2015learning} & 0.698 & 0.668 & 0.630 \\ 
\textbf{WELL-MM}          & \textbf{0.709} & \textbf{0.700} & \textbf{0.682} \\ \hline
\end{tabular}

\end{table}

\subsection{Experiments on YFCC100M}
In the experiments on YFCC100M, we train 101 concept detectors on YFCC100M and test them on the TRECVID MED dataset which includes 32,000 Internet videos. Since there are no manual labels, to evaluate the performance, we manually annotate the top 10 videos in the test set and report their precisions in Table~\ref{exps-yfcc}. The MED evaluation is done by four annotators and the final results are averaged from all annotations. The Fleiss' Kappa value for these four annotators is 0.64.
A similar pattern can be observed where the comparisons substantiate the rationality of the proposed webly learning framework. Besides, the promising results on the largest multimedia set YFCC100M verify the scalability of the proposed method.

\begin{table}[ht]
\centering
\caption{Baseline comparison on YFCC100M}
\label{exps-yfcc}

\begin{tabular}{|l||c|c|c|c|c|c|}
\hline
Method      & P@3 & P@5   & P@10\\ \hline \hline
	BatchTrain              & 0.535  &  0.513 &  0.487 \\
	SPL~\cite{kumar2011learning}              &0.485  &   0.463 &  0.454\\
    GoogleHNM~\cite{varadarajan2015efficient}       &0.541  &  0.525 &  0.500 \\ 
	BabyLearning~\cite{liang2015towards}     &0.548  & 0.519 &  0.466 \\ 
	\textbf{WELL-MM} &\textbf{0.667}    &   \textbf{0.663}   & \textbf{0.649}\\ \hline
	\end{tabular}
\end{table}

\begin{figure}
	\centering
		\includegraphics[width=0.9\textwidth]{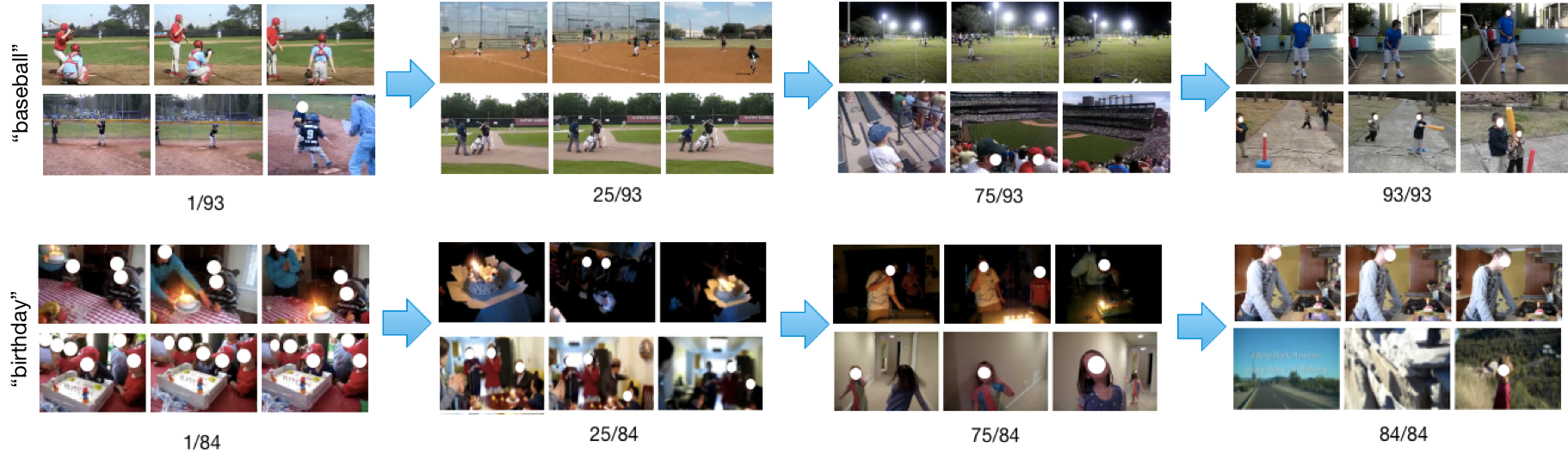}
	\caption{Illustration of representative videos selected by WELL-MM at different iterations}
	\label{fig:well_qual}
\end{figure}

\subsection{Qualitative Analysis}
    In this section we show training examples of WELL-MM. In Fig.~\ref{fig:well_qual}, we demonstrate the positive samples that WELL select at different stage of training the concept "baseball" and "birthday". 
    For the concept "baseball", at the early stage (1/93, 25/93), WELL-MM selects easier and clearer samples such as the ones with camera directly pointing at the playground, while at the later stage (75/93, 93/93) WELL-MM starts to train with harder samples with different lighting conditions and untypical samples for the concept. 
    For the concept "birthday", as we see, at later stage of the training, complex samples for birthday event like a video with two girl singing birthday song (75/84) and a video of celebrating birthday during hiking (84/84) are included in the training, while at the early stage, only typical "birthday" videos with birthday cake and candles are included.
    
\section{Related Work}
\textbf{Curriculum and Self-paced Learning}: Recently a learning paradigm called~\textit{curriculum learning} (CL) was proposed by Bengio et al., in which a model is learned by gradually incorporating from easy to complex samples in training so as to increase the entropy of training samples~\cite{bengio2009curriculum}. A curriculum determines a sequence of training samples and is often derived by predetermined heuristics in particular problems. 
For example, Chen et al. designed a curriculum where images with clean backgrounds are learned before the images with noisy backgrounds~\cite{chen2015webly}, i.e. their method first builds a feature representation by a Convolutional Neural Network (CNN) on images with clean background and then they fine tune the models on images with noisy background. 
In~\cite{spitkovsky2009baby}, the authors approached grammar induction, where the curriculum is derived in terms of the length of a sentence. Because the number of possible solutions grows exponentially with the length of the sentence, and short sentences are easier and thus should be learned earlier.

The heuristic knowledge in a problem often proves to be useful. However, the curriculum design may lead to inconsistency between the fixed curriculum and the dynamically learned models. That is, the curriculum is predetermined a prior and cannot be adjusted accordingly, taking into account the feedback about the learner. To alleviate the issue of CL, Kumar et al. designed a learning paradigm, called \emph{self-paced learning} (SPL)~\cite{kumar2010self}. SPL embeds curriculum design as a regularizer into the learning objective. Compared with CL, SPL exhibits two advantages: first, it jointly optimizes the learning objective with the curriculum, and thus the curriculum and the learned model are consistent under the same optimization problem; second, the learning is controlled by a regularizer which is independent of the loss function in specific problems. This theory has been successfully applied to various applications, such as action/event detection~\cite{jiang2014self}, domain adaption~\cite{tang2012shifting}, tracking~\cite{supancic2013self} and segmentation~\cite{kumar2011learning}, reranking~\cite{jiang2014easy}, etc.

\textbf{Learning Detectors in Web Data}: Many recent studies have been proposed to utilize a large amount of noisy data from the Internet. For example, \cite{mitchell2015never} proposed a Never-Ending Language Learning (NELL) paradigm and built adaptive learners that make use of the web data by learning different types of knowledge and beliefs continuously. 
In the image domain, existing methods try to tackle the problem of constructing qualified training sets based on the search results of text or image search engines~\cite{chen2013neil,divvala2014learning,liang2015towards,xiao2015learning}. 
For example,
NEIL~\cite{chen2013neil} followed the idea of NELL and learned from web images to form a large collection of concept detectors iteratively via a semi-supervised fashion. By combining the classifiers and the inter-concept relationships it learned, NEIL can be used for scene classification and object detection task.
\cite{liang2015towards} presented a weakly-supervised method called Baby Learning for object detection from a few training images and videos. They first embed the prior knowledge into a pre-trained CNN. When given very few samples for a new concept, a simple detector is constructed to discover much more training instances from the online weakly labeled videos. As more training samples are selected, the concept detector keeps refining until a mature detector is formed. 
Another recent work in image domain \cite{chen2015webly} proposed a webly supervised learning of Convolutional Neural Network. They utilized easy images from search engine like Google to bootstrap a first-stage network and then used noisier images from photo-sharing websites like Flickr to train an enhanced model.  

In video domain, only a few studies \cite{duan2012visual,han2015fast,varadarajan2015efficient} have been proposed for noisy data learning since training robust video concept detectors is more challenging than the problem in the image domain.
\cite{duan2012visual} tackled visual event detection problem by using SVM based domain adaptation method in web video data.
\cite{han2015fast} described a fast automatic video retrieval method using web images. Given a targeted concept, compact representations of web images obtained from search engines like Google, Flickr are calculated and matched to compact features of videos. Such method can be utilized without any pre-defined concepts.
\cite{varadarajan2015efficient} discussed a method that exploits the YouTube API to train large-scale video concept detectors on YouTube. The method utilized a calibration process and hard negative mining to train a second order mixture of experts model in order to discover correlations within the labels.

\section{Summary}
In this chapter, we propose a novel method called WELL-MM for webly labeled video data learning. WELL-MM extracts multi-modal informative knowledge from noisy weakly labeled video data from the web through a general framework and achieves the best performance only using webly-labeled data on two major video datasets. The comprehensive experimental results demonstrate that WELL-MM outperforms state-of-the-art studies by a statically significant margin on learning concepts from noisy web video data. 
In addition, the results also verify that WELL-MM is robust to the level of noisiness in the video data. The result suggests that with more webly-labeled data, which is not hard to obtain, WELL-MM can potentially outperform models trained on any existing manually-labeled data.

\chapter{Spatial-Temporal Alignment Network for Action Recognition and Detection
}  \label{chap:0103_viewpoint}

\newcommand{\fancyname}{\textit{STAN}}

In this chapter, we explore viewpoint-invariant feature representations that aim to have better generalization abilities for action recognition and detection. We propose a novel Spatial-Temporal Alignment Network (\fancyname) that aims to learn geometric invariant representations for action recognition and action detection.

\section{Motivation}
Human vision can recognize video actions efficiently despite the variations of viewpoints. 

Convolutional neural networks (CNNs) \cite{cnn-lecun-eccv10,cnn-dtran-iccv15,cnn-Carreira-Zisserman-cvpr17,DBLP:journals/corr/SigurdssonDFG16,DBLP:conf/nips/FeichtenhoferPW16} 
fully utilize the power of GPUs/TPUs and employ spatial-temporal filters to recognize actions, which
outperforms traditional models including oriented filtering in space time (HOG3D) \cite{DBLP:conf/bmvc/KlaserMS08}, spatial-temporal interest points \cite{DBLP:conf/cvpr/LaptevMSR08}, motion history images \cite{DBLP:journals/tsmc/TianCLZ12}, and trajectories \cite{DBLP:conf/iccv/WangS13a}.
However, due to the high variations in space-time, the state of the art of action recognition is still far from satisfactory, compared with the success of 2D CNNs in image recognition \cite{he2016deep}.

A key challenge of action recognition is to capture the variations across space and time.
Since CNN assumes the filters share weights at different locations, it can not explicitly model the viewpoint changes and other variations. To solve such limitations, a practical way is to expand feature representations to accommodate a higher degree of freedom. 
For example, the two-stream network \cite{twostream-Simonyan-nips14} proposes to integrate optical flow with RGB features. More recently, SlowFast
\cite{feichtenhofer2019slowfast} combines both slow and fast pathways to learn different temporal information, and obtain good performance. However, such feature expanding approaches quickly lead to
cumbersome, high-dimensional feature maps, which not only make the computation more expensive but also miss the geometric interpretation of the subjects. 
In this chapter we show that our method can improve on both simpler backbone networks like ResNet3D and more complex ones like SlowFast, with minimal computation overhead.

\begin{figure}[t!]
	\centering
		\includegraphics[width=0.8\textwidth]{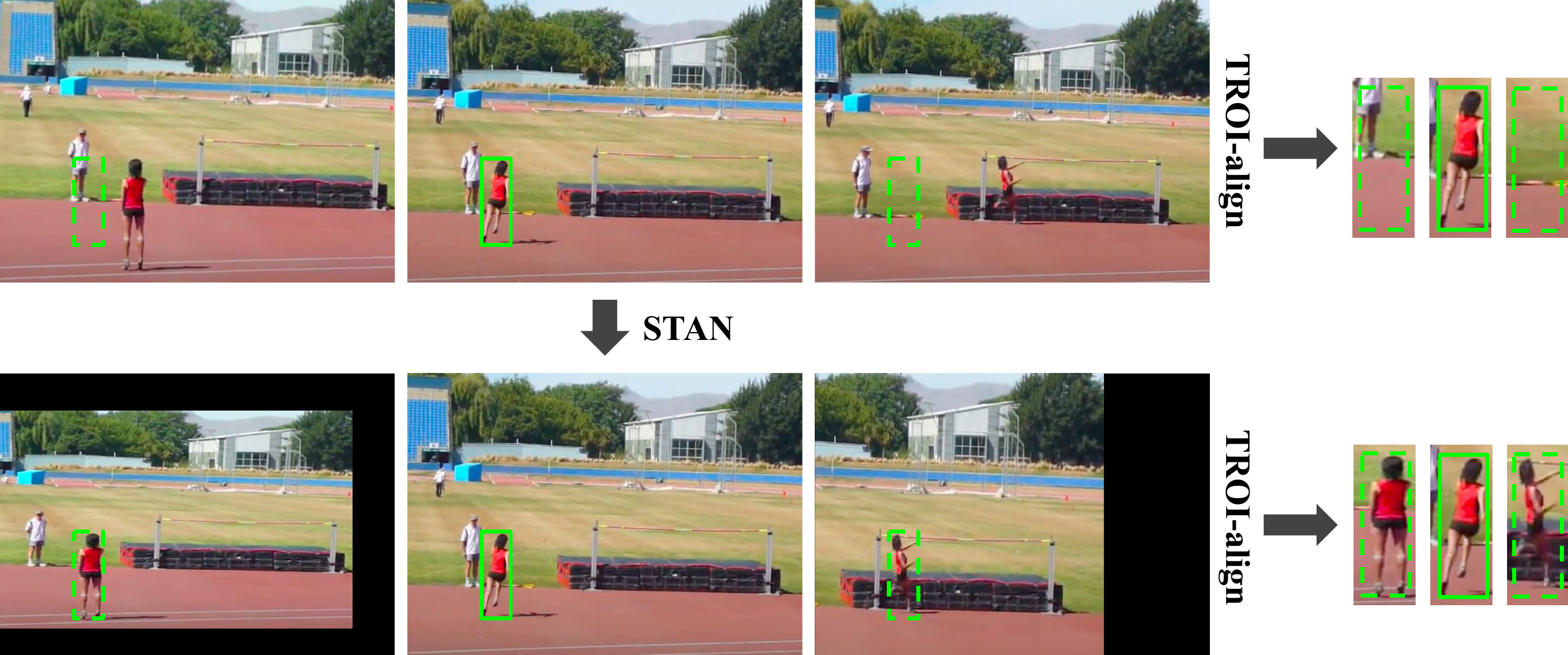} 
	\caption{An illustrative example of how {\fancyname} can address the bounding box misalignment issue in spatial temporal localization. In prior work~\cite{gu2018ava,feichtenhofer2019slowfast}, the detected person box (solid line) is expanded along the temporal axis (dotted line) to form a 3D cube for the subsequent temporal ROI-align (TROI-align). If the actor's movement is large across frames as shown in the top row (or there are substantial camera movements), the TROI-aligned features will be noisy with the majority being background pixels. {\fancyname} learns a spatial temporal transformation 
	that puts feature maps in the same coordinate system and in the bottom row example it is the central frame.}
	\label{fig:title}
\end{figure}

We proposes a different approach to capture the variation in action recognition. 
Instead of relying on stacking deeper CNN layers, we aims to explicitly learn geometric transformations and viewpoint invariant features. 
Our idea is motivated by \cite{kosiorek2019stacked}, which believes that  human vision relies on coordinate frames. 
However, the stacked capsule autoencoder in \cite{kosiorek2019stacked} is designed for 2D images, and too expensive for large scale visual recognition.

Fig.~\ref{fig:title} illustrates the importance of alignment in the problem of spatial temporal localization. Following the existing action detection pipeline~\cite{gu2018ava,feichtenhofer2019slowfast}, the action representation is obtained by first cropping a 3D cube within the spatial-temporal feature maps centered at the detected actor’s bounding box, followed by a Max Pooling operation. Without alignment, the representation is contaminated by background pixels as shown in the top row in Fig.~\ref{fig:title}. With alignment the representation as shown in the bottom row is much more focused on the actor.

In this chapter,
we propose a Spatial Temporal Alignment Network ({\fancyname}) that aims to learn viewpoint invariant representations for action recognition and action detection. 
The {\fancyname} block is very light-weighted and generic, which could be plugged into existing action recognition models like ResNet3D, Non-Local Network~\cite{wang2018non} and the SlowFast network~\cite{feichtenhofer2019slowfast}. 
As discussed in Section~\ref{sec:arch}, we insert a {\fancyname} block between $res_2$ and $res_3$ in the ResNet3D architecture, 
and add it at the same location on the Fast pathway in the SlowFast model.
For the SlowFast + {\fancyname} model, the FLOPS increase is only \textbf{relatively 2.1\%} (134.5 GFLOPS vs. 131.7 GFLOPS) for action recognition on Charades, but we achieve \textbf{5\% relative} improvement on mean average precision.

The contribution of this chapter is three-fold: 
(1) To the best of our knowledge, this is the first work to explore explicit spatial-temporal alignment in 3D CNNs for action detection.
(2) Our {\fancyname} requires very low extra FLOPS in addition to the backbone network. (3) Extensive experiments on different datasets suggest the model using {\fancyname} can outperform the state of the art.

\section{The {\fancyname} Model}
\label{sec:method}

In this section, we describe our spatial-temporal alignment network for action recognition and detection, which we call {\fancyname}.
Motivated by the previous  works in image understanding \cite{DBLP:conf/nips/HuangMLL12,jaderberg2015spatial,lin2017inverse}, our work considers a generalized model in video domain, so that it can handle dynamic viewpoint changes in action recognition and detection.
The key idea of {\fancyname} is to utilize spatial-temporal alignment for feature maps to account for viewpoint changes and actor movements in the videos.
The learned affine transformations (Eqn.~\ref{eqn:affine}) at
each temporal group can achieve effects including
camera stabilization (See Fig.~\ref{fig:overview}), which is important in Internet videos to counter camera motions.
In general, given an input spatial-temporal feature map $\mathcal{I}_{in} \in \mathbf{R}^{C \times T \times H \times W}$ where $H$ stands for height, $W$ for width, $T$ for time and $C$ for channels, the alignment function is defined as
\begin{align}
\label{eqn:stan}
    \text{STAN}(\mathcal{I}_{in}) = \mathcal{T}(\theta, \mathcal{I}_{in}) + \mathcal{I}_{in},
    \text{where} \; \theta = \mathcal{D}(\mathcal{I}_{in})
\end{align}

The output of {\fancyname} function is of the same dimension as $\mathcal{I}_{in}$. 
Function $ \mathcal{D}$ represents the deformation network, where the feature map alignment parameter $\theta \in \mathbf{R}^{4 \times 4}$ is computed based on the input feature map.
Function $ \mathcal{D}$ can be in the form of a simple feed-forward network using spatial-temporal features \cite{tran2018closer}.
Function $ \mathcal{T}$ is defined as the warping function, where input feature maps are warped based on the alignment parameter.
We add a residual connection between the input feature maps and output feature maps for faster training and avoiding boundary effect described in ~\cite{lin2017inverse}.
The {\fancyname} layer can be added to different locations of the backbone to account for the alignment needs for different level of feature maps.

\subsection{Network Architecture}
\noindent The overall {\fancyname} architecture is shown in Fig.~\ref{fig:overview}.
Our model uses a 3D CNN as the backbone network to extract spatial-temporal feature maps from video frames. In addition,  {\fancyname}
has the following key components:


\noindent\textbf{Deformation Network} produces the alignment / deformation parameter $\theta$ in Eqn.~\ref{eqn:stan}. 

\noindent\textbf{Warping Module} samples from the input feature maps based on $\theta$ and outputs the final transformed feature maps.


\noindent\textbf{Temporal Grouping} learns a separate alignment for different temporal segments within the video clip, similar to multihead attention~\cite{vaswani2017attention} and correlation networks~\cite{wang2020video}.

In the following, we will introduce the above modules in details.

\subsection{Deformation Network}\label{sec:deformnet}
The deformation network $\mathcal{D}$ produces a transformation parameter, $\theta \in \mathbf{R}^{4 \times 4}$.
Our network is based on R(2+1)D~\cite{tran2018closer} although other options are possible, such as a simple feed-forward network, or a recurrent network and compositional function as proposed in ~\cite{lin2017inverse}.
Suppose the number of input channels is $C$, the details of our network architecture is presented in Table~\ref{tab:deformnet}.  
\begin{table}
\centering
\begin{tabular}{c | c | c }
Layer  & Filter size &  Output channels \\
\hline
conv & $1\times7^2$ & $C // 4$ \\
\hline
conv & $7\times1^2$ & $C // 2$\\
\hline
\multirow{2}{*}{maxpool} & $4\times7^2$ & \multirow{2}{*}{$C // 2$} \\
& stride $4\times7^2$ &  \\
\hline
conv & $7\times7^2$ & $C$ \\
\hline
globalpool &  & $C$ \\
\hline
fc &  & variable \\
\end{tabular}
\vspace{-2mm}
\caption{Deformation network architecture used in our experiments. The filter sizes are $T\times S^2$ where $T$ is the temporal kernel size and $S$ is the spatial size. The global pool layer averages features across all spatial-temporal locations. ``fc'' is the final fully-connected layer.} 
\label{tab:deformnet}
\vspace{-9mm}
\end{table}
All convolution layers are followed by batch normalization~\cite{ioffe2015batch} and ReLU~\cite{hahnloser2000digital,glorot2011deep}. The dimension of the final FC layer depends on the type of parameterization we choose for the spatial-temporal alignment.
Taking affine transformation for example, the dimension of the network output ($\mathbf{p}_{\text{affine}}=[p_1 \; ... \;p_{12}]^{T}$) is of size 12 and $\theta$ is constructed as 
\begin{align}\label{eqn:affine}
    \theta(\mathbf{p}_{\text{affine}}) = 
\begin{pmatrix}
1 + p_1 & p_2 & p_3 & p_4\\
p_5 & 1 + p_6 & p_7 & p_8\\
p_9 & p_{10} & 1 + p_{11} & p_{12}\\
0 & 0 & 0 & 1\\
\end{pmatrix}
\end{align}
$\theta$ can be more restrictive as in the case of attention~\cite{xu2015show}, where cropping, translation and scaling are allowed for transformation, the dimension of the network output $\mathbf{p}_{\text{att}}=[p_1,...p_6]^T$ is a vector of size 6 and $\theta$ is constructed as 
\begin{align}\label{eqn:att}
    \theta(\mathbf{p}_{\text{att}}) = 
\begin{pmatrix}
1 + p_1 & 0 & 0 & p_4\\
0 & 1 + p_2 & 0 & p_5\\
0 & 0 & 1 + p_3 & p_6\\
0 & 0 & 0 & 1\\
\end{pmatrix}
\end{align}
The design of {\fancyname} is flexible and can be any type of transformation.
The key intuition of the deformation on CNN feature maps is to compensate for the fact that CNNs are not rotation, scale, and shear transformation equivariant~\cite{kosiorek2019stacked}.

\subsection{Warping Module}
\begin{figure}[ht]
	\centering
		\includegraphics[width=0.8\textwidth]{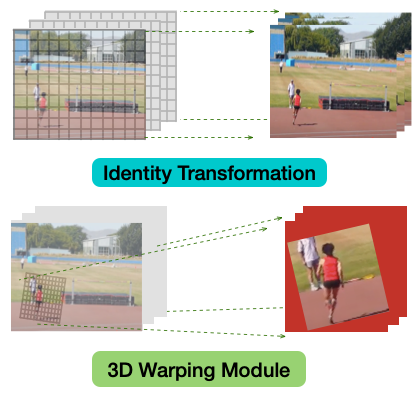} 
	\caption{Two examples of applying spatial-temporal warping/transform to an input sequence. The top row shows an identity transformation while the bottom row shows an example 3D warping where the actor is rotated and zoomed in.
	}
	\label{fig:warp}
\end{figure}

After computing the transformation matrix $\theta$, we utilize a differentiable warping function $\mathcal{T}$ to transform the feature maps with better alignment of the content.
See Fig.~\ref{fig:warp}.
The warping function is essentially a resampling of features from the input feature maps to the output at each corresponding pixel location.
Note that the feature maps could also be images.
Extending from the notation of 2D alignment~\cite{jaderberg2015spatial} , we define the output feature maps $\mathcal{I} \in \mathbf{R}^{C \times T \times H \times W}$ to lie on a spatial-temporal regular grid $G = \{G_i\}$, where each element of the grid $G_i = (t^o_i, x^o_i, y^o_i)$ corresponds to a vector of output features of size $\mathbf{R}^{C}$.  
Hence, the pointwise sampling between the input and output feature maps is written as
\begin{align}
    \begin{bmatrix}
    t^s_i \\
    x^s_i \\
    y^s_i \\
    1 \\
    \end{bmatrix}
    = \mathcal{T}_{g}(\theta, G_i) =
    \begin{pmatrix}
    1\!+\!p_1 & 0 & 0 & p_4\\
    0 & 1\!+\!p_2 & 0 & p_5\\
    0 & 0 & 1\!+\!p_3 & p_6\\
    0 & 0 & 0 & 1\\
    \end{pmatrix}
    \begin{bmatrix}
    t^o_i \\
    x^o_i \\
    y^o_i \\
    1 \\
    \end{bmatrix}
\end{align}
where ($t^o_i, x^o_i, y^o_i$) are the output feature map coordinates in the regular grid and ($t^s_i, x^s_i, y^s_i$) are the corresponding input feature map coordinates for feature sampling.
Here we use attention transformation as an example, where the deformation matrix is $\theta$ parameterized by  $\mathbf{p}_{\text{att}}$ (Eqn.~\ref{eqn:att}).
Given the coordinates mapping, as the computed corresponding coordinates in the input feature maps might not be integers, we utilize the differentiable trilinear interpolation to sample input features from the eight closest points based on their distance to the computed point ($t^s_i, x^s_i, y^s_i$).
In this way, we iterate through every point in the regular grid, ($t^o_i \in [1, ..., T], x^o_i \in [1, ..., W], y^o_i \in [1, ..., H]$) and compute the output feature maps identically for each channel.

\subsection{Temporal Grouping}\label{sec:tg}
\begin{figure}[ht]
	\centering
		\includegraphics[width=0.9\textwidth]{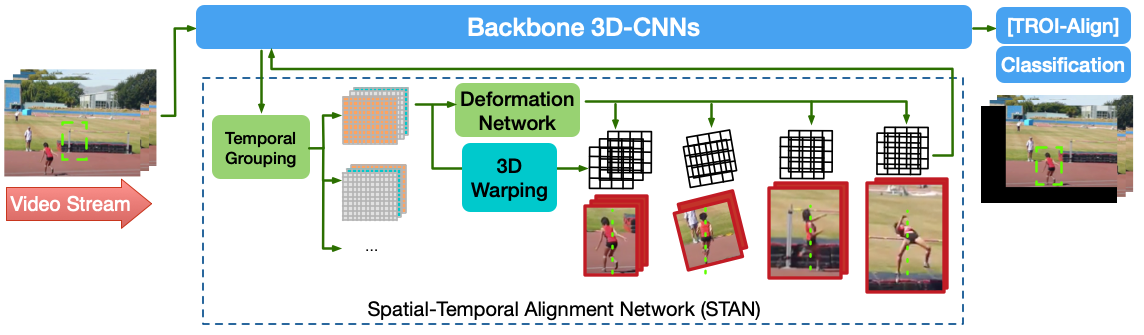} 
	\caption{Spatial Temporal Alignment Network ({\fancyname}) design. In our design, different temporal slices of the feature map can undergo different transformations to account for intra-clip camera motion and actor movement. 
	We show four alignment examples with a reference line where the actor is aligned on for demonstration. After the features are transformed and processed through the backbone network, they are passed to a temporal ROI-Align layer (TROI-Align) if the task is action detection, and then finally to the classification layer.
	}
	\label{fig:overview}
\end{figure}

Many actions are composed of a sequence of sub-actions that require different alignments. For example, action ``High Jump'' may consist of ``stand'', ``run'' and ``jump'' (see Fig.~\ref{fig:overview}), and the actor may have moved around between video frames.
Based on this observation, we propose temporal grouping to allow the model to learn different alignments at different time periods.
It is written as 
\begin{align}
\label{eqn:tg}
    \text{STAN}_{tg}(\mathcal{I}_{in}) = \text{concat}\left(\text{STAN}(\mathcal{I}^1_{in}),..., \text{STAN}(\mathcal{I}^{T_g}_{in}) \right)
\end{align}
where ${T_g}$ is the number of temporal groups and concatenation is operated on the temporal axis.
Each group alignment $\text{STAN}(\mathcal{I}^{T_g}_{in})$ is of size $\mathbf{R}^{C \times T // T_g \times H \times W}$.

\subsection{Integration with Backbone CNNs}\label{sec:arch}
Now that we have defined {\fancyname} function we now discuss effective ways to add it into existing backbone CNN networks. In this chapter, we focus on designing {\fancyname} for the RGB stream and the extension to optical flow will be left as one of our future work.
Recent works~\cite{wang2018non,feichtenhofer2019slowfast} and their variants~\cite{pan2020actor} achieve single-stream state-of-the-art performance on action recognition and detection with 3D CNNs. 
We therefore explore adding {\fancyname} layers to the ResNet3D model~\cite{wang2018non,feichtenhofer2019slowfast} and the SlowFast~\cite{feichtenhofer2019slowfast} network.
Intuitively, placing {\fancyname} layer at shallower layers allows earlier feature alignments that could potentially lead to cleaner representation for better action recognition. However, shallow layers may not have enough abstraction in the feature maps for {\fancyname} to learn the right alignments.
Based on this trade-off, we experiment with adding one {\fancyname} layer after $res_2$ block. 
For SlowFast, we only add {\fancyname} layers on the Fast pathway since it contains more temporal information compared to the Slow pathway.
Other insertion locations are explored in the ablation experiments (Section~\ref{sec:ablation}).

\subsection{Detection Architecture}\label{sec:detect}
For action recognition, the final CNN outputs are passed through global averaging (and a concatenation for SlowFast) and a fully-connected layer to get the action class probabilities.
For action detection, following previous works~\cite{gu2018ava,sun2018actor,feichtenhofer2019slowfast}, we use pre-computed actor proposals.
The bounding boxes are used to extract region-of-interest (RoI)
features using RoI-Align~\cite{he2017mask} at the last feature map of ``$res_5$'' after temporal global average pooling. 
Our {\fancyname} can fix the actor misalignment problem by aligning the 2D RoIs along the temporal axis. 
Finally, the RoI features are then max-pooled and fed to a fully-connected layer.

\section{Experiments}
\label{sec:exp}
To demonstrate the efficacy of our models, specifically, the viewpoint-invariant design that helps action models to generalize, we experiment on two recent action detection datasets, including AVA~\cite{gu2018ava} and AVA-Kinetics~\cite{li2020ava}, and several major action recognition datasets, Kinetics-400~\cite{kay2017kinetics}, Charades~\cite{sigurdsson2016hollywood}, Charades-Ego~\cite{sigurdsson2018charades},
MEVA~\cite{corona2021meva}.
In the experiments, we aim to showcase
that STAN can achieve significant performance improvement with minimal computation overhead.

\subsection{Action Detection}
\label{sec:exp_detect}
This section evaluates our {\fancyname} model for the task of spatial temporal localization for actions using AVA and AVA-Kinetics.
We show significant improvement over baselines while only adding a fraction of computation. 

\noindent\textbf{Datasets.}
The AVA dataset~\cite{gu2018ava} is an action detection dataset where models are required to output action classification results with bounding boxes. 
Spatial-temporal labels are provided for one frame per second, with every person annotated with a bounding box and (possibly multiple) actions.
There are 211k training and 57k validation video clips. 
We use AVA v2.1 and follow the standard protocol~\cite{gu2018ava,feichtenhofer2019slowfast} of evaluating on 60 classes.
The performance metric is mean Average Precision (mAP) over 60 classes, using an IoU threshold of 0.5.
We report mAP on the official validation set.

The AVA-Kinetics dataset~\cite{li2020ava} is a recent action detection dataset which follows AVA annotation protocol to annotate relevant videos on the Kinetics-700~\cite{carreira2019short} dataset.
We utilize the part where all videos are from Kinetics for training and testing.
There are about 141k training videos and 32k validation videos.
The AVA-Kinetics also contains the same 60 classes for evaluation and we utilize the mAP metric. Following~\cite{li2020ava}, we evaluate action detection model performance with both ground truth person boxes and pre-computed person boxes.
We report mAP on the official validation set as well.

\noindent\textbf{Action Proposals.}
The action detection models have to output action/person bounding boxes, and only predicted boxes with IoU (intersection over union) area w.r.t the ground truth boxes above a threshold of 0.5 are considered true positives.
As mentioned in Section~\ref{sec:detect}, we follow previous works~\cite{gu2018ava,sun2018actor,feichtenhofer2019slowfast} and use pre-computed person boxes as action proposals.
For AVA, we utilize the same person proposals from ~\cite{feichtenhofer2019slowfast}.
For AVA-Kinetics, we finetune a Mask-RCNN~\cite{he2017mask} with ResNet-101 backbone trained on COCO~\cite{lin2014microsoft} on the person boxes in the training set and extract boxes as described in Section~\ref{sec:detect}.
The region proposals for action detection are detected person boxes with a confidence score of larger than 0.8, which has a recall of 83.1\% and a precision of 62.1\% for the person class, given IoU threshold of 0.5.
The average precision of the person class is 0.732 on the validation set.

\noindent\textbf{Training.}
We initialize the network weights from the Kinetics-400 classification models, following previous works~\cite{feichtenhofer2019slowfast}. 
We use a learning rate of 0.1 and cosine learning rate decay. We use synchronized SGD to train for 10 epochs with a batch size of 16 on a 4-GPU machine, with a linear warm-up from 0.000125 for the first 2 epochs.
We use a weight decay of $10^{-7}$. 
Ground-truth boxes and video clips centered at the annotated key frames are sampled for training. The video is first resized to 256x320, and then 
we use random 224×224 crops and horizontal flipping  following~\cite{feichtenhofer2019slowfast}.

\noindent\textbf{Inference.}
Since the annotations on AVA and AVA-Kinetics are one (key) frame per second, we sample a single video clip temporally centered around the key frame for evaluation.
Following~\cite{feichtenhofer2019slowfast}, we resize the spatial dimension such that its shorter side is 256 pixels. 
Ground truth boxes or pre-computed boxes are used as inputs.
We report the inference time computational cost (in FLOPs) of a single 256x320 clip using Tensorflow's Profiler. 

\noindent\textbf{Implementation Details.}
We add one {\fancyname} layer to the backbone CNN network as described in Section~\ref{sec:arch}.
We use affine transformation (Eqn.~\ref{eqn:affine}) and the deformation network is defined in Table~\ref{tab:deformnet}.
We use a temporal group of 2 and the number of base convolution filters in the deformation network is capped at 8 for ResNet3D to keep the FLOPs low. For SlowFast, we set this number as described in Table~\ref{tab:deformnet}.

\noindent\textbf{Baselines.}
To demonstrate the effectiveness of our proposed model, we experiment with recent 3D-CNN based models for action recognition and detection.
\textbf{ResNet3D} is a model based on ResNet-50~\cite{he2016deep} with additional 3D convolutional filters. 

The number of input frames is 8 and we sample 1 frame every 8 frames (i.e., 8x8 frames).
\textbf{ResNet3D + {\fancyname}} is our proposed model added to ResNet3D as described in Section~\ref{sec:arch} with the same inputs.
\textbf{SlowFast} is a recent efficient model~\cite{feichtenhofer2019slowfast} with a Slow pathway and a Fast pathway, which takes 8x8 and 32x2 frames as inputs respectively. We use ResNet-50 backbone for SlowFast as well.
If we use only the Slow pathway, the model will become the same as ResNet3D. 
\textbf{SlowFast + {\fancyname}} is our proposed model added to SlowFast as described in Section~\ref{sec:arch}.

\begin{table}[t!]
\centering
\begin{tabular}{l|c|c|c}
\hline
Models                & mAP  & GFLOPs & MParams  \\ \hline
ResNet3D   (8x8)            & 0.234 &   208.0 & 31.75      \\ 
ResNet3D + {\fancyname} & \textbf{0.247} & 216.6 & 32.02  \\ \hline
SlowFast~\cite{feichtenhofer2019slowfast}    (32x2)              & 0.252&     242.6 & 33.77     \\ 
SlowFast + {\fancyname} & \textbf{0.268} & 247.4 & 33.96 \\ \hline
\end{tabular}
\caption{Experiment results on AVA dataset. We show the model performance on mAP, computation cost (in billions) and number of parameters (in millions). The models are based on ResNet-50 backbone. The computational cost is of a single 256x320 video clip of length 8x8 or 32x2 (number of frames x sampling rate) frames.
}
\label{tab:ava}
\end{table}

\begin{table}[t!]
\centering
\begin{tabular}{l|c|c}
\hline
Models                & mAP & GFLOPs \\ \hline
Action Transformer~\cite{li2020ava}    & 0.337 / 0.191  & -                \\ \hline
ResNet3D  (8x8)            & 0.315 / 0.224      & 208.0                 \\ 
ResNet3D + {\fancyname} & \textbf{0.336} / \textbf{0.238}  & 216.6      \\ \hline
SlowFast~\cite{feichtenhofer2019slowfast}         (32x2)           & 0.341 / 0.242 &     242.6                \\ 
SlowFast + {\fancyname} & \textbf{0.358} / \textbf{0.253}  & 247.4  \\ \hline
\end{tabular}
\caption{Experiment results on AVA-Kinetics dataset. We show both mAP with ground truth boxes and detected boxes. We are not able to get Action Transformer's FLOPs as the code is not public. Note that the pre-computed boxes for Action Transformer is different from ours.}
\label{tab:ava-kinetics}
\end{table}

\subsubsection{Main Results}
\noindent\textbf{AVA.}
Table~\ref{tab:ava} shows the results on AVA dataset~\cite{gu2018ava}.
We follow the SlowFast~\cite{feichtenhofer2019slowfast} paper's evaluation protocol and use the same predicted person bounding boxes provided by the authors with ROIAlign to classify actions.
For both ResNet3D and SlowFast, we use ResNet-50 as their base architecture.
Compared with baseline methods, our model is able to achieve significant improvement with minimal computation overhead (about 2-4\% more). The parameter increase is also minimal.

\noindent\textbf{AVA-Kinetics} is a recent action detection dataset with Internet videos from the Kinetics-700 dataset~\cite{carreira2019short}, which has more moving cameras and objects with smaller bounding boxes than AVA.
Table~\ref{tab:ava-kinetics} shows the results on AVA-Kinetics dataset~\cite{li2020ava}.
We follow the baseline~\cite{li2020ava} paper's evaluation protocol and experiment with both ground truth person boxes and detected person boxes from the described finetuned Mask-RCNN model.
Our model is able to achieve 2.1\% absolute improvement on mAP for ResNet3D with only relatively 4.1\% more computation and 1.7\% improvement with 2\% more computation on SlowFast.
These results show
the consistently significant improvement of STAN.

\subsubsection{Ablation Experiments}\label{sec:ablation}

\begin{table}[t!]
\centering
\begin{tabular}{l|c|c|c}
\hline
               & Diff & mAP & GFLOPs   \\ \hline
SlowFast             & - & 0.252 &   242.55     \\ 
+ {\fancyname} & \textbf{+1.6\%} & \textbf{0.268} & 247.40 \\ \hline
+ {\fancyname} ($pool_1$) & - & 0.253 & 242.88 \\
+ {\fancyname} ($res_3$) & +0.7\% & 0.259 & 247.35 \\
+ {\fancyname} ($res_4$) & +0.3\% & 0.255 & 247.39 \\ \hline
+ {\fancyname} (Att, 6) & +0.7\% & 0.259 & 247.40  \\
+ {\fancyname} (H, 15) &  +0.3\% & 0.255 & 247.40  \\ \hline
+ {\fancyname} (no tg) &  +0.8\% & 0.260 & 247.40  \\ 
+ {\fancyname} (tg=\#frames) &  - & 0.254 &  246.16 \\ \hline
+ {\fancyname} (fixed $W_{\theta}$) &  +1.2\% & 0.264 & 247.40 \\\hline
\end{tabular}
\caption{Ablation experiment results (on AVA). The computational cost is of a single 256x320 video clip of length 32x2 (number of frames x sampling rate) frames. The ``Diff'' column shows absolute improvement of the variant models compared to the baseline SlowFast model. ``(Att, 6)'' means attention transformation with 6 degree-of-freedom (DoF),
and ``(H, 15)'' means homography transformation with 15 DoF. ``tg'' means temporal grouping. See text for details.}
\label{tab:ablation}
\end{table}

In this section, we perform ablation studies on the AVA dataset with the SlowFast model as the backbone network.
We run each ablation
experiments two times and show the averaged results. The
differences between the same run are within 0.1\%.
To understand how action models can benefit from {\fancyname}, we explore the following questions (results are shown in Table~\ref{tab:ablation}):

\noindent\textbf{Where to insert {\fancyname} layer?}
In CNN networks, shallower layers tend to encode low-level visual features like edges and patterns while deeper layers may contain more abstract information.
Placing {\fancyname} layer at shallower layers allows earlier feature alignments that could potentially lead to cleaner representation. 
To verify this hypothesis, we experiment with adding one {\fancyname} layer at deeper layers than $res_2$ block and before. 
We first try adding {\fancyname} layer after $res_3$ and $res_4$.
As we see, the model performance deteriorates significantly.
We then try adding STAN layer after $pool_1$ (before $res_2$).
There is virtually no improvement and the addition of FLOPS is low due to the low number of features channels at that location. We suspect
that STAN at earlier layer struggles to understand the visual content hence it is unable to produce meaningful alignment. 
We find that right after $res_2$ seems to be
a good sweet spot. 
We have also experimented with multiple insertions of STAN layers but it only leads to marginal improvement. 

\noindent\textbf{What is the best parameterization for the deformation network?}
In Section~\ref{sec:deformnet}, we have discussed two ways of parameterization for the deformation network, affine transformation (Eqn.~\ref{eqn:affine}) and attention transformation (Eqn.~\ref{eqn:att}). 
Each has 12 and 6 degree-of-freedom (DoF), respectively.
We use affine transformation in our main experiments and here we experiment with attention transformation
and homography transformation.
For homography transformation, the last element in the $4\times4$ matrix is set as 1 hence the DoF is 15, which is shown in Table~\ref{tab:ablation} (``H, 15'').
Results are shown in Table~\ref{tab:ablation}. As we see, the model with the most free parameters does not necessarily lead to better performance.

\noindent\textbf{Does temporal grouping help?}
In this experiment, we validate the efficacy of temporal grouping (Section~\ref{sec:tg}). 
The main experiments are conducted with a temporal group of 2, and the model performance drops by a big margin if temporal grouping is removed.
We then try applying a temporal group of 32, which is the same as the number of video frames, to see whether having individual transformations on every frame would help.
As we see, the computation cost drops slightly and no improvement over the baseline.

\noindent\textbf{Does {\fancyname} transfer well?}
Finally, we conduct an experiment to see whether the deformation network learned from a dataset can be generalized to another.
We train the original {\fancyname} model on Kinetics-400 dataset, and then only fine-tune the layers after the {\fancyname} layer on AVA.
As we see in Table~\ref{tab:ablation} (``fixed $W_{\theta}$''), {\fancyname} can still achieve reasonable improvement on AVA, suggesting the deformation can be transferred from one dataset to another.

\subsection{Action Recognition}
\label{sec:exp_rec}
The action recognition task is defined to be a classification task given a trimmed video clip. To evaluate the generalization abilities of our proposed model, we consider three major datasets, Kinetics-400~\cite{kay2017kinetics}, Charades~\cite{sigurdsson2016hollywood} and Charades-Ego~\cite{sigurdsson2018charades}.

\noindent\textbf{Datasets.}
Kinetics-400~\cite{kay2017kinetics} consists of about 240k training videos and 20k validation videos in 400 human action classes. The videos are about 10 seconds long. Following previous works~\cite{feichtenhofer2019slowfast,wang2018non}, we report top-1 and top-5 classification accuracy.
Charades~\cite{sigurdsson2016hollywood} is a dataset with longer ( about 30 seconds on average) videos of indoor activities. There are about 9.8k training videos and 1.8k validation videos in 157 classes in a multi-label classification setting.
Charades-Ego~\cite{sigurdsson2018charades} has the same 157 action labels but consists of both third-person view and first-person view videos. Essentially, this dataset shows different perspectives of the same actions.
MEVA~\cite{corona2021meva} is a multi-view dataset with actions of 35 classes. Many action instances are captured by multiple synchronized cameras, which makes it ideal to test our proposed method for viewpoint-invariant representations.
Performance is measured in mean Average Precision (mAP).

\noindent\textbf{Training.}
Our models on Kinetics are trained from scratch with random initialization, without using any pre-training (same as in ~\cite{feichtenhofer2019slowfast}). 
We use a learning rate of 0.2 and cosine learning rate decay. We use synchronized SGD to train for 100 epochs with a batch size of 16 on a 4-GPU machine, with a linear warm-up from 0.01 for the first 20 epochs.
We use a weight decay of $10^{-7}$. 
On Charades, Charades-Ego and MEVA, we initialize the models using models trained on Kinetics-400.
We use a learning rate of 0.02 and train for 50 epochs, where learning rate reduces to its 1/10 at epoch 40. We use a linear warm-up from 0.000125 for the first 2 epochs.
For all four datasets, we use random $224\times224$ crops and horizontal flipping from a video clip, which is randomly sampled from the full-length video and resized to a shorter edge side of randomly sampled in [256, 320] pixels.

\noindent\textbf{Inference.}
Following previous works~\cite{feichtenhofer2019slowfast,wang2018non}, we sample $3\times10$ clips for each video during testing: we uniformly sample 10 clips for the temporal domain and 3 spatial crops of size $256\times256$ after the shorter edge size are resized to 256 pixels.
We average the softmax scores across all clips for final prediction for Kinetics-400 and use the maximum of the softmax scores for Charades, Charades-Ego and MEVA.
We report the computational cost of a single, spatially center-cropped clip of size $256\times256$.

\begin{table}[]
\centering
\begin{tabular}{l|c|c|c}
\hline
        Models              & top-1          & top-5   & GFLOPs       \\ \hline
        I3D~\cite{kay2017kinetics} & 0.711          & 0.893 & - \\
        R(2+1)D~\cite{tran2018closer} & 0.720         & 0.900 & - \\
        \small{DynamoNet (32 frames)}~\cite{diba2019dynamonet} & 0.714          & 0.900 & -          \\ 
NL-R50 (32 frames)~\cite{wang2018non}     & 0.749          & 0.916 & -          \\ \hline
ResNet3D (8x8)             & 0.735          & 0.908   & 109.2       \\ 
ResNet3D + {\fancyname} & \textbf{0.751} & \textbf{0.916} & 113.2 \\ \hline
SlowFast~\cite{feichtenhofer2019slowfast}  (32x2)*           & 0.759          & 0.920     & 131.7     \\
SlowFast + {\fancyname} & \textbf{0.774} & \textbf{0.931} & 134.5\\ \hline
\end{tabular}
\caption{Experiment results on Kinetics-400 dataset. We compare recent models with ResNet-50 backbone as well as some classic methods.
*We have implemented the ResNet3D and SlowFast model using Tensorflow based on the official code. We convert the released model weights and notice a drop of top-1 accuracy from 77.0\% in the paper~\cite{feichtenhofer2019slowfast} to 75.9\%. 
The reason might be that a small portion of videos
in Kinetics-400 validation set are unavailable at the time of
our experiments. 
In terms of FLOPs, we
compute FLOPs using Tensorflow’s profiler on each dataset
(vs. PyTorch profiler in the SlowFast paper). The FLOPs
difference might be due to profiler differences. 
We will release our code and models
for reproduction.}
\label{tab:kinetics}
\end{table}

\subsubsection{Recognition Results}

We compare the same baselines, as mentioned in the previous section. 
For Kinetics-400, the input frames are the same as the previous section. For Charades and Charades-Ego, we use $16\times8$ input frames for ResNet3D and $32\times4$ input frames for SlowFast, following the SlowFast paper~\cite{feichtenhofer2019slowfast}.

\noindent\textbf{Kinetics-400.} Table~\ref{tab:kinetics} shows the experiments on Kinetics-400. Our method can improve top-1 accuracy by 1.6 and 1.5 point for ResNet3D and SlowFast, respectively, at the cost of 3.6\% and 2.1\% relatively more computation.

\begin{table}[t!]
\centering
\begin{tabular}{l|c|c|c}
\hline
Models         & mAP  & GFLOPs & MParams\\ \hline
ResNet3D  (16x8)            &  0.354   &    218.4 & 32.40   \\ 
ResNet3D + {\fancyname} & \textbf{0.377} & 226.4 & 32.47 \\ \hline
SlowFast~\cite{feichtenhofer2019slowfast}   (32x4)       & 0.386 & 131.7 & 34.51        \\ 
SlowFast + {\fancyname} & \textbf{0.406}  & 134.5 & 34.53    \\\hline
\end{tabular}
\caption{Experiment results on the Charades dataset. We show the model performance on mAP, computation cost (in billions), and number of parameters (in millions). The computational cost is of a single 256x256 video clip of length 16x8 or 32x4 (number of frames x sampling rate) frames.}
\label{tab:charades}
\end{table}

\noindent\textbf{Charades.} Table~\ref{tab:charades} shows the experiments on the Charades dataset. 
Our method is able to improve mAP by 2.3 and 2 points for ResNet3D and SlowFast, respectively, with only 3.6\% and 2.1\% relatively more computation.
Our method only contains a few additional parameters.

\begin{table}[t!]
\centering
\begin{tabular}{l|c|c}
\hline
Models         & 1st-person & 3rd-person  \\ \hline
Baseline v1.0~\cite{sigurdsson2018charades}   &    0.282                           & 0.232     \\ \hline
ResNet3D  (16x8)            & 0.298                           & 0.361   \\ 
ResNet3D + {\fancyname} & \textbf{0.318}                  & \textbf{0.366} \\ \hline
SlowFast~\cite{feichtenhofer2019slowfast}   (32x4)      & 0.316                           & 0.391    \\ 
SlowFast + {\fancyname} & \textbf{0.326}                  & \textbf{0.396}   \\ \hline
\end{tabular}
\caption{Experiment results on Charades-Ego dataset. Models are trained on both 1st-person and 3rd-person view videos. We show the test set with 1st-person and 3rd-person view videos separately. The computation cost and number of parameters are the same as the Charades experiment.}
\label{tab:charades-ego}
\end{table}

\begin{table}[t!]
\centering
\begin{tabular}{l|c}
\hline
Models         & mAP  \\ \hline
ResNet3D  (16x8)                                      & 0.455   \\ 
ResNet3D + {\fancyname}                   & \textbf{0.497} \\ \hline
SlowFast~\cite{feichtenhofer2019slowfast}   (32x4)      & 0.484                               \\ 
SlowFast + {\fancyname}                   & \textbf{0.531}   \\\hline
\end{tabular}
\caption{Experiment results on MEVA dataset.}
\label{tab:action_meva}
\end{table}

\noindent\textbf{Charades-Ego.} Table~\ref{tab:charades-ego} show the results on the Charades-Ego dataset.
The test set is divided into 1st-person videos and 3rd-person videos. Note that the training set of Charades-Ego dataset includes mostly 3rd-person videos, and we can observe from the results that our {\fancyname} model achieves more significant improvement on the 1st-person test set, verifying the efficacy of our model's generalization ability.
With SlowFast and {\fancyname} model, we are able to achieve state-of-the-art performance on this dataset.

\noindent\textbf{MEVA.} Table~\ref{tab:action_meva} show the results on the MEVA dataset.
We can observe from the results that our {\fancyname} model achieves significantly better results with both backbones.

\begin{figure}[t!]
	\centering
		\includegraphics[width=0.47\textwidth]{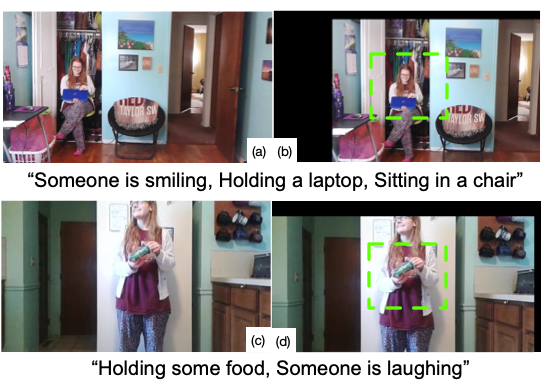} 
	\caption{Qualitative analysis. Videos are from Charades. (a) and (c) are original frames. (b) and (d) are  transformed frames. The green bounding box denotes a center reference. The sentences below the images are the ground truth labels for the video clip. See main text for details of analysis.
	}
	\label{fig:qual}
\end{figure}

\noindent\textbf{Qualitative Analysis.}
{\fancyname} can provide intuitive geometric interpretation of human actions.  
To illustrate such effects, we visualize example transformations of our model on Charades videos on two videos with correct classification labels in Fig.~\ref{fig:qual}. We use the output ($\theta$, Eqn.~\ref{eqn:affine}) from the {\fancyname} layer (deformation network) of the first temporal group and visualize the spatial transformation using the middle frame. 
Fig.~\ref{fig:qual} (a) and (c) are the original frames.
Fig.~\ref{fig:qual} (b) and (d) are the transformed frames using the transformation matrix predicted by {\fancyname}.
We use a center bounding box as a reference in the visualization. As we see, for the first example, the person originally is located to the left of the scene and {\fancyname} learns a transformation that centers the main actor.
In the second example, the person is already in the center and the transformation does not alter much of the frames.

\noindent\textbf{Limitations.}
When examining the model transformation, we have noticed that the model does not learn much meaningful temporal transformation as we have expected.
Ideally, for action videos like dunking basketball, the model should be able to slow-down video segments like jumping and speed-up parts like running. 
More work needs to be done during training of the transformation to allow the model to learn better temporal transformation. For example, removing residual connection for STAN layer might help, but it may require much longer training time.
In addition, different people might perform the same action differently. This kind of personal variances should also be considered in future work.

\section{Related Work}

The research of action recognition has advanced with both new datasets and new models.  As one of the earliest action recognition benchmarks, KTH\cite{kth-icpr04} collects videos of individual actors repetitively performing six types of human actions (walking, jogging, running, boxing, hand waving and clapping) with a clean background. Because these videos are very simple, KTH dataset turns out to be a very easy benchmark since studies quickly obtained near-perfect accuracy on it \cite{DBLP:conf/cvpr/CaoLH10,DBLP:conf/cvpr/WangKSL11}. 

To overcome the limitation of KTH, 
the HMDB dataset \cite{hmdb51} was proposed in 2011 with 51 actions in 7000 video clips, while 
UCF101 \cite{ucf101} extended this effort by collecting 101 action classes in 13000 clips. Both benchmarks are captured 
with more diversified backgrounds. In the past decade, we have witnessed a steady improvement of accuracy on these two datasets by different methods including features fusion \cite{DBLP:journals/cviu/PengWWQ16}, two-stream network \cite{twostream-Simonyan-nips14}, C3D \cite{cnn-dtran-iccv15},  I3D \cite{cnn-Carreira-Zisserman-cvpr17}, graph-based approaches \cite{wang2018videos,chen2019graph,qi2018learning,zhang2019structured} and others \cite{xu2017r,chao2018rethinking,hou2017tube}. However, some clips in the UCF101 test set are taken from the same YouTube video as the training set \cite{kay2017kinetics}, which makes it relatively easy to obtain good accuracy on UCF101. As a result, the SOTA on UCF101 dataset is more than 98\%\cite{kay2017kinetics}. 

The modern benchmarks for action recognition and detection is the Kinetics dataset \cite{kay2017kinetics} and the AVA dataset \cite{li2020ava}, respectively. The Kinetics dataset proposes a bigger benchmark with more categories and more videos (e.g., 400 categories 160,000 clips in  \cite{kay2017kinetics}) as a harder benchmark. The action labels in AVA \cite{li2020ava} are annotated with spatial temporal locations, which is more challenging than the setting of one label per clip.

Many new approaches ~\cite{tran2018closer,zhao2018trajectory,lin2019tsm,feichtenhofer2019slowfast,yang2020temporal} have been carried on these two datasets, of which the SlowFast network \cite{feichtenhofer2019slowfast} obtains good performance. Note that the SlowFast contains more parameters than C3D or I3D networks, by integrating features at both high and low frame rates. 
We can see the trend of action recognition in the last two decades is to collect bigger datasets (e.g. Kinetics) as well as build bigger models (e.g., I3D and SlowFast).

In the recent years, there has been a consistent effort to use alignment for image recognition
\cite{DBLP:conf/nips/HuangMLL12,Xiong_2013_CVPR,jaderberg2015spatial,lin2017inverse,kosiorek2019stacked,dai2017deformable}. 
However, many previous studies show that alignment models are not as competitive as data-driven approaches like data augmentation or spatial pooling for image recognition.
Some recent works have to rely on very expensive models such as
 recurrent networks \cite{lin2017inverse} or stacked capsules \cite{kosiorek2019stacked}.
As a result, a lot of alignment-based recognition methods are limited to MNIST \cite{kosiorek2019stacked} and face recognition \cite{Xiong_2013_CVPR}. 
Some follow-up works on capsule network~\cite{duarte2018videocapsulenet} and 2D spatial alignment network~\cite{huang2019part,yang2019step} have been proposed but they are limited to action recognition on small datasets like UCF101~\cite{ucf101} and JHMDB~\cite{jhuang2013towards}.
One popular related work, deformable convolution~\cite{dai2017deformable}, computes spatial offsets to deform traditional convolution operations based on the visual content in images. Our method defers in that: (1) we aim for spatial-temporal alignment of features; (2) we explicitly compute geometric transformation.
Another related work, trajectory convolutions~\cite{zhao2018trajectory}, relies on optical flow estimation to compute spatial offset at different times for 3D convolutions.
Our method utilizes higher level content understanding to
account for actor movements and viewpoint changes (act as
camera stabilization, etc.). Our method achieves 0.853 (vs.
0.798 in \cite{zhao2018trajectory}) on Kinetics-400 on avg. top-1\&5 acc.
In this chapter, we show that it is possible to build an efficient spatial-temporal alignment for both action recognition and detection, and improve recent networks with very few extra parameters.

\section{Summary}
In this chapter we introduce a new spatial-temporal alignment network, {\fancyname}, for action recognition and detection.
Our study is the first to explore explicit spatial-temporal alignment in 3D CNNs for action detection.
Our model can be conveniently inserted into existing networks and provides significant improvement with a low extra computation cost.
We have shown that our method achieves state-of-the-art performance on multiple challenging action recognition and detection benchmarks.

\part{Pedestrian Trajectory Prediction with Scene Semantics}
\label{part:future_prediction}
In this part, we focus on the human future trajectory prediction problem.
We study how trajectory prediction can benefit from scene semantic understanding of the scene.
Since the future is uncertain, we first introduce the \textit{Multiverse} model to tackle the multiple-future trajectory prediction problem (\autoref{chap:0201_multi}).
To alleviate the limited training data challenge as mentioned in previous section, we propose a machine learning algorithm called \textit{SimAug}, to efficiently learn from 3D simulation data for trajectory prediction (\autoref{chap:0202_3d}).
\chapter{Multiple-future Pedestrian Trajectory Prediction 
}  \label{chap:0201_multi}

In this chapter, we study the uncertainty of future trajectory predictions, by proposing the \textit{Multiverse} model~\cite{liang2020garden}, which generates multiple-future trajectories with probability distributions. 
We also develop a novel multiple-future trajectory benchmark, called the \textit{ForkingPaths} dataset, using 3D simulation.

\section{Motivation}

\begin{figure}[ht]
	\centering
		\includegraphics[width=0.97\textwidth]{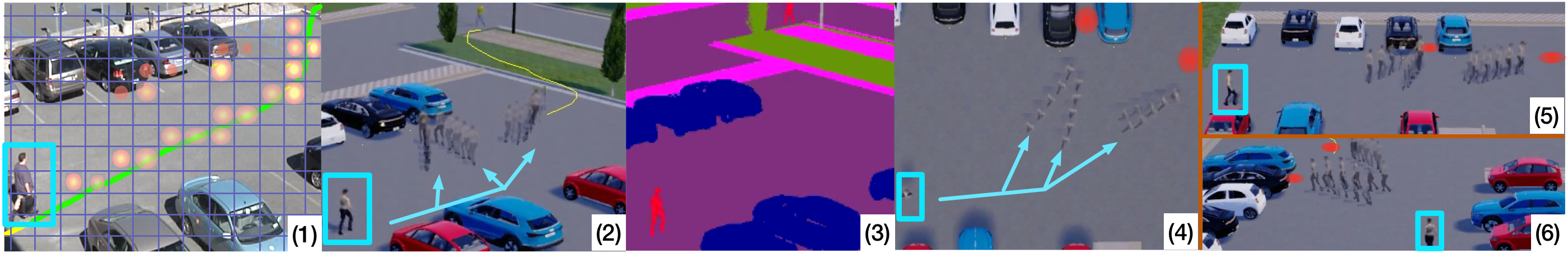} 
	\caption{Illustration of person trajectory prediction.
    (1) A person walks towards a car (data from
    the VIRAT/ActEV dataset).
    The green line is the actual future trajectory and the yellow-orange heatmaps are example future predictions.
    Although these predictions near the cars
    are plausible, they would be considered errors in
    the real video dataset. 
    (2) To combat this,
     we propose a new dataset called ``Forking Paths'';
     here we illustrate 3 possible futures
     created by human annotators controlling agents
      in a synthetic world derived from real data.
     (3) Here we show semantic segmentation of the scene.
     (4-6) Here we show the same scene rendered from different viewing angles, where the red circles are future destinations.}
	\label{fig:multiFutures}
\end{figure}

Forecasting future
human behavior is a fundamental problem in video understanding. In particular, future path prediction, which aims at forecasting a pedestrian's future trajectory in the next few seconds, has received 
a lot of attention in our community~\cite{kitani2012activity,alahi2016social,gupta2018social,li2019way}. This functionality 
is a key component in a variety of applications such as autonomous driving~\cite{bansal2018chauffeurnet,chai2019multipath}, long-term object tracking~\cite{kalman1960new, sadeghian2017tracking}, safety monitoring~\cite{liang2019peeking}, robotic planning~\cite{rhinehart2017first,rhinehart2018r2p2},
etc.

Of course, the future is often very uncertain:
Given the same historical trajectory, a person
may take different paths, depending on their 
(latent) goals. Thus recent work has started focusing on  \emph{multi-future trajectory prediction}~\cite{tang2019multiple,chai2019multipath,li2019way,makansi2019overcoming, thiede2019analyzing,lee2017desire}.

Consider the example in Fig.~\ref{fig:multiFutures}.
We see a person moving from the bottom left towards the top right of
the image, and our task is to predict where he will go next.
Since there are many possible future trajectories this person
might follow, we are interested in learning a model
that can generate multiple plausible futures.
However, since the ground truth data only contains one trajectory, it is difficult to evaluate such
probabilistic models.

To overcome the aforementioned challenges,
our first contribution is the creation of a realistic synthetic dataset that allows us to compare models in a quantitative way in terms of their ability to predict multiple plausible futures,
rather than just evaluating them against a single observed trajectory as in existing studies.
We create this dataset using the
3D CARLA~\cite{dosovitskiy2017carla} simulator, where the scenes are manually designed to be similar to those found in the  challenging real-world benchmark VIRAT/ActEV~\cite{oh2011large,2018trecvidawad}.
Once we have recreated the static scene, we automatically reconstruct trajectories by projecting real-world data to the 3D simulation world. See Fig.~\ref{fig:multiFutures} and~\ref{fig:multiverse_dataset}.
We then semi-automatically select a set of plausible future destinations (corresponding to semantically meaningful locations in the scene), and ask human annotators to create multiple possible
continuations of the real trajectories towards each such goal. In this way, our dataset is ``anchored'' in reality,
and yet contains plausible variations in high-level human behavior, which is impossible to simulate automatically.

We call this dataset the ``Forking Paths'' dataset,
a reference to the short story by Jorge Luis Borges.\footnote{
\footnotesize{\url{https://en.wikipedia.org/wiki/The_Garden_of_Forking_Paths}}
} %
As shown in Fig.~\ref{fig:multiFutures}, different human annotations have created forkings of future trajectories for the identical historical past.
So far, we have collected 750 sequences, with each covering about 15 seconds, 
from 10 annotators, controlling 127 agents in 7 different scenes. 
Each agent contains 5.9 future trajectories on average.
We render each sequence from 4 different views,
and automatically generate dense labels,
as illustrated in Fig.~\ref{fig:multiFutures} and \ref{fig:multiverse_dataset}.
In total,
this amounts to 3.2 hours of trajectory sequences, which is comparable to the largest person trajectory benchmark VIRAT/ActEV~\cite{2018trecvidawad, oh2011large} (4.5 hours), or 5 times bigger than the common ETH/UCY~\cite{lerner2007crowds,luber2010people} benchmark.
We therefore believe this will serve as a benchmark
for evaluating models that can predict multiple futures.

Our second contribution is to propose a new probabilistic model, \textit{\textit{Multiverse}}, which can generate multiple plausible trajectories given the past history of locations and the scene.
The model contains two novel design decisions. First, we use a multi-scale representation of locations. 
In the first scale, the coarse scale, we represent locations on a 2D grid, as shown in Fig.~\ref{fig:multiFutures}(1). This captures high level uncertainty about possible destinations and leads to a better representation of multi-modal distributions. In the second fine scale, we predict a real-valued offset for each grid cell, to get more precise localization. This two-stage approach is partially inspired by object detection methods~\cite{ren2015faster}. 
The second novelty of our model is to design convolutional RNNs~\cite{xingjian2015convolutional} over the spatial graph as a way of encoding inductive bias about the movement patterns of people.

In addition, we empirically validate our model on the challenging real-world benchmark VIRAT/ActEV~\cite{oh2011large,2018trecvidawad} for single-future trajectory prediction, in which our model achieves the best-published result. On the proposed simulation data for multi-future prediction, experimental results show our model compares favorably against the state-of-the-art models across different settings. To summarize, the main contributions of this chapter are as follows:
\textit{(i)} We introduce the first dataset and evaluation methodology that allows us to compare models in a quantitative way in terms of their ability to predict multiple plausible futures.
\textit{(ii)} We propose a new effective model for multi-future trajectory prediction.
\textit{(iii)} We establish a new
state of the art result
on the challenging VIRAT/ActEV benchmark,
and compare
various methods
on our multi-future prediction dataset.

\section{The \textit{Multiverse} Model}

\begin{figure}[ht]
	\centering
		\includegraphics[width=\textwidth]{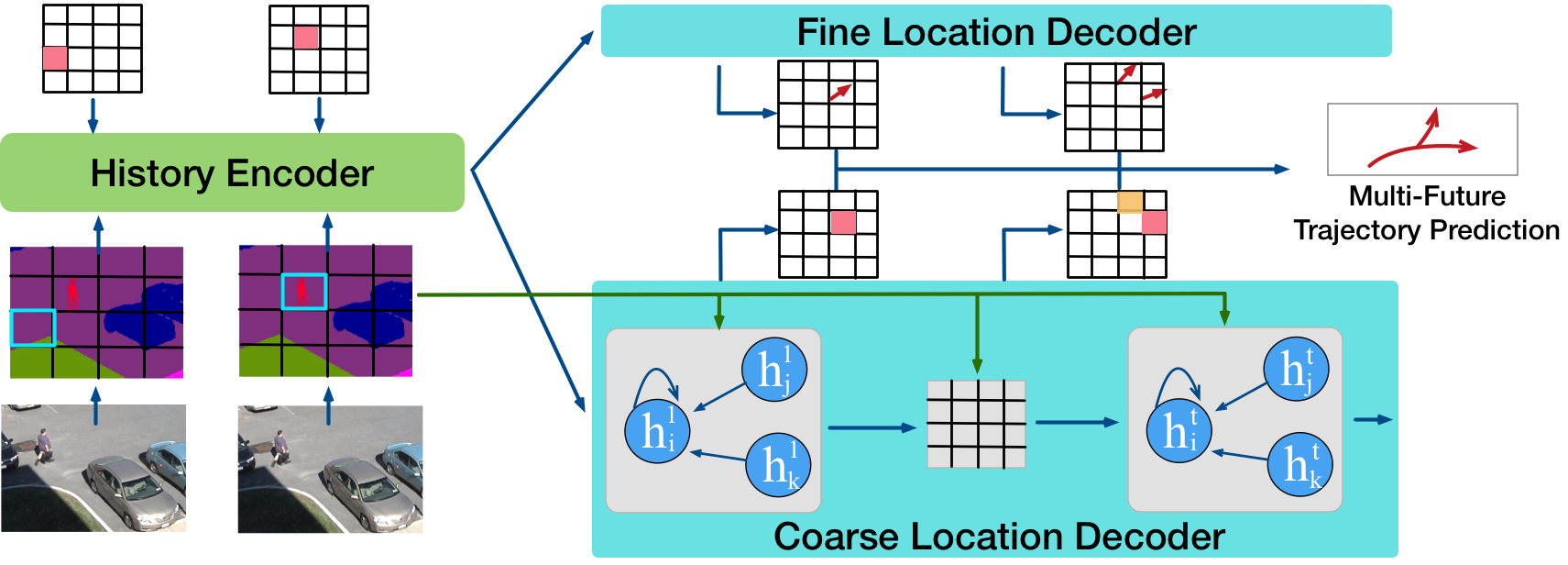}
	\caption{Overview of our model.
	The input to the model is 
	the ground truth location history,
	and a set of video frames,
	which are preprocessed by a semantic segmentation model. 
	This is encoded by the ``History Encoder''
	convolutional RNN.
	The output of the encoder is fed
	to the convolutional RNN decoder for  location prediction.
	The coarse location decoder outputs
	a heatmap over the 2D grid of size $H \times W$.
	The fine location decoder
	outputs 
	a vector offset within each grid cell.
	These are combined to generate a multimodal distribution
	over $\real^2$ for predicted locations.
	}
	\label{fig:multiverse_model}
\end{figure}

In this section, we describe our 
model for forecasting agent
trajectories,
which we call \textit{Multiverse}.
We focus on predicting the locations of a single agent for multiple steps into the future, $L_{h+1:T}$, 
given a sequence of past video frames, $V_{1:h}$,
and agent locations, $L_{1:h}$,
where $h$ is the history length and $T-h$ is the prediction length.
Since there is inherent uncertainty in this task, our goal is to design a model
that can effectively predict multiple plausible future trajectories,
by computing the multimodal distribution
$p(L_{h+1:T}|L_{1:h}, V_{1:h})$.
See Fig.~\ref{fig:multiverse_model} for a high level summary of the model,
and the sections below for more details.

\vspace{-1mm}
\subsection{History Encoder}
\vspace{-1mm}
The encoder computes a representation of the scene
from the history of past locations,
$L_{1:h}$, and frames, $V_{1:h}$.
We encode each ground truth location
$L_t$ by an index $Y_t \in G$ representing
the nearest cell in a 2D grid $G$ of size $H \times W$, indexed from $1$ to $HW$.
 Inspired by~\cite{lazebnik2006beyond,lin2017feature}, 
 we encode location with two different grid scales
 ($36 \times 18$ and $18 \times 9$);
 we show the benefits of this multi-scale
 encoding in Section~\ref{sec:multiverse_ablation}.
For simplicity of presentation,
we focus on a  single $H \times W$ grid.

To make the model more invariant to low-level visual details, and thus more robust to domain shift (e.g., between different scenes, different views of the same scene, or between real and synthetic images), we preprocess each video frame $V_t$ using a pre-trained semantic segmentation model, with $K=13$ possible class labels per pixel. 
We use the Deeplab model \cite{chen2017deeplab} trained on the ADE20k~\cite{zhou2017scene} dataset,
and keep its weights frozen.
Let $S_t$ be this semantic segmentation map modeled as a tensor of size $H \times W \times K$.

We then pass these inputs to a convolutional RNN 
\cite{xingjian2015convolutional,wang2019eidetic}
to compute a spatial-temporal feature history:
\begin{equation}
H_t^e = \text{ConvRNN}(\text{one-hot}(Y_{t}) \odot
(W * S_t), H^e_{t-1})
\end{equation}
where
$\odot$ is element wise product, 
and $*$ represents 2D-convolution. The function $\text{one-hot}(\cdot)$ projects a cell index into an one-hot embedding of size $H \times W$ according to its spatial location.

We use the final state of this encoder $H_t^e \in \mathbb{R}^{H \times W \times d_{enc}}$, where $d_{enc}$ is the hidden size, to initialize the state of the decoders. 
We also use the temporal average of the semantic maps,
$\overline{S} = \frac{1}{h} \sum_{t=1}^h S_t$, during each decoding step.
The context is represented as $\context=[H_h^e,\overline{S}]$.

\subsection{Coarse Location Decoder}

After getting the context $\context$, our goal is to forecast future locations.
We initially focus on predicting locations
at the level of grid cells, $Y_t \in G$.
In Section~\ref{sec:multiverse_fineDecoder}, 
we discuss how to predict a continuous offset in $\real^2$, which specifies a ``delta''
from the center of each grid cell, to get a fine-grained location prediction.

Let the coarse
distribution over grid
locations at time $t$ (known as the ``belief state'') be denoted by
$C_t(i)=p(Y_t=i|Y_{h:t-1},\context)$,
for $\forall i \in G$ and $t \in [h+1, T]$. For brevity, we use a single
index $i$ to represent a cell in the 2D grid.
Rather than assuming a Markov model,
we update this using a 
convolutional recurrent neural network,
with hidden states $H_t^C$.
We then compute the belief state by:
\begin{align}
    C_t = \text{softmax}(W * H_t^C) \in \mathbb{R}^{HW}
    \label{eqn:multiverse_fC}
\end{align}
Here we use 2D-convolution with one filter and flatten the spatial dimension before applying softmax.
The hidden state is updated using:
\begin{align}
    H_t^C = \text{ConvRNN}(\text{GAT}(H^C_{t-1}), 
    \text{embed}(C_{t-1}))
    \label{eqn:multiverse_fH}
\end{align}
where
$\text{embed}(C_{t-1})$ embeds into a 3D tensor of size $H \times W \times d_e$ and $d_e$ is the embedding size.
$\text{GAT}(H_{t-1}^C)$ is a 
graph attention network~\cite{velivckovic2017graph},
where the graph structure corresponds to the 2D grid in $G$.
More precisely,
let $h_i$ be the feature vector corresponding
to the $i$-th grid cell in $H^C_{t-1}$,
and let $\tilde{h}_i$ be the 
corresponding output
in
$\tilde{H}^C_{t-1} = \text{GAT}(H^C_{t-1}) \in \mathbb{R}^{H \times W \times d_{dec}}$, where $d_{dec}$ is the size of the decoder hidden state.
We compute these outputs of GAT using:
\begin{align}
    \tilde{h}_i = \frac{1}{|\mathcal{N}_i|}
    \sum_{j \in \mathcal{N}_i} f_e([v_i, v_j]) + h_i
\end{align}
where $\mathcal{N}_i$ are the neighbors of node $v_i$ in $G$ with each node represented as 
$v_i = [h_i, \overline{S}_{i}]$, where $\overline{S}_{i}$ collects the cell $i$'s feature in $\overline{S}$.
$f_e$ is some edge function (implemented as an MLP in our experiments) that computes the attention weights.

The graph-structured
update function for the RNN ensures that the probability
mass  ``diffuses out'' to nearby grid cells
in a controlled manner, reflecting the prior knowledge
that people do not suddenly jump between distant locations.
This inductive bias is also encoded in the convolutional structure,
but adding the graph attention network gives improved results,
because the weights are input-dependent and not fixed.

\subsection{Fine Location Decoder}
\label{sec:multiverse_fineDecoder}

The 2D heatmap is useful for capturing multimodal distributions, but does not give very precise location predictions. 
To overcome this, we train a second convolutional RNN decoder
$H_t^O$
to compute an offset vector for each possible
grid cell
using a regression output, $O_t = \text{MLP}(H_t^O) \in \mathbb{R}^{H \times W \times 2}$.
This RNN is updated using
\begin{align}
    H_t^O = \text{ConvRNN}(\text{GAT}(H^O_{t-1}), O_{t-1}) \in \mathbb{R}^{H \times W \times d_{dec}}
\end{align}

\noindent To compute the final prediction location, we first flatten the spatial dimension of $O_{t}$ into $\tilde{O}_{t} \in \mathbb{R}^{HW \times 2}$. Then we use
\begin{align}
    L_t = Q_{i} + \tilde{O}_{ti}
\label{eq:multiverse_L_t}
\end{align}
where $i$ is the index of the
selected grid cell,
$Q_{i} \in \real^2$ is the center of that cell,
and $\tilde{O}_{ti} \in \real^2$ is the predicted offset for that cell at time $t$.
For single-future prediction, we use greedy search, namely $i=\argmax C_t$ over the belief state. For multi-future prediction, we use beam search in Section~\ref{sec:multiverse_inference}.

This idea of combining classification
and regression is partially inspired by
object detection methods
(e.g., \cite{ren2015faster}).
It is worth noting that in concurrent work,
\cite{chai2019multipath} also designed a two-stage
model for trajectory forecasting.
However, their classification targets are pre-defined anchor trajectories.
Ours is not limited by the predefined anchors.

\subsection{Training}
Our model trains on the observed trajectory from time 1 to $h$ and predicts the future trajectories (in  $xy$-coordinates) from time $h+1$ to $T$.
We supervise this training by providing ground
truth targets for both the heatmap (belief state), $C_t^*$,
and regression offset map, $O_t^*$.
In particular, for the coarse decoder,
 the cross-entropy loss is used:
\begin{equation}
    \mathcal{L}_{cls} = -\frac{1}{T} \sum_{t=h+1}^{T} 
    \sum_{i \in G} C_{ti}^* \log(C_{ti})
\end{equation}
For the fine decoder, we use the smoothed $L_1$ loss
used in object detection~\cite{ren2015faster}:
\begin{equation}
 \mathcal{L}_{reg} = \frac{1}{T} \sum_{t=h+1}^{T} 
 \sum_{i \in G}
 \text{smooth}_{L_1}(O_{ti}^*, O_{ti})
\end{equation}
where 
$O_{ti}^* = L_t^* - Q_{i}$ 
is the delta between the true location and the center of the grid cell at $i$ and $L_t^*$ is the ground truth for $L_t$ in Eq.\eqref{eq:multiverse_L_t}. We impose this loss on every cell to improve the robustness.

The final loss is then calculated using
\begin{align}
\mathcal{L}(\theta) = \mathcal{L}_{cls} + \lambda_1 \mathcal{L}_{reg} + \lambda_2 \|\theta\|_2^2
\end{align}
where $\lambda_2$ controls the $\ell_2$ regularization (weight decay),
and $\lambda_1=0.1$ is used
to balance the regression and classification losses.

Note that during training, when updating the 
RNN, we feed in the predicted
soft distribution
over locations, $C_{t}$. See Eq.~\eqref{eqn:multiverse_fC}.
An alternative would be to feed in the true
values, $C_t^*$, i.e., use teacher forcing.
However, this is known to suffer
from problems \cite{Ranzato2016}.

\subsection{Inference}\label{sec:multiverse_inference}

To generate multiple qualitatively distinct
 trajectories, we use the diverse beam search strategy from \cite{li2016simple}.
 To define this precisely,
 let $B_{t-1}$ be the beam at time $t-1$;
 this set contains $K$ trajectories (history
 selections) $M^k_{t-1}=\{\hat{Y}^k_1, \ldots , \hat{Y}^k_{t-1}\}$, $k \in[1 , K]$, where $\hat{Y}^k_t$ is an index in $G$,
 along with
 their accumulated log probabilities,
 $P^k_{t-1}$.
 Let $C^k_t=f(M^k_{t-1}) \in \mathbb{R}^{HW}$ be the coarse location output probability from Eq.~\eqref{eqn:multiverse_fC} and \eqref{eqn:multiverse_fH} at time $t$ given inputs $M^k_{t-1}$.
 
 The new beam is computed using

\begin{align}
 \small
    B_{t} = \text{topK}
    \left( \{
    P^k_{t-1} \!+\! \log(C^k_{t}(i)) \!+\! \gamma(i) | \forall i \in G, k \in [1 ,K]\}
    \right)
\end{align}
where $\gamma(i)$ is a diversity penalty term,
and we take the top $K$ elements from the set
produced by considering 
values with $k=1:K$.
If $K=1$, this reduces to greedy search.

Once we have computed the top $K$ future predictions,
we add the corresponding offset vectors to get
$K$ trajectories by $L^k_{t} \in \real^2$.
This constitutes the final output of our model.

\section{The Forking Paths Dataset}
\begin{figure}[!t]
	\centering
		\includegraphics[width=1.0\textwidth]{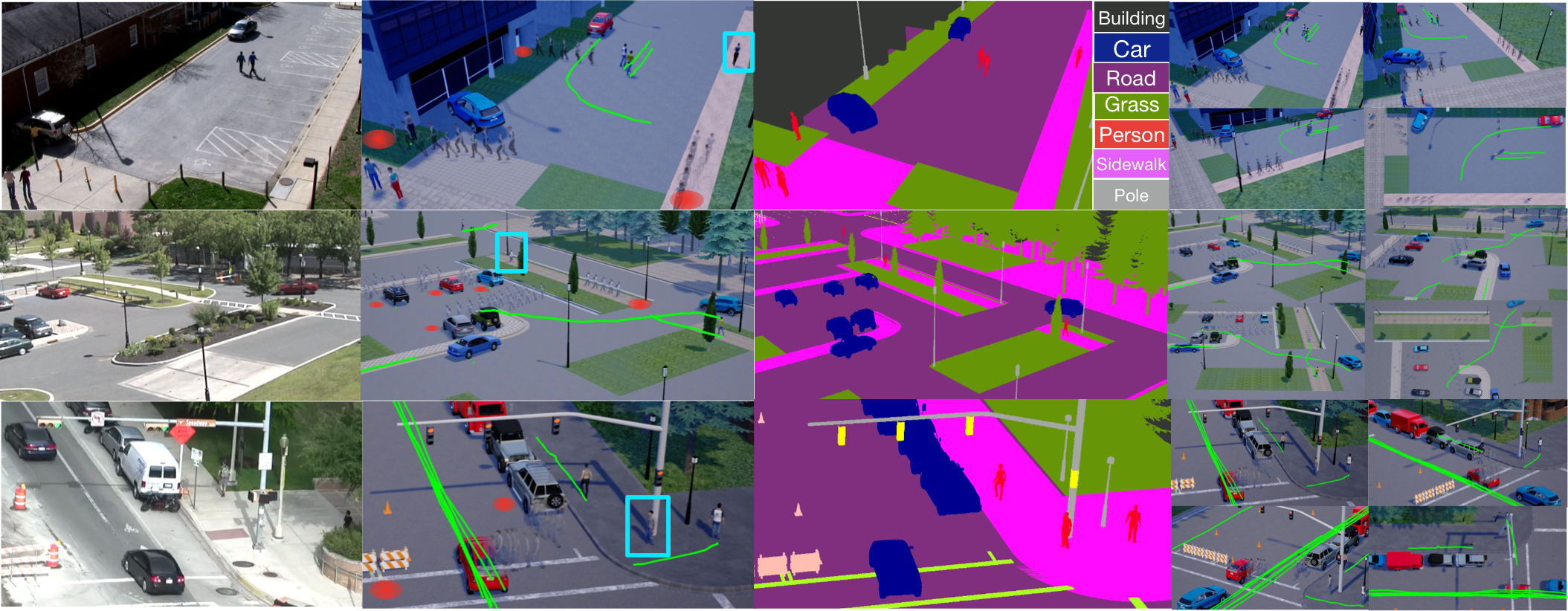}
	\caption{Visualization of the Forking Paths dataset. On the left is examples of the real videos and the second column shows the reconstructed scenes. The person in the blue bounding box is the controlled agent and multiple future trajectories annotated by humans are shown by overlaid person frames. 
	The red circles are the defined destinations. The green trajectories are future trajectories of the reconstructed uncontrolled agents. The scene semantic segmentation ground truth is shown in the third column and the last column shows all four camera views including the top-down view.} 
	\label{fig:multiverse_dataset}
\end{figure}

In this section, we describe our human-annotated simulation dataset, called Forking Paths, for multi-future trajectory evaluation.

\noindent\textbf{Existing datasets.}
There are several real-world datasets for trajectory evaluation,
such as SDD~\cite{robicquet2016learning}, ETH/UCY~\cite{pellegrini2010improving,lerner2007crowds}, KITTI~\cite{geiger2013vision}, nuScenes~\cite{caesar2019nuscenes} and VIRAT/ActEV~\cite{2018trecvidawad,oh2011large}.
However, they all share the fundamental problem
that
one can only observe one out of many possible future trajectories sampled from the underlying distribution. This is broadly acknowledged in prior works~\cite{makansi2019overcoming,thiede2019analyzing,chai2019multipath,gupta2018social,rhinehart2019precog,rhinehart2018r2p2} but has not yet been addressed. 

The closest work to ours is the simulation used in~\cite{makansi2019overcoming, thiede2019analyzing,chai2019multipath}.
However,
these only contain artificial trajectories,
not human generated ones.
Also, they use a highly simplified  2D space, with pedestrians oversimplified as points and vehicles as blocks; no other scene semantics are provided.

\noindent\textbf{Reconstructing reality in simulator.}
In this work, we use CARLA~\cite{dosovitskiy2017carla}, a near-realistic open source simulator built on top of the Unreal Engine 4. 
Following prior simulation datasets~\cite{gaidon2016virtual,ros2016synthia}, we \textit{semi-automatically} reconstruct static scenes and their dynamic elements from the real-world videos in ETH/UCY and VIRAT/ActEV.
There are 4 scenes in ETH/UCY and 5 in VIRAT/ActEV. We exclude 2 cluttered scenes (UNIV \& 0002)
that we are not able to reconstruct in CARLA, leaving 7 static scenes in our dataset.

For dynamic movement of vehicle and pedestrian, we first convert the ground truth trajectory annotations from the real-world videos to the ground plane using the provided homography matrices.
We then match the real-world trajectories' origin to correct locations in the re-created scenes. 

\noindent\textbf{Human generation of plausible futures.}
We manually select sequences with more than one pedestrian.
We also require that
at least one pedestrian could have multiple plausible alternative destinations.
We insert plausible pedestrians into the scene to increase the diversity of the scenarios.
We then select one of the pedestrians to be the ``controlled agent'' (CA) for each sequence,  and set meaningful destinations within reach, like a car or an entrance of a building.
On average, each agent has about 3 destinations to move towards.
In total, 
we have 127 CAs from 7 scenes. We call each CA and their corresponding scene a scenario.

For each scenario,  
there are on average 5.9 human annotators to control the agent to the defined destinations. 
Specifically, they are asked to watch the first 5 seconds of video, from a first-person view (with the camera slightly behind the pedestrian) and/or an overhead view (to give more context). They are then asked to control the motion of the agent so that it moves towards the specified destination in a ``natural'' way, e.g., without colliding with other moving objects (whose motion is derived from the real videos, and is therefore unaware of the controlled agent).
The annotation is considered successful if the agent reached the destination without colliding within the time limit of 10.4 seconds. 
All final trajectories in our dataset are examined by humans to ensure reliability.

Note that our videos
are up to 15.2 seconds long.
This is slightly longer than previous works
(e.g. \cite{alahi2016social,gupta2018social,liang2019peeking,sadeghian2018sophie,li2019way,zhang2019sr,zhao2019multi})
that use
3.2 seconds of observation and 
4.8 seconds for prediction.
(We use 10.4 seconds for the future
 to allow us to evaluate longer term forecasts.)

\noindent\textbf{Generating the data.}
Once we have collected human-generated trajectories, 750 in total after data cleaning, 
we render each one in four camera views (three 45-degree and one top-down view). Each camera view has 127 scenarios in total and each scenario has on average 5.9 future trajectories.
With CARLA,
we can also simulate different weather conditions,
although we did not do so in this work.
In addition to agent location, we collect ground truth for pixel-precise scene semantic segmentation from 13 classes including sidewalk, road, vehicle, pedestrian, etc. See Fig.~\ref{fig:multiverse_dataset}.

\section{Experimental Results}
\label{sec:0201_exp}
This section evaluates various methods,
including our \emph{\textit{Multiverse}} model,
for multi-future trajectory prediction
on the proposed Forking Paths dataset. 
To allow comparison with previous works,
we also evaluate our model on the challenging VIRAT/ActEV~\cite{2018trecvidawad,oh2011large} benchmark for single-future path prediction.

\noindent\textbf{Multi-Future Evaluation.}
Let $Y^{ij}=Y^{ij}_{t=(h+1)\cdots T}$ be the $j$-th true future trajectory for the $i$-th test sample, for $\forall j \in [1,J]$,
and let $\hat{Y}^{ik}$ be the $k$'th sample from the predicted
distribution over trajectories, for $k \in [1,K]$.
Since there is no agreed-upon evaluation metric for this setting,
we simply extend the above metrics, as follows:
\noindent i) \textit{Minimum Average Displacement Error Given K Predictions with Multi-modal Ground-Truth} ($\text{minADE}^M_K$): similar to the metric described in ~\cite{chai2019multipath, rhinehart2018r2p2,rhinehart2019precog, gupta2018social}, for each true trajectory $j$ in test sample $i$,
we select
the closest overall prediction (from the $K$ model predictions),
and then measure its average error:
\begin{equation}
    \text{minADE}^M_K = \frac{  \sum^{N}_{i=1} \sum^{J}_{j=1} \min_{k=1}^K \sum^{T}_{t=h+1} \lVert Y^{ij}_t - \hat{Y}^{ik}_t \rVert_{2}}{N \times (T-h) \times J}
\end{equation}

\noindent ii) \textit{Minimum Final Displacement Error Given K Predictions with Multi-modal Ground-Truth} ($\text{minFDE}^M_K$): similar to $\text{minADE}^M_K$, but we only consider the predicted points and the ground truth point at the final prediction time instant:
\begin{equation}
    \text{minFDE}^M_K = \frac{  \sum^{N}_{i=1} \sum^{J}_{j=1} \min_{k=1}^K \lVert Y^{ij}_{T} - \hat{Y}^{ik}_{T} \rVert_{2}}{N \times J}
\end{equation}

\noindent iii) \textit{Negative Log-Likelihood} (NLL):
Similar to NLL metrics used in~\cite{makansi2019overcoming,chai2019multipath}, we measure the
fit of ground-truth samples to the predicted distribution.

\subsection{Multi-Future Prediction on Forking Paths}\label{sec:multiverse_exp-multi}

\begin{table}
\centering
\begin{tabular}{l|c|c|c|c|c}
\hline
\multirow{2}{*}{Method} & \multirow{2}{*}{Input Types} &  \multicolumn{2}{c|}{$\text{minADE}^M_{20}$} & \multicolumn{2}{c}{$\text{minFDE}^M_{20}$}  \\ \cline{3-6} 
 & & 45-degree         & top-down        & 45-degree         & top-down                   \\ \hline \hline
Linear & Traj. &  213.2    & 197.6    & 403.2    & 372.9   \\ 
LSTM  & Traj.              &  201.0 {\small$\pm$2.2} & 183.7 {\small$\pm$2.1} & 381.5 {\small$\pm$3.2} & 355.0 {\small$\pm$3.6}  \\ 
Social-LSTM \cite{alahi2016social}& Traj.         &  197.5 {\small$\pm$2.5} & 180.4 {\small$\pm$1.0}  & 377.0 {\small$\pm$3.6} & 350.3 {\small$\pm$2.3}   \\ 
Social-GAN (PV) \cite{gupta2018social}& Traj.         &  191.2 {\small$\pm$5.4} & 176.5 {\small$\pm$5.2}  & 351.9 {\small$\pm$11.4} & 335.0 {\small$\pm$9.4}   \\ 
Social-GAN (V) \cite{gupta2018social}& Traj.         &  187.1 {\small$\pm$4.7} & 172.7 {\small$\pm$3.9}  & 342.1 {\small$\pm$10.2} & 326.7 {\small$\pm$7.7}   \\ 
Next \cite{liang2019peeking}& \scriptsize{Traj.+Bbox+RGB+Seg.}   &    186.6 {\small$\pm$2.7} & 166.9 {\small$\pm$2.2}  & 360.0 {\small$\pm$7.2} & 326.6 {\small$\pm$5.0}    \\ 
 Ours& Traj.+Seg. & \textbf{168.9} {\small$\pm$2.1} & \textbf{157.7} {\small$\pm$2.5} & \textbf{333.8} {\small$\pm$3.7}  & \textbf{316.5} {\small$\pm$3.4}    \\ \hline
\end{tabular}
\caption{Comparison of different methods on the Forking Paths dataset. Lower numbers are better. 
The numbers for the column labeled ``45 degrees'' are averaged
over 3 different 45-degree views.
For the input types, ``Traj.'', ``RGB'', ``Seg.'' and ``Bbox.'' mean the inputs are $xy$ coordinates, raw frames, semantic segmentations and bounding boxes of all objects in the scene, respectively.
All models are trained on real VIRAT/ActEV videos
and tested on synthetic (CARLA-rendered) videos.
}
\label{tab:multiverse_exp-multi}
\end{table}

\noindent\textbf{Dataset \& Setups.} 
The proposed Forking Paths dataset is used for multi-future trajectory prediction evaluation. 
Following the setting in previous works~\cite{liang2019peeking, alahi2016social,gupta2018social,sadeghian2018sophie,makansi2019overcoming}, 
we downsample the videos to 2.5 fps and extract person trajectories using code released in~\cite{liang2019peeking}, and let the models observe 3.2 seconds (8 frames) of the controlled agent before outputting trajectory coordinates in the pixel space. 
Since the length of the ground truth future trajectories are different, each model needs to predict the maximum length at test time but we evaluate the predictions using the actual length of each true trajectory.

\noindent\textbf{Baseline methods.} 
We compare our method with  two simple baselines, and three recent methods with released source code,
including  a recent model for multi-future prediction and the state-of-the-art model for single-future prediction:
\textbf{\textit{Linear}} is a single layer model that predicts the next coordinates using a linear regressor based on the previous input point.
\textbf{\textit{LSTM}} is a simple LSTM~\cite{hochreiter1997long} encoder-decoder model with coordinates input only. 
\textbf{\textit{Social LSTM}}~\cite{alahi2016social}: We use the open source implementation from {\footnotesize (\url{https://github.com/agrimgupta92/sgan/})}.
\textbf{\textit{Next}}~\cite{liang2019peeking} is the state-of-the-art method for single-future trajectory prediction on the VIRAT/ActEV dataset. We train the Next model without the activity labels for fair comparison using the code from {\footnotesize (\url{https://github.com/google/next-prediction/})}.
\textbf{\textit{Social GAN}}~\cite{gupta2018social} is a recent multi-future trajectory prediction model trained using Minimum over N (MoN) loss. We train two model variants (called PV and V) detailed in the paper using the code from~\cite{gupta2018social} .

All models are trained on real videos
(from VIRAT/ActEV -- see Section~\ref{sec:multiverse_exp-virat} for  details)
and tested on our synthetic videos
(with CARLA-generated pixels,
and annotator-generated trajectories).
Most models just use trajectory data as input,
except for our model
(which uses trajectory and semantic segmentation)
and Next
(which uses trajectory, bounding box,
semantic segmentation, and RGB frames).

\noindent\textbf{Implementation Details.}
We use ConvLSTM~\cite{xingjian2015convolutional} cell for both the encoder and decoder. 
The embedding size is set to 32, and the hidden sizes for the encoder and decoder are both 256. 
The scene semantic segmentation features are extracted from the deeplab model~\cite{chen2017deeplab}, pretrained on the ADE-20k~\cite{zhou2017scene} dataset.
We use Adadelta optimizer~\cite{zeiler2012adadelta} with an initial learning rate of 0.3 and weight decay of 0.001. 
Other hyper-parameters for the baselines are the same to the ones in ~\cite{gupta2018social, liang2019peeking}.
We evaluate the top $K=20$ predictions for multi-future trajectories. For the models that only output a single trajectory, including Linear, LSTM, Social-LSTM, and Next, we duplicate the output for $K$ times before evaluating. For Social-GAN, we use $K$ different random noise inputs to get the predictions. For our model, we use diversity beam search~\cite{li2016simple,plotz2018neural} as described in Section~\ref{sec:multiverse_inference}.

\begin{table}[]
\centering
\begin{tabular}{l||c|c|c}
\hline
Method & $T_{pred}=1$   & $T_{pred}=2$ & $T_{pred}=3$  \\ \hline \hline
(PV) [14]    &  10.08 {\scriptsize$\pm$0.25} & 17.28 {\scriptsize$\pm$0.42} &23.34 {\scriptsize$\pm$0.47}  \\ 
(V) [14]    &  9.95 {\scriptsize$\pm$0.35} & 17.38 {\scriptsize$\pm$0.49} &23.24 {\scriptsize$\pm$0.54} \\ 
Next [27]  & 8.32 {\scriptsize$\pm$0.10} & 14.98 {\scriptsize$\pm$0.19} &22.71 {\scriptsize$\pm$0.11} \\ 
 Ours & \textbf{2.22} {\scriptsize$\pm$0.54} & \textbf{4.46} {\scriptsize$\pm$1.33} & \textbf{8.14} {\scriptsize$\pm$2.81} \\ \hline
\end{tabular}
\caption{Negative Log-likelihood comparison of different methods on the Forking Paths dataset. For methods that output multiple trajectories, we quantize the xy-coordinates into the same grid as our method and get a normalized probability distribution prediction.} 
\label{tab:multiverse_exp-nll}
\end{table}

\noindent\textbf{Quantitative Results.} 
Table~\ref{tab:multiverse_exp-multi} lists the multi-future evaluation results, where we divide the  evaluation according to the viewing angle of camera, 45-degree vs. top-down view. 
We repeat all experiments (except ``linear'') 5 times with random initialization to produce the mean and standard deviation values.
As we see, our model outperforms baselines in all metrics and it performs significantly better on the $\text{minADE}^M_K$ metric, which suggests better prediction quality over all time instants.
Notably, our model outperforms Social GAN by a large margin of at least 8 points on all metrics.
We also measure the standard negative log-likelihood (NLL) metric for the top methods in Table~\ref{tab:multiverse_exp-nll}.

\begin{figure}[!t]
	\centering
		\includegraphics[width=1.0\textwidth]{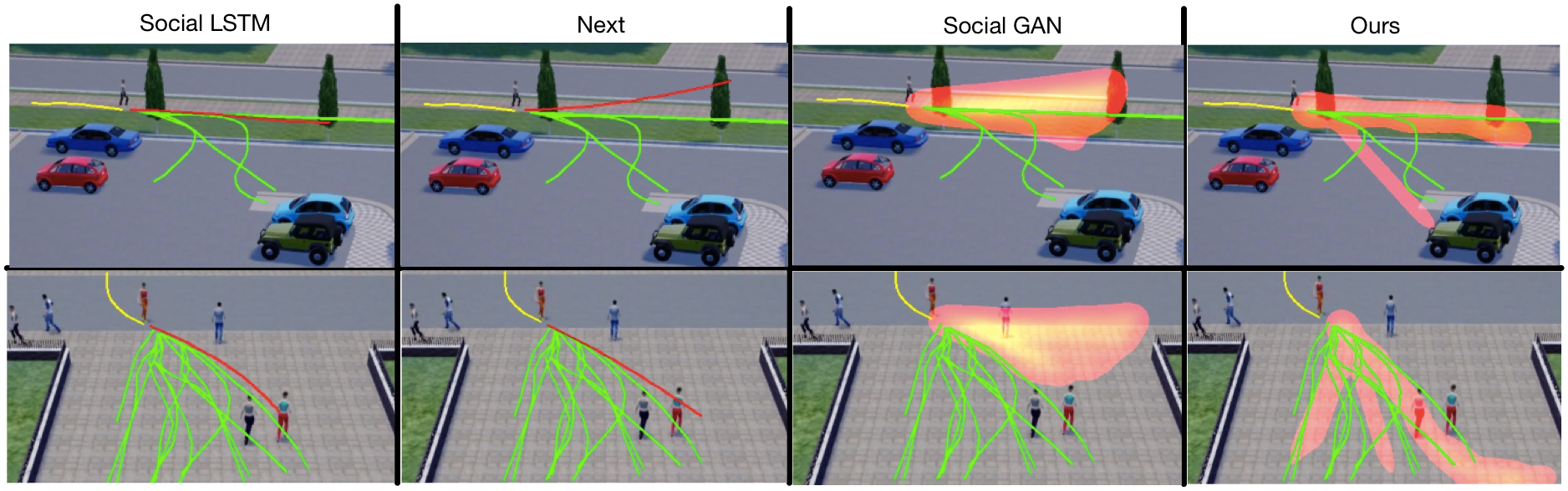}
	\caption{Qualitative analysis. The red trajectories are single-future method predictions and the yellow-orange heatmaps are multi-future method predictions. The yellow trajectories are observations and the green ones are ground truth multi-future trajectories. See text for details.}  
	\label{fig:multiverse_qual}
\end{figure}

\noindent\textbf{Qualitative analysis.} 
We visualize some outputs of the top 4 methods in Fig.~\ref{fig:multiverse_qual}.
In each image, the yellow trajectories are the history trajectory of each controlled agent (derived from real 
video data)
and the green trajectories are the ground truth
 future trajectories from human annotators.
The predicted trajectories are shown in yellow-orange heatmaps for multi-future prediction methods, and in red lines for single-future prediction methods. 
As we see, our model correctly generally
puts probability
mass where there is data, and does not ``waste''
probability mass where there is no data.

\begin{figure}[!t]
	\centering
		\includegraphics[width=0.7\textwidth]{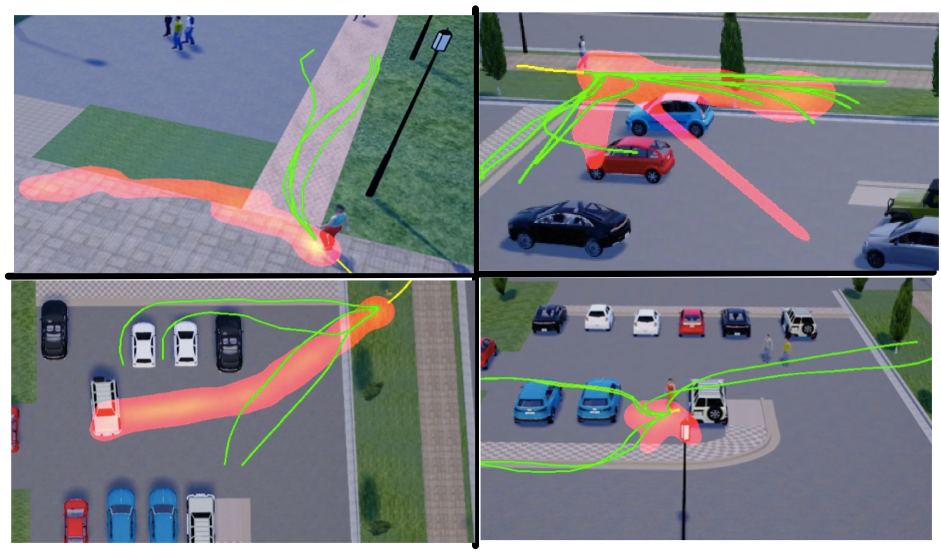}
	\caption{Error analysis. See text for details.}  
	\label{fig:multiverse_error_analysis}
\end{figure}

\noindent\textbf{Error analysis.}
We show some typical errors our model makes in Fig.~\ref{fig:multiverse_error_analysis}.
The first image shows our model misses the correct direction,
perhaps due to lack of diversity in our sampling procedure.
The second image shows our model 
sometimes predicts the person will ``go through'' the car
(diagonal red beam)
instead of going around it. This may be addressed by adding more training examples of ``going around'' obstacles.
The third image shows
our model predicts the person will go to a moving car.
This is due to the lack of modeling of the dynamics of other far-away agents in the scene. 
The fourth image shows a hard case where the person just exits the vehicle and there is no indication of where they will go next (so our model ``backs off'' to a sensible
``stay nearby'' prediction).
We leave solutions to these problems to future work.

\subsection{Single-Future Prediction on VIRAT/ActEV}
\label{sec:multiverse_exp-virat}

\noindent\textbf{Dataset \& Setups.} NIST released VIRAT/ActEV~\cite{2018trecvidawad} for activity detection research in streaming videos in 2018. 
This dataset is a new version of the VIRAT~\cite{oh2011large} dataset, with more videos and annotations. The length of videos with publicly available annotations is about 4.5 hours.
Following~\cite{liang2019peeking}, we use the official training set for training and the official validation set for testing. Other setups are the same as in Section~\ref{sec:multiverse_exp-multi},
except 
we use the single-future evaluation metric.

\noindent\textbf{Quantitative Results.} 
Table~\ref{tab:multiverse_exp-single} (first column)
shows the evaluation results. 
As we see, our model achieves state-of-the-art performance.
The improvement is especially large on Final Displacement Error (FDE) metric, attributing to the coarse location decoder that helps regulate the model prediction for long-term prediction.
The gain shows that our model does well at both
single future prediction (on real data)
and multiple future prediction on our quasi-synthetic data.

\noindent\textbf{Generalizing from simulation to real-world.} As described in the previous section, we generate simulation data first by reconstructing from real-world videos.
To verify the quality of the reconstructed data,
and the efficacy of learning from simulation videos, we train 
all the models on the simulation videos 
derived from the real data.
We then evaluate on the real test set of VIRAT/ActEV.
As we see from the right column in Table~\ref{tab:multiverse_exp-single},
all models do worse in this scenario,
due to the difference between synthetic and real data.
We find  the  performance  ranking  of  different  methods  are consistent between the real and our simulation training data.  This suggests the errors mainly coming from the model,  and  substantiates  the  rationality  of  using  the  proposed dataset to compare the relative performance of different methods.

There are two sources of error.
The synthetic trajectory data only contains about 60\%
of the real trajectory data, due to difficulties
reconstructing all the real data in the simulator.
In addition, the synthetic images are
not photo realistic.
Thus
methods (such as Next \cite{liang2019peeking})
that rely on RGB input obviously suffer the most,
since they have never been trained on ``real pixels''.
Our method, which uses trajectories plus high level
semantic segmentations (which transfers from synthetic
to real more easily)
suffers the least drop in performance,
showing its robustness to ``domain shift''. See Table~\ref{tab:multiverse_exp-multi} for input source comparison between methods.

\begin{table}[]
\centering
\small
\begin{tabular}{l||c||c}
\hline
Method & Trained on Real.   & Trained on Sim.        \\ \hline \hline
Linear &  32.19 / 60.92 & 48.65 / 90.84 \\ 
LSTM                &  23.98 / 44.97 & 28.45 / 53.01 \\ 
Social-LSTM \cite{alahi2016social}         &  23.10 / 44.27 & 26.72 / 51.26 \\ 
Social-GAN (V) \cite{gupta2018social}       &  30.40 / 61.93 & 36.74 / 73.22 \\ 
Social-GAN (PV) \cite{gupta2018social}         &  30.42 / 60.70 & 36.48 / 72.72 \\ 
Next ~\cite{liang2019peeking}   & 19.78 / 42.43 & 27.38 / 62.11 \\ 
 Ours & \textbf{18.51} / \textbf{35.84} & \textbf{22.94} / \textbf{43.35} \\ \hline
\end{tabular}
\caption{Comparison of different methods on the VIRAT/ActEV dataset.
We report ADE/FDE metrics.
First column  is for models trained on real video training set and
second column 
 is for models trained on the simulated version of this
 dataset.} 
\label{tab:multiverse_exp-single}
\end{table}

\subsection{Ablation Experiments}
\label{sec:multiverse_ablation}
We test various ablations of our model
on both the single-future and multi-future trajectory prediction to substantiate our design decisions.
Results are shown in Table~\ref{tab:multiverse_exp-ablation}, where the ADE/FDE metrics are shown in the ``single-future'' column and $\text{minADE}^M_{20}$/$\text{minFDE}^M_{20}$ metrics (averaged across all views) in the ``multi-future'' column. We verify three of our key designs by leaving the module out from the full model.

(1) \textit{Spatial Graph:} Our model is built on top of a spatial 2D graph that uses graph attention to model the scene features. We train model without the spatial graph.
As we see, the performance drops on both tasks.
(2) \textit{Fine location decoder:} We test our model without the fine location decoder and only use the grid center as the coordinate output.
As we see, the significant performance drops on both tasks verify the efficacy of this new module proposed in our study. 
(3) \textit{Multi-scale grid:} We utilize two different grid scales (36 $\times$ 18) and (18 $\times$ 9) in training.
We see that performance is slightly worse if we only use the fine scale (36 $\times$ 18) .

\begin{table}
\centering
\small
\begin{tabular}{l||c|c}
\hline
Method                             & Single-Future & Multi-Future \\ \hline
Our full model      &   18.51 / 35.84  &    166.1 / 329.5    \\ \hline
No spatial graph         &   28.68 / 49.87  & 184.5 / 363.2     \\ 
No fine location decoder                 &   53.62 / 83.57   &  232.1 / 468.6 \\ 
No multi-scale grid                       &  21.09 / 38.45     &    171.0 / 344.4     \\ \hline
\end{tabular}
\caption{Performance on ablated versions of our model on single and multi-future trajectory prediction. 
Lower numbers are better.}
\label{tab:multiverse_exp-ablation}
\end{table}

\section{Related Work}
This work falls under the category of sequential models that utilize both static and dynamic environmental cues in the human motion prediction literature~\cite{rudenko2020human}. 
In the following, we also review a few relevant recent approaches based on their outputs. Then we also review the trajectory prediction datasets.

\noindent\textbf{Single-future trajectory prediction.}
Recent works have tried to predict a single best trajectory for pedestrians or vehicles.
Early works~\cite{manh2018scene, xue2018ss,zhang2019sr} focused on modeling person motions by considering them as points in the scene.
These research works~\cite{kooij2014context,yagi2018future,ma2017forecasting,liang2019peeking} have attempted to predict person paths by utilizing visual features. 
Recently Liang et al. ~\cite{liang2019peeking} proposed a joint future activity and trajectory prediction framework that utilized multiple visual features using focal attention~\cite{liang2018focal,liang2019focal}.
Many works~\cite{lee2017desire,sadeghian2018car,bansal2018chauffeurnet,hong2019rules,zhao2019multi} in vehicle trajectory prediction have been proposed.
CAR-Net~\cite{sadeghian2018car} proposed attention networks on top of scene semantic CNN to predict vehicle trajectories.
Chauffeurnet~\cite{bansal2018chauffeurnet} utilized imitation learning for trajectory prediction.

\noindent\textbf{Multi-future trajectory prediction.}
Many works have tried to model the uncertainty of trajectory prediction.
Various papers (e.g. \cite{kitani2012activity,rhinehart2018r2p2,rhinehart2019precog} use Inverse Reinforcement Learning (IRL) to forecast human trajectories. 
Social-LSTM~\cite{alahi2016social} is a popular method using social pooling to predict future trajectories.
Other works~\cite{sadeghian2018sophie,gupta2018social,li2019way,amirian2019social} like Social-GAN~\cite{gupta2018social} have utilized generative adversarial networks~\cite{goodfellow2014generative} to generate diverse person trajectories.
In vehicle trajectory prediction, DESIRE~\cite{lee2017desire} utilized variational auto-encoders (VAE) to predict future vehicle trajectories.
Many recent works~\cite{thiede2019analyzing,chai2019multipath,tang2019multiple,makansi2019overcoming} also proposed probabilistic frameworks for multi-future vehicle trajectory prediction. 
Different from these previous works, we present a flexible two-stage framework that combines multi-modal distribution modeling and precise location prediction.

\noindent\textbf{Trajectory Prediction Datasets.}
Many vehicle trajectory datasets~\cite{caesar2019nuscenes,chang2019argoverse} have been proposed as a result of self-driving's surging popularity. 
With the recent advancement in 3D computer vision research~\cite{zhang2015fast,liang2017event,shah2018airsim,dosovitskiy2017carla,richter2016playing,ros2016synthia,heess2017emergence}, many research works~\cite{qiu2017unrealcv, gaidon2016virtual,de2017procedural,das2018embodied,wu2019revisiting,zhu2017target,sun2019stochastic} have looked into 3D simulated environment for its flexibility and ability to generate enormous amount of data.
We are the first to propose a 3D simulation dataset that is reconstructed from real-world scenarios complemented with a variety of human trajectory continuations for multi-future person trajectory prediction. 

\section{Summary}

In this chapter,
we have introduced the Forking Paths dataset, 
and the \textit{\fancyname} model for multi-future forecasting.
Our study is the first to provide a quantitative benchmark and evaluation methodology for 
multi-future trajectory prediction by using human annotators to create a variety
of trajectory continuations under the identical past.
Our model utilizes multi-scale location decoders with 
graph attention model to predict multiple future locations.
We have shown that our method achieves state-of-the-art performance on two challenging benchmarks: the large-scale real video dataset and our proposed multi-future trajectory dataset.
We believe our dataset, together with our models, will facilitate future research and applications on multi-future prediction.

\chapter{Learning from 3D Simulation 
}  \label{chap:0202_3d}

In this chapter, we explore the benefit of multi-camera-view video data from 3D simulation created in \autoref{chap:0201_multi} to train a robust trajectory prediction model that could perform fairly well on out-of-domain testing datasets.

\section{Motivation}

\begin{figure}[ht]
	\centering
		\includegraphics[width=\hsize]{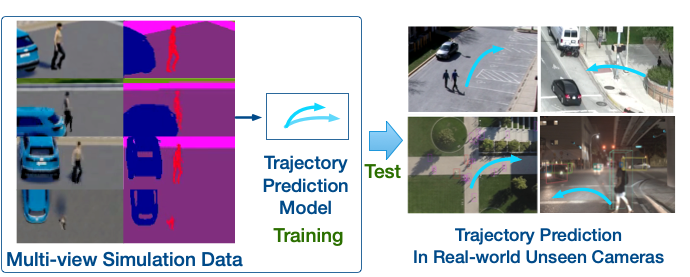}
	\caption{Illustration of pedestrian trajectory prediction in unseen cameras. We propose to learn robust representations only from 3D simulation data that could generalize to real-world videos captured by unseen cameras.}
	\label{fig:simaug_intro}
\end{figure}

Future trajectory prediction~\cite{kitani2012activity,alahi2016social,gupta2018social,liang2019peeking,sadeghian2018sophie,lee2017desire,liang2020garden} is a fundamental problem in video analytics, which aims at forecasting a pedestrian's future path in the video in the next few seconds. Recent advancements in future trajectory prediction have been successful in a variety of vision applications such as
self-driving vehicles~\cite{bansal2018chauffeurnet,chai2019multipath,chang2019argoverse}, safety monitoring~\cite{liang2019peeking}, robotic planning~\cite{rhinehart2017first,rhinehart2018r2p2}, among others.

A notable bottleneck for existing works is that the current model is closely coupled with the video cameras on which it is trained, and generalizes poorly on new cameras with novel views or scenes. 
For example, prior works have proposed various models to forecast a pedestrian's trajectories in video cameras of different types such as stationary outdoor cameras~\cite{oh2011large,liang2019focal,alahi2016social,gupta2018social,lerner2007crowds,luber2010people}, drone cameras~\cite{sadeghian2018sophie,deo2020trajectory,li2019way}, ground-level egocentric cameras~\cite{yagi2018future,rhinehart2017first,styles2019multiple}, or dash cameras~\cite{mangalam2019disentangling,styles2019forecasting,chang2019argoverse}. 
However, existing models are all separately trained and tested within one or two datasets, and there have been no attempts at successfully generalizing the model across datasets of novel camera views.
This bottleneck significantly hinders the application whenever there is a new camera because it requires annotating new data to fine-tune the model, resulting in a procedure that is not only expensive but also tardy in deploying the model.

An ideal model should be able to disentangle human behavioral dynamics from specific camera views, positions, and scenes. It should produce robust trajectory prediction despite the variances in these camera settings. 
Motivated by this idea, in this work, we learn a robust representation for future trajectory prediction that can generalize to unseen video cameras. Different from existing works, we study a \emph{real-data-free} setting where a model is trained only on synthetic data but tested, out of the box, on unseen real-world videos, without further re-training or fine-tuning the model. 
Following the success of learning from simulation~\cite{ruiz2018learning,de2017procedural,varol2019synthetic,zhang2019rsa,gaidon2016virtual,richter2016playing}, our synthetic data is generalized from a 3D simulator, called CARLA~\cite{dosovitskiy2017carla}, which anchors to the static scene and dynamic elements in the VIRAT/ActEV videos~\cite{oh2011large}. By virtue of the 3D simulator, we can generate multiple views and pixel-precise semantic segmentation labels for each training trajectory, 
as illustrated in Figure~\ref{fig:simaug_intro}.
Meanwhile, following the previous works~\cite{sadeghian2018sophie,liang2020garden}, scene semantic segmentation is used instead of RGB pixels to alleviate the influence of different lighting conditions, scene textures, subtle noises produced by camera sensors, etc. At test time, we extract scene features from real videos using pretrained segmentation models.
The use of segmentation features is helpful but is insufficient for learning robust representation for future trajectory prediction.

To tackle this issue, we propose a novel data augmentation method called \emph{SimAug} to augment the features of the simulation data with the goal of learning robust representation to various semantic scenes and camera views in real videos. 
To be specific, first, after representing each training trajectory by high-level scene semantic segmentation features, we defend our model from adversarial examples generated by white-box attack methods~\cite{goodfellow2014explaining}. 
Second, to overcome the changes in camera views, we generate multiple views for the same trajectory, and encourage the model to focus on overcoming the ``hardest'' view to which the model has learned. Following~\cite{jiang2018mentornet,jiang2019synthetic}, the classification loss is adopted and the view with the highest loss is favored during training. 
Finally, the augmented trajectory is computed as a convex combination of the trajectories generated in previous steps.
Our trajectory prediction backbone model is built on a recent work called Multiverse~\cite{liang2020garden}. The final model is trained to minimize the empirical vicinal risk over the distribution of augmented trajectories.
Our method is partially inspired by recent robust deep learning methods using adversarial training~\cite{kurakin2016adversarial,cheng2020advaug}, Mixup~\cite{zhang2017mixup}, and MentorMix~\cite{jiang2019synthetic}.

We empirically validate our model, which is trained only on simulation data, 
on three real-world benchmarks for future trajectory prediction: VIRAT/ActEV \cite{oh2011large,2018trecvidawad}, Stanford Drone~\cite{robicquet2016learning}, and Argoverse~\cite{chang2019argoverse}. 
These benchmarks represent three distinct camera views: 45-degree view, top-down view and dashboard camera view with ego-motions.
The results show our method performs favorably against baseline methods including standard data augmentation, adversarial learning, and imitation learning. 
Notably, our method achieves better results compared to the state-of-the-art on the VIRAT/ActEV and Stanford Drone benchmark. 
Our code and models are released at \url{https://next.cs.cmu.edu/simaug}. To summarize, our contribution is threefold:
\begin{itemize}
    \item We study a new setting of future trajectory prediction in which the model is trained only on synthetic data and tested, out of the box, on any unseen real video with novel views or scenes.
    \item We propose a novel and effective approach to augment the representation of trajectory prediction models using multi-view simulation data.
    \item Ours is the first work on future trajectory prediction to demonstrate the efficacy of training on 3D simulation data, and establishes new state-of-the-art results on three public benchmarks.
\end{itemize}

\section{The \textit{SimAug} Model}

\begin{figure}[ht]
	\centering
		\includegraphics[width=\textwidth]{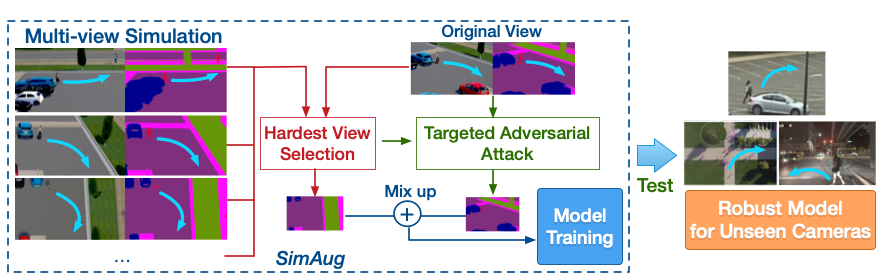}
	\caption{Overview of our method \emph{SimAug} that is trained on simulation and tested on real unseen videos. Each training trajectory is represented by multi-view segmentation features extracted from the simulator. \emph{SimAug} mixes the feature of the hardest camera view with the adversarial feature of the original view.}
	\label{fig:simaug_method}
\end{figure}

\begin{figure}[]
	\centering
		\includegraphics[width=\textwidth]{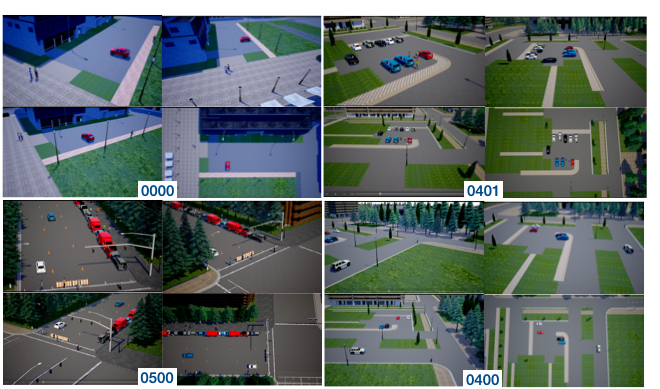}
	\caption{Visualization of the multi-view 3D simulation data used in \textit{SimAug} training. Data generation process is described in Section~\ref{sec:simaug_data}. We use 4 camera views from 4 scenes defined in~\cite{liang2020garden}. ``0400'' and ``0401'' scene have overlapping views. The top-left views are the original views from VIRAT/ActEV dataset.}
	\label{fig:simaug_data}
\end{figure}

\section{Approach}
In this section, we describe our approach to learn robust representation for future trajectory prediction, which we call \textit{SimAug}. Our goal is to train a model only on simulation training data that can effectively predict the future trajectory in the real-world test videos that are unseen during training.

\subsection{Problem Formulation}
We focus on predicting the locations of a single agent for multiple steps into the future. Given a sequence of historical video frames $V_{1:h}$ of the past $h$ steps and the past agent locations $L_{1:h}$ in training, we learn a probabilistic model on simulation data to estimate $P(L_{h+1:T}|L_{1:h}, V_{1:h})$ for $T-h$ steps into the future. At test time, our model takes as input an agent's observable past $(V_{1:h}, L_{1:h})$ in real videos to predict the agent's future locations $L_{h+1:T} = \{y_{h+1}, \ldots, y_T\}$, where $y_t$ is the location coordinates. As the test real videos are unseen during training, the model is supposed to be invariant to the variances in semantic scenes, camera views, and camera motions.

\subsection{Training Data Generation From Simulation}
\label{sec:simaug_data}
Our model is trained only on simulation data. To ensure high-level realism, the training trajectories are generated by CARLA~\cite{dosovitskiy2017carla}, an open-source 3D simulator built on top of the state-of-the-art game engine \textit{Unreal Engine 4}. We use the trajectories from the Forking Paths dataset~\cite{liang2020garden} that are semi-manually recreated from the VIRAT/ActEV benchmark that projects real-world annotations to the 3D simulation world. Note that it is not our intention to build an exact replica of the real-world scene, nor it is necessary to help train a model for real-world tasks as suggested in previous works~\cite{gaidon2016virtual,ros2016synthia,liang2020garden,zhang2019rsa}.

With CARLA, we record multiple views of the same trajectory of different camera angles and positions. For a trajectory $(V_{1:T}, L_{1:T})$ in original view, let $\mathcal{S} = \{(V_{1:T}^{(i)}, L_{1:T}^{(i)})\}_{i=1}^{|\mathcal{S}|}$ denote a set of additional views for the same trajectory.
In our experiments, we use four camera parameters pre-specified in~\cite{liang2020garden}, including three 45-degree views and one top-down view. 
We use a total of 4 scenes shown in Fig.~\ref{fig:simaug_data}. 
The ground-truth location varies under different camera views i.e. $L_{1:T}^{(i)} \ne L_{1:T}^{(j)}$ for $i \ne j$.
Note that these camera positions and angles are defined in~\cite{liang2020garden} specifically for VIRAT/ActEV dataset.
The top-down view cameras in Stanford Drone dataset~\cite{robicquet2016learning} are still considered unseen to the model since the scenes and camera positions are quite different.

In simulation, we also collect the ground-truth scene semantic segmentation for $K=13$ classes including sidewalk, road, vehicle, pedestrian, etc. At test time, we extract the semantic segmentation feature from real videos using a pre-trained model with the same number of class labels per pixel. To be specific, we use the Deeplab model \cite{chen2017deeplab} trained on the ADE20k \cite{zhou2017scene} dataset and keep its weights frozen. To bridge the gap between real and simulated video frames, we represent all trajectory $V_{1:T}$ as a sequence of scene semantic segmentation features, following previous works~\cite{liang2019peeking,liang2020garden,deo2020trajectory,sadeghian2018sophie}. As we show in our experiments, the use of segmentation features is helpful but is still insufficient for learning the robust representation.

\subsection{Multi-view Simulation Augmentation (\textit{SimAug})}
\label{sec:simaug}
In this subsection, we first describe \textit{SimAug} for learning robust representations. Our trajectory prediction backbone model is built on the Multiverse model~\cite{liang2020garden} and will be discussed in Section~\ref{sec:simaug_multiverse}.

Given a trajectory in its original view $(V_{1:T}, L_{1:T})$, we generate a set of additional views in $\mathcal{S} = \{(V_{1:T}^{(i)}, L_{1:T}^{(i)})\}_{i=1}^{|\mathcal{S}|}$ as described in the previous section, where $V_{t}^{(i)}$ represents the scene semantic feature of view $i$ at time $t$. $L_{1:T}^{(i)}$ is a sequence of ground-truth locations for the $i$-th view. 

Each time given a camera view, we use it as an anchor to search for the ``hardest'' view that is most inconsistent with what the model has learned. 
Inspired by~\cite{jiang2018mentornet}, we use the classification loss as the criteria and compute:
\begin{align}
    \label{eqn:simaug_select_c}
    j^* = \argmax_{j \in [1,|\mathcal{S}|]} \{ 
    \mathcal{L}_\text{cls}(V_{1:h} + \delta, L_{h+1:T}^{(j)}) \},
\end{align}
where $\delta$ is the $\ell_{\infty}$-bounded random perturbation applied to the input features.  $\mathcal{L}_\text{cls}$ is the location classification loss used in our backbone Multiverse model and will be discussed in the next subsection.

Then for the original view, we generate an adversarial trajectory by the targeted-FGSM attack~\cite{kurakin2016adversarial}:
\begin{align}
    V_{1:h}^{adv} &= V_{1:h} - \epsilon  \cdot \text{sign}(\nabla_{V_{1:h}} \mathcal{L}_\text{cls}( V_{1:h} + \delta, L_{h+1:T}^{(j^*)})),
    \label{eqn:simaug_adv}
\end{align}
where $\epsilon$ is the hyper-parameter. 
The attack tries to make the model predict the future locations in the selected ``hardest'' camera view rather than the original view.
In essence, the resulting adversarial feature is ``warped'' to the ``hardest'' camera view by a small perturbation.
By defending against such adversarial trajectory, our model learns representations that are robust against variances in camera views.  

Finally, we mix up the trajectory locations of the selected view and the adversarial trajectory locations by a convex combination function~\cite{zhang2017mixup} over their features and one-hot location labels.
\begin{align}
\begin{split}
    V_{1:h}^{aug} &= \lambda \cdot V_{1:h}^{adv} + (1-\lambda) \cdot V_{1:h}^{(j^*)} \\
    y_t^{aug} &= \lambda \cdot \text{one-hot}(y_t) + (1-\lambda) \cdot\text{one-hot}(y_t^{(j^*)}) \quad \forall t \in [h+1,T] \\
    L_{h+1:T}^{aug} &= [y_{h+1}^{aug}, \ldots, y_T^{aug}]
    \label{eqn:simaug_mixup}
\end{split}
\end{align}
where $[y_{h+1}, \cdots, y_T]=L_{h+1:T}$ are the ground-truth locations of the original view. The $\text{one-hot}(\cdot)$ function projects the location in $xy$ coordinates into an one-hot embedding over the predefined grid used in our backbone trajectory prediction model. Please find the details in~\cite{liang2020garden}.
Following~\cite{zhang2017mixup}, $\lambda$ is drawn from a Beta distribution $\text{Beta}(\alpha, \alpha)$ controlled by the hyper-parameter $\alpha$.

The algorithm for training with one training step is listed in Algorithm~\ref{alg:simaug}.
To train robust models to various camera views and semantic scenes, we learn representations over augmented training trajectories to overcome (i) feature perturbations (Step 3 and 5) (ii) targeted adversarial attack (Step 5), and (iii) the ``hardest'' feature from other views (Step 4).
By the mix-up operation in Eq.~\eqref{eqn:simaug_mixup}, 
our model is trained to minimize the empirical vicinal risk over a new distribution constituted by the generated augmented trajectories, which is proved to be useful in improving model robustness to real-world distributions under various settings~\cite{jiang2019synthetic}.

\begin{algorithm}
\setcounter{AlgoLine}{0}
\SetKwData{Left}{left}\SetKwData{This}{this}\SetKwData{Up}{up}
\SetKwInOut{Input}{Input}\SetKwInOut{Output}{Output}
\LinesNumbered
\Input{Mini-batch of trajectories; hyper-parameters $\alpha$ and $\epsilon$}
\Output{Classification loss $\mathcal{L}_\text{cls}$ computed over augmented trajectories}
\BlankLine
\For{each trajectory $(V_{1:T}, L_{1:T})$ in the mini-batch}{
    Generate trajectories from additional views $\mathcal{S}= \{(V_{1:T}^{(i)}, L_{1:T}^{(i)})\}$\;
    
    Compute the loss for each camera view $\mathcal{L}_\text{cls}( V_{1:h} + \delta, L_{h+1:T}^{(j)})$\;
    
    Select the view with the largest loss $j^*$ by Eq.~\eqref{eqn:simaug_select_c} \;
    
    Generate an adversarial trajectory $V_{1:h}^{adv}$ by Eq.~\eqref{eqn:simaug_adv}\;
    
    Mix up ($V_{1:h}^{adv}$, $L_{h+1:T}$) and ($V_{1:h}^{(j^*)}$, $L_{h+1:T}^{(j*)}$) by Eq.~\eqref{eqn:simaug_mixup}\;
    
    Compute $\mathcal{L}_\text{cls}$ over the augmented trajectory ($V^{aug}_{1:h}$, $L_{h+1:T}^{aug}$) from Step 6\;
}
\Return  averaged $\mathcal{L}_\text{cls}$ over the augmented trajectories
\caption{\small Multi-view Simulation Adversarial Augmentation (\textit{SimAug})}
\label{alg:simaug}
\end{algorithm}

\subsection{Backbone Model for Trajectory Prediction}
\label{sec:simaug_multiverse}

We employ Multiverse~\cite{liang2020garden} as our backbone network, a state-of-the-art multi-future trajectory prediction model. Although we showcase the use of \emph{SimAug} to improve the robustness of Multiverse, \emph{SimAug} is a general approach that can be applied to other trajectory prediction models.

\noindent\textbf{Input Features.}
The model is given the past locations, $L_{1:h}$, and the scene, $V_{1:h}$. 
Each ground-truth location $L_t$ is encoded by an one-hot vector $y_{t} \in \mathbb{R}^{HW}$ representing the nearest cell in a 2D grid of size $H \times W$.
In our experiment, we use a grid scale of $36 \times 18$.
Each video frame $V_t$ is encoded as semantic segmentation feature of size $H \times W \times K$ where $K=13$ is the total number of class labels as in~\cite{liang2020garden,liang2019peeking}.
As discussed in the previous section, we use \textit{SimAug} to generate augmented trajectories $(V_{1:h}^{aug}, L_{1:h}^{aug})$ as our training features.

\noindent\textbf{History Encoder.} 
A convolutional RNN~\cite{xingjian2015convolutional,wang2019eidetic} is used to get the final spatial-temporal feature state $H_t \in \mathbb{R}^{H \times W \times d_{enc}}$, where $d_{enc}$ is the hidden size.
The context is a concatenation of the last hidden state and the historical video frames, $\context=[H_h,V_{1:h}]$.

\noindent\textbf{Location Decoder.}
After getting the context $\context$, a coarse location decoder is used to predict locations at the level of grid cells at each time-instant by:
\begin{align}
    \hat{y}_{t} = \text{softmax}(f_c(\context, H^c_{t-1})) \in \mathbb{R}^{HW}
    \label{eqn:simaug_fc}
\end{align}
where $f_c$ is the convolutional recurrent neural network (ConvRNN) with graph attention proposed in~\cite{liang2020garden} and $H^c_t$ is the hidden state of the ConvRNN.
Then a fine location decoder is used to predict a continuous offset in $\real^2$, which specifies a ``delta''
from the center of each grid cell, to get a fine-grained location prediction:
\begin{align}
    \hat{O}_t = \text{MLP}(f_o(\context, H^o_{t-1})) \in \mathbb{R}^{HW \times 2},
    \label{eqn:simaug_fo}
\end{align}
where $f_o$ is a separate ConvRNN and $H^o_t$ is its hidden state.
To compute the final prediction location, we use
\begin{align}
    \hat{L}_t = Q_{g} + \hat{O}_{tg}
\end{align}
where $g=\argmax \hat{y}_{t}$ is the index of the selected grid cell, $Q_{g} \in \real^2$ is the center of that cell,
and $\hat{O}_{tg} \in \real^2$ is the predicted offset for that cell at time $t$.

\noindent\textbf{Training.} 
We use \textit{SimAug} (see Section~\ref{sec:simaug}) to generate $L_{h+1:T}^{aug}= \{y_{h+1}^{aug}, \ldots, y_T^{aug} \}$ as labels for training.
For the coarse decoder, the cross-entropy loss is used:
\begin{equation}
    \mathcal{L}_\text{cls} = -\frac{1}{T} \sum_{t=h+1}^{T} 
    \sum^{HW}_{c=1} y^{aug}_{tc} \log(\hat{y}_{tc})
    \label{eqn:simaug_loss_cls}
\end{equation}
For the fine decoder, we use the original ground-truth location label $L_{h+1:T}$:
\begin{equation}
 \mathcal{L}_\text{reg} = \frac{1}{T} \sum_{t=h+1}^{T} 
 \sum^{HW}_{c=1}
 \text{smooth}_{l_1}(O_{tc}, \hat{O}_{tc})
\end{equation}
where 
$O_{tc} = L_t - Q_{c}$ 
is the delta between the ground true location and the center of the $c$-th grid cell. 
The final loss is then calculated using
\begin{align}
\mathcal{L}(\theta) = \mathcal{L}_\text{cls} + \lambda_1 \mathcal{L}_\text{reg} + \lambda_2 \|\theta\|_2^2
\end{align}
where $\lambda_2$ controls the $\ell_2$ regularization (weight decay),
and $\lambda_1=0.5$ is used
to balance the regression and classification losses.

\section{Experiments}
\label{sec:simaug_exp}

In this section, we evaluate various methods, including our \textit{SimAug} method, on three public video benchmarks of real-world videos captured under different camera views and scenes: the VIRAT/ActEV~\cite{2018trecvidawad,oh2011large} dataset, the Stanford Drone Dataset (SDD)~\cite{robicquet2016learning}, and the autonomous driving dataset Argoverse~\cite{chang2019argoverse}.
We demonstrate the efficacy of our method for unseen cameras in Section~\ref{sec:simaug_exp_main} and how our method can also improve state-of-the-art when fine-tuned on the real training data in Section~\ref{sec:simaug_exp_sdd} and Section~\ref{sec:simaug_exp_actev}.

\subsection{Evaluation Metrics}
\label{sec:simaug_metrics}
Following prior works~\cite{alahi2016social,liang2020garden}, we utilize two common metrics for trajectory prediction evaluation. Let $L^{i}=L^{i}_{t=(h+1)\cdots T}$ be the true future trajectory for the $i^{th}$ test sample, and $\hat{L}^{ik}$ be the corresponding $k^{th}$ prediction sample, for $k \in [1,K]$.

\noindent i) \textit{Minimum Average Displacement Error Given K Predictions} (minADE\textsubscript{K}): for each true trajectory sample $i$,
we select the closest $K$ predictions,
and then measure its average error:
\begin{equation}
    \text{minADE}_K = \frac{  \sum^{N}_{i=1} \min_{k=1}^K \sum^{T}_{t=h+1} \lVert L^i_t - \hat{L}^{ik}_t \rVert_{2}}{N \times (T-h)}
\end{equation}

\noindent ii) \textit{Minimum Final Displacement Error Given K Predictions} (minFDE\textsubscript{K}): similar to minADE\textsubscript{K}, but we only consider the predicted points and the ground truth point at the final prediction time instant:
\begin{equation}
    \text{minFDE}_K = \frac{  \sum^{N}_{i=1} \min_{k=1}^K \lVert L^{i}_{T} -  \hat{L}^{ik}_{T} \rVert_{2}}{N}
\end{equation}
\noindent iii) \textit{Grid Prediction Accuracy} (Grid\_Acc):
As our base model also predicts coarse grid locations as described in Section~\ref{sec:simaug_multiverse}, we also evaluate the accuracy between the predicted grid $\hat{y}_t$ and the ground truth grid $y_t$. This is an intermediate metric and hence is less indicative than the minADE\textsubscript{K} and minFDE\textsubscript{K}.

\subsection{Main Results}
\label{sec:simaug_exp_main}
\noindent\textbf{Dataset \& Setups.} 
We compare \textit{SimAug} with classical data augmentation methods as well as adversarial learning methods to train robust representations.
All methods are trained using the same backbone on the same \textit{simulation training data} described in Section~\ref{sec:simaug_data}, and tested on the same benchmarks. Real videos are not allowed to be used during training except in our finetuning experiments. For VIRAT/ActEV and SDD, we use the standard test split as in \cite{liang2019peeking,liang2020garden} and \cite{sadeghian2018sophie,deo2020trajectory}, respectively.
For Argoverse, we use the official validation set from the 3D tracking task, and the videos from the ``ring\_front\_center'' camera are used.

These datasets have different levels of difficulties. VIRAT/ActEV is the easiest one because its training trajectories have been projected in the simulation training data. SDD is more difficult as its camera positions and scenes are different from the training data. Argoverse is the most challenging one with distinct scenes, camera views, and ego-motions.


Following the setting in previous works~\cite{liang2019peeking,alahi2016social,gupta2018social,alahi2016social,gupta2018social,sadeghian2018sophie,makansi2019overcoming,liang2020garden,deo2020trajectory}, 
the models observe 3.2 seconds (8 frames) of every pedestrian and predict the future 4.8 seconds (12 frames) of the person trajectory. 
We use the pixel values for the trajectory coordinates as it is done in~\cite{yagi2018future,liang2019peeking,lee2017desire,chai2019multipath,li2019way,makansi2019overcoming,bansal2018chauffeurnet,hong2019rules,deo2020trajectory}. 
By default, we evaluate the top $K=1$ future trajectory prediction of all models.

\noindent\textbf{Baseline methods.} 
We compare \textit{SimAug} with the following baseline methods for learning robust representations. All methods are built on the same backbone network and trained using the same simulation training data.
\textit{Base Model} is the trajectory prediction model proposed in~\cite{liang2020garden}.
\textit{Standard Aug} is the base model trained with standard data augmentation techniques including horizontal flipping and random input jittering. 
\textit{Fast Gradient Sign Method (FGSM)} is the base model trained with adversarial examples generated by the targeted-FGSM attack method~\cite{goodfellow2014explaining}. Random labels are used for the targeted-FGSM attack. 
\textit{Projected Gradient Descent (PGD)} is learned with an iterative adversarial learning method~\cite{madry2017towards,xie2019feature}. The number of iterations is set to 10 and other hyper-parameters following~\cite{xie2019feature}.

\noindent\textbf{Implementation Details.}
We use $\alpha=0.2$ for the $\text{Beta}$ distribution in Eq~\eqref{eqn:simaug_mixup} and we use $\epsilon=\delta=0.1$ in Eq~\eqref{eqn:simaug_adv}.
As the random perturbation is small, we do not normalize the perturbed features and the normalized features yield comparable results.
All models are trained using Adadelta optimizer~\cite{zeiler2012adadelta} with an initial learning rate of 0.3 and a weight decay of 0.001.
Other hyper-parameters for the baselines are the same as the ones in~\cite{liang2020garden}.

\noindent\textbf{Quantitative Results.}
Table~\ref{tab:simaug_main} shows the evaluation results. 
Our method performs favorably against other baseline methods across all evaluation metrics on all three benchmarks. 
In particular, ``Standard Aug'' seems to be not generalizing well to unseen cameras.
FGSM improves significantly on the ``Grid\_Acc'' metric but fails to translate the improvement to final location predictions. 
\textit{SimAug} is able to improve the model overall stemming from the effective use of multi-view simulation data.
All other methods are unable to improve trajectory prediction on Argoverse, whose data characteristics include ego-motions and distinct dashboard-view cameras.
The results substantiate the efficacy of \textit{SimAug} for future trajectory prediction in unseen cameras. Note as the baseline methods use the same features as ours, the results indicate the use of segmentation features is insufficient for learning robust representations.

\begin{figure}[ht]
	\centering
		\includegraphics[width=\textwidth]{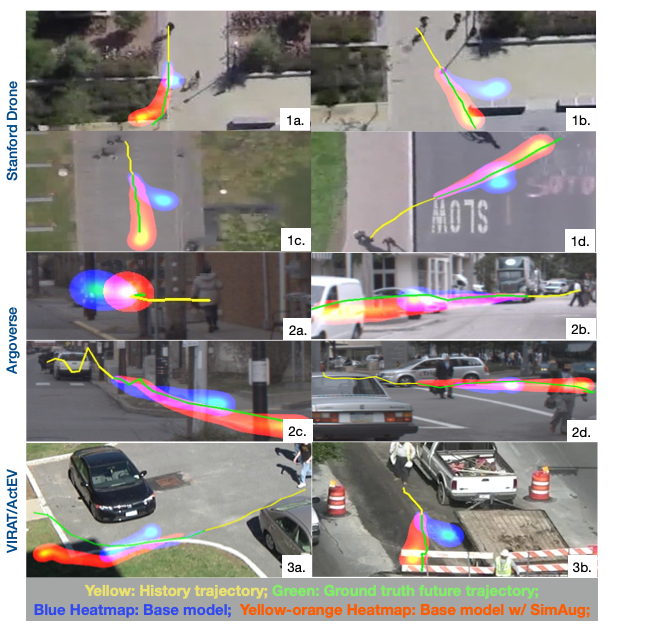}
	\caption{Qualitative analysis. Trajectory predictions from different models are colored and overlaid in the same image. See text for details. }
	\label{fig:simaug_qualitative}
\end{figure}

\noindent\textbf{Qualitative Analysis.}
We visualize outputs of the base model with and without \textit{SimAug} in Fig.~\ref{fig:simaug_qualitative}. We show visualizations on all three datasets. 
In each image, the yellow trajectories denote historical trajectories and the green ones are ground truth future trajectories. Outputs of the base model without {SimAug} are colored with blue heatmaps and the yellow-orange heatmaps are from the same model with {SimAug}.
As we see, the base model with {SimAug} augmentation yields more accurate trajectories for turnings (Fig.~\ref{fig:simaug_qualitative} 1a., 3a.) while without it the model sometimes predicts the wrong turns (Fig.~\ref{fig:simaug_qualitative} 1b., 1c., 2a., 3a., 3b.).
In addition, the length of {SimAug} predictions is more accurate (Fig.~\ref{fig:simaug_qualitative} 1d., 2b., 2c., 2d.).

\begin{table}[]
\centering
\caption{Comparison to the standard data augmentation method and recent adversarial learning methods on three datasets. We report three metrics: Grid\_Acc($\uparrow$)/minADE\textsubscript{1}($\downarrow$)/minFDE\textsubscript{1}($\downarrow$). The units of ADE/FDE are pixels. All methods are built on the same backbone model in~\cite{liang2020garden} and trained using the same multi-view simulation data described in Section~\ref{sec:simaug_data}.}
\setlength\tabcolsep{2mm}
\begin{tabular}{lccc}
\toprule
Method & VIRAT/ActEV & Stanford Drone  & Argoverse    \\ 
\midrule
Base Model~\cite{liang2020garden} & 44.2\%/26.2/49.7 & 31.4\%/21.9/42.8  &  26.6\%/69.1/183.9 \\ 
Standard Aug &   45.5\%/25.8/48.3 & 21.3\%/23.7/47.6  &  28.9\%/70.9/183.4 \\ 
PGD~\cite{madry2017towards,xie2019feature} &  47.5\%/25.1/48.4 & 28.5\%/21.0/42.2 & 25.9\%/72.8/184.0 \\
FGSM~\cite{goodfellow2014explaining} &  48.6\%/25.4/49.3 & 42.3\%/19.3/39.9 & 29.2\%/71.1/185.4\\
SimAug &  \textbf{51.1}\%/\textbf{21.7}/\textbf{42.2} &  \textbf{45.4\%}/\textbf{15.7}/\textbf{30.2} & \textbf{30.9\%}/\textbf{67.9}/\textbf{175.6}\\
\bottomrule
\end{tabular}
\label{tab:simaug_main}
\end{table}

\subsection{State-of-the-Art Comparison on Stanford Drone Dataset}
\label{sec:simaug_exp_sdd}
In this section, we compare our \textit{SimAug} model with the state-of-the-art generative models, including Social-LSTM~\cite{alahi2016social}, Social-GAN~\cite{gupta2018social}, DESIRE~\cite{lee2017desire}, and SoPhie~\cite{sadeghian2018sophie}. We also compare with the imitation learning model, IDL~\cite{li2019way}, and the inverse reinforcement learning model, P2T\textsubscript{IRL}~\cite{deo2020trajectory} for trajectory prediction on the Stanford Drone Dataset. Following previous works, we evaluate the minimal errors over $K=20$ predictions.

\noindent\textbf{Results \& Analysis.} The results are shown in Table~\ref{tab:simaug_sdd}, where \textit{SimAug} is built on top of the \textit{Multiverse} model. 
As it shows, \textit{SimAug} model trained only on the simulation data (second to the last row) achieves comparable or even better performance than other state-of-the-art models that are trained on in-domain real videos. By further fine-tuning on the learned representations of \textit{SimAug}, we achieve the state-of-the-art performance on the Stanford Drone Dataset. 
The promising results demonstrate the efficacy of \textit{SimAug} for future trajectory prediction in unseen cameras.

\begin{table}[]

\centering
\caption{State-of-the-art comparison on the VIRAT/ActEV dataset. Numbers are minimal errors over 1 predictions and lower the better. 
}
\begin{tabular}{lcc}
\toprule
Method & minADE\textsubscript{1}($\downarrow$)   & minFDE\textsubscript{1} ($\downarrow$)      \\ 
\midrule
Social-LSTM~\cite{alahi2016social}  & 23.10 & 44.27 \\
Social-GAN~\cite{gupta2018social}  & 30.42   & 60.70  \\ 
Next~\cite{liang2019peeking}  & 19.78 & 42.43\\
Multiverse~\cite{liang2020garden} & 18.51 & 35.84 \\
\midrule
Multiverse (Trained on Sim.)~\cite{liang2020garden} & 22.94 & 43.35 \\
SimAug & 21.73  & 42.22 \\
SimAug + finetune  & \textbf{17.96}  & \textbf{34.68} \\
\bottomrule
\end{tabular}
\label{tab:simaug_actev}
\end{table}

\begin{table}[]
\centering
\caption{State-of-the-art comparison on the Stanford Drone Dataset (SDD). Numbers are minimal errors over 20 predictions and lower the better. Baseline numbers are taken from \cite{sadeghian2018sophie,deo2020trajectory}. ``SimAug'' is trained without using SDD training data and ``SimAug + finetune'' is further finetuned on SDD training data.}
\begin{tabular}{lcc}
\toprule
Method & minADE\textsubscript{20}($\downarrow$)   & minFDE\textsubscript{20} ($\downarrow$)      \\
\midrule
Social-LSTM~\cite{alahi2016social} & 31.19 & 56.97 \\
Social-GAN~\cite{gupta2018social} & 27.25   &  41.44 \\ 
DESIRE~\cite{lee2017desire} &  19.25 & 34.05 \\ 
SoPhie~\cite{sadeghian2018sophie} &  16.27 & 29.38 \\ 
Multiverse~\cite{liang2020garden} & 14.78 & 27.09 \\
IDL~\cite{li2019way} &  13.93 & 24.40 \\ 
P2T\textsubscript{IRL}~\cite{deo2020trajectory} &  12.58 & 22.07 \\ 
\midrule
SimAug & 12.03  & 23.98 \\
SimAug + finetune & \textbf{10.27}  & \textbf{19.71} \\
\bottomrule
\end{tabular}
\label{tab:simaug_sdd}
\end{table}

\subsection{State-of-the-Art Comparison on VIRAT/ActEV}
\label{sec:simaug_exp_actev}
In this section, we compare our \textit{SimAug} model with state-of-the-art models on VIRAT/ActEV.
Following the previous work~\cite{liang2020garden}, we compute the errors for the top $K=1$ prediction. 
Experimental results are shown in Table~\ref{tab:simaug_actev} (b), where all models in the top four rows are trained on the real-world training videos in VIRAT/ActEV. 
Our model trained on simulation data achieves competitive performance and outperforms \textit{Multiverse}~\cite{liang2020garden} model that is trained on the same data.
With fine-tuning, which means using exactly the same training data without any extra annotation of real trajectories compared to ~\cite{alahi2016social,gupta2018social,liang2019peeking,liang2020garden}, we achieve the best performance on the VIRAT/ActEV benchmark.



\subsection{Ablation Experiments}

We test various ablations of our approach to validate our design decisions.
Results are shown in Table~\ref{tab:simaug_ablation}, where the top-1 prediction is used in the evaluation.
We verify four key design choices by removing each, at a time, from the full model.
The results show that by introducing viewpoint selection (Eq.~\eqref{eqn:simaug_select_c}) and adversarial perturbation (Eq.~\eqref{eqn:simaug_adv}), our method improves model generalization. 

(1) \textit{Multi-view data:} Our method is trained on multi-view simulation data and we use 4 camera views in our experiments. We test our method without the top-down view because it is similar to the ones in the SDD dataset.
As we see, the performance drops due to the fewer number of data and less diverse views, suggesting that we should use more views in augmentation (which is effortless to do in 3D simulators).

(2) \textit{Random perturbation:} We test our model without random perturbation on the original view trajectory samples by setting $\delta=0$ in Eq.~\eqref{eqn:simaug_select_c}.
This leads to the performance drop on all three datasets and particularly on the more difficult Argoverse dataset.

(3) \textit{Adversarial attack:} We test our model without adversarial attack by replacing Eq.~\eqref{eqn:simaug_adv} with $V_{1:h}^{adv} = V_{1:h}$.
This is similar to applying the Mixup method~\cite{zhang2017mixup} to two views in the feature space. The performance drops across all three benchmarks. 

(4) \textit{View selection:} We replace Eq.~\eqref{eqn:simaug_select_c} with random search to see the effect of view selection. As we see, the significant performance drops, especially on the Stanford Drone dataset, verifying the effectiveness of this design.

\begin{table}[]
\centering
\caption{Performance on ablated versions of our method on three benchmarks. We report 
the minADE\textsubscript{1}($\downarrow$)/minFDE\textsubscript{1}($\downarrow$) metrics.} 
\begin{tabular}{l c c c}
\toprule
Method & VIRAT/ActEV   & Stanford Drone & Argoverse       \\
\midrule
SimAug full model&  21.7 / 42.2 &  15.7 / 30.2 & 67.9 / 175.6\\
\midrule
- top-down view data &  22.8 / 43.6 & 18.4 / 35.6 &  68.4 / 178.3\\ 
- random perturbation & 23.6 / 43.8 & 18.7 / 35.6  & 69.1 / 180.2 \\ 
- adversarial attack &  23.1 / 43.8 & 17.4 / 32.9 & 68.0 / 177.5 \\ 
- view selection &  23.0 / 42.9 & 19.6 / 38.2 & 68.6 / 177.0\\ 
\bottomrule
\end{tabular}
\label{tab:simaug_ablation}
\end{table}

\section{Related Work}
This work studies the novel direction of robustness in trajectory prediction model~\cite{rudenko2020human}.
We are the first to look into the generalization of trajectory prediction model and study the invariant representation for different environments and camera views.
In the following, we also review some of the trajectory prediction works based on their specific camera views.
We then review more broadly of learning from simulation and robust model learning.

\noindent\textbf{Trajectory prediction.}
Recently there is a large body of work on predicting person future trajectories in a variety of scenarios.
Many works~\cite{alahi2016social,xue2018ss,zhang2019sr,liang2019peeking,liang2020garden,sadeghian2018sophie} focused on modeling person motions in videos recorded with 45-degree-view stationary cameras.
Datasets like VIRAT/ActEV~\cite{oh2011large}, ETH/UCY~\cite{lerner2007crowds,luber2010people} have been used for such direction.
Meanwhile, many works~\cite{lee2017desire,sadeghian2018car,bansal2018chauffeurnet,hong2019rules,zhao2019multi,makansi2019overcoming,li2019way,rhinehart2018r2p2} have been proposed for top-down view videos for trajectory prediction.
Notably, the Stanford Drone Dataset (SDD)~\cite{robicquet2016learning} is used in many works~\cite{sadeghian2018sophie,deo2020trajectory,li2019way} for trajectory prediction with drone videos.
Other works have also looked into pedestrian prediction in dashcam videos~\cite{mangalam2019disentangling,styles2019forecasting,kooij2014context,lee2017desire} and first-person videos~\cite{yagi2018future,styles2019multiple}.
Many vehicle trajectory datasets~\cite{caesar2019nuscenes,chang2019argoverse,yu2018bdd100k} have been proposed as a result of self-driving's surging popularity.

\noindent\textbf{Learning from 3D simulation data.}
As the increasing research focus in 3D computer vision~\cite{zhang2015fast,liang2017event,shah2018airsim,dosovitskiy2017carla,richter2016playing,ros2016synthia,heess2017emergence}, many research works have used 3D simulation for training and evaluating real-world tasks~\cite{gaidon2016virtual,de2017procedural,wu2019revisiting,zhu2017target,sun2019stochastic,liang2020garden,sun2018multi,bak2018domain,kar2019meta,chen2019data}.
Many works~\cite{qiu2017unrealcv,gaidon2016virtual,de2017procedural} were proposed to use data generated from 3D simulation for video object detection, tracking, and action recognition analysis. 
For example, Sun et al.~\cite{sun2019stochastic} proposed a forecasting model by using a gaming simulator.
AirSim ~\cite{shah2018airsim} and CARLA ~\cite{dosovitskiy2017carla} were proposed for robotic autonomous controls for drones and vehicles.
Zeng et al.~\cite{zeng2019adversarial} proposed to use 3D simulation for adversarial attacks. RSA~\cite{zhang2019rsa} used randomized simulation data for human action recognition.
The ForkingPaths dataset~\cite{liang2020garden} was proposed for evaluating multi-future trajectory prediction. Human annotators were asked to control agents in a 3D simulator to create a multi-future trajectory dataset. 

\noindent\textbf{Robust Deep Learning.}
Traditional domain adaptation approaches \cite{bousmalis2017unsupervised,ganin2016domain,tzeng2017adversarial,kang2019contrastive} may not be applicable as our target domain is considered ``unseen'' during training. Methods for learning using privileged information~\cite{lambert2018deep,vapnik2015learning,lopez2015unifying,luo2018graph} is not applicable for a similar reason. Closest to ours is robust deep learning methods. In particular, our approach is inspired by the following methods: (i) \textit{adversarial training}~\cite{goodfellow2014explaining,madry2017towards,xie2019feature,zeng2019adversarial} to defend the adversarial attacks generated on-the-fly during training using gradient-based methods~\cite{madry2017towards,goodfellow2014explaining,tramer2017ensemble,cheng2019robust}; (ii) data augmentation methods to overcome unknown variances between training and test examples such as Mixup~\cite{zhang2017mixup}, MentorMix~\cite{jiang2019synthetic}, AugMix~\cite{cheng2020advaug}, etc; (iii) example re-weighting or selection~\cite{jiang2018mentornet,ren2018learning,jiang2015self,liang2016learning,northcutt2019confident} to mitigate network memorization.
Different from prior work, ours uses 3D simulation data as a new perspective for data augmentation and is carefully designed for future trajectory prediction.

\section{Summary}

In this chapter,
we have introduced \textit{\fancyname}, a novel simulation data augmentation method to learn robust representations for trajectory prediction. Our model is trained only on 3D simulation data and applied out-of-the-box to a wide variety of real video cameras with novel views or scenes. 
We have shown that our method achieves competitive performance on three public benchmarks with and without using the real-world training data.
We believe our approach will facilitate future research and applications on learning robust representation for trajectory prediction with limited or zero training data. Other directions to deal with camera view dependence include using a homography matrix, which may require an additional step of manual or automatic calibration of multiple cameras. We leave them for future work.

\part{Joint Analysis of Human Actions and Trajectory Prediction}
\label{part:joint}
In the final part, we aim to build robust prediction model with enhanced contextual cues from scene semantics and human actions through multi-task learning.
We first study jointly predicting short-term pedestrian trajectories and activities on common benchmarks (\autoref{chap:0301_joint}).
Since short-term future prediction is not enough to ensure safe operations in autonomous driving applications, we introduce a new, human-annotated long-term trajectory and action prediction benchmark with multi-view camera data in urban traffic scenes. (\autoref{chap:0302_longterm_data}).
Finally, we utilize our findings from the first two parts and develop a robust model for the aforementioned challenging long-term trajectory prediction task (\autoref{chap:0303_longterm_model}).
\chapter{Joint Pedestrian Trajectory and Action Prediction on Common Benchmarks
}  \label{chap:0301_joint}

In this chapter, we explore joint trajectory and action prediction in common benchmarks for short-term prediction (short-term means prediction 3-5 seconds into the future). 
We propose the \textit{Next} model~\cite{liang2019peeking} for pedestrian prediction which utilizes rich visual features and multi-task learning.

\section{Motivation}

With the advancement in deep learning, systems now are able to analyze an unprecedented amount of rich visual information from videos.
An important analysis is forecasting the future path of pedestrians, called future person trajectory prediction.
This problem has received increasing attention in the computer vision community~\cite{kitani2012activity,alahi2016social,gupta2018social}. It is regarded as an essential building block in video understanding because looking at the visual information from the past to predict the future is useful in many applications like self-driving cars, socially-aware robots~\cite{luber2010people}, etc.

\begin{figure}[ht]
	\centering
		\includegraphics[width=0.8\textwidth]{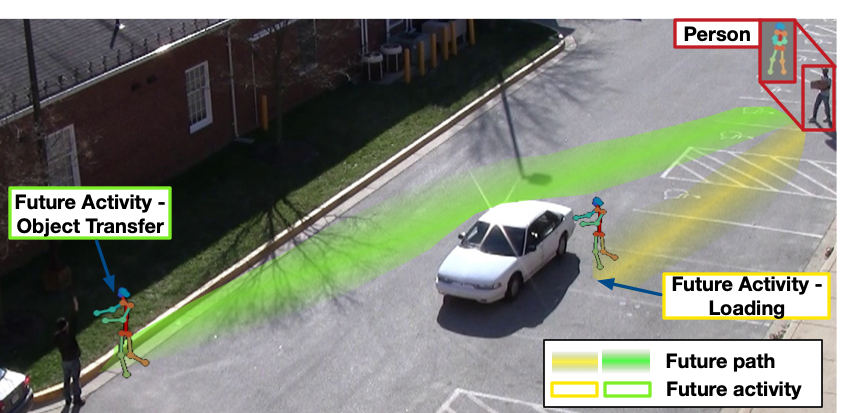} 
	\caption{Our goal is to jointly predict a person's future path and activity. The green and yellow line show two possible future trajectories and two possible activities are shown in the green and yellow boxes. Depending on the future activity, the person (top right) may take different paths, e.g., the yellow path for ``loading'' and the green path for ``object transfer''. }
	\label{fig:next}
\end{figure}

Humans navigate through public spaces often with specific purposes in mind, ranging from simple ones like entering a room to more complicated ones like putting things into a car. Such intention, however, is mostly neglected in existing work. 
Consider the example in Fig.~\ref{fig:next}, the person (at the top-right corner) might take different paths depending on their intention, e.g., they might take the green path to~\emph{transfer object} or the yellow path to~\emph{load object into the car}. Inspired by this, this work is interested in modeling the future path jointly with such intention in videos. 
We model the intention in terms of a predefined set of 29 activities provided by the NIST such as ``loading'', ``object transfer'', etc. See supplementary material for the full list.

The joint prediction model can have two benefits. First, learning the activity together with the path may benefit the future path prediction. 
Intuitively, humans are able to read from others' body language to anticipate whether they are going to cross the street or continue walking along the sidewalk. 
In the example of Fig.~\ref{fig:next}, the person is carrying a box, and the man at the bottom left corner is waving at the person. Based on common sense, we may agree that the person will take the green path instead of the yellow path. 
Second, the joint model advances the capability of understanding not only the future path but also the future activity by taking into account the rich semantic context in videos. 
This increases the capabilities of automated video analytics for social good,
such as safety applications like anticipating pedestrian movement at traffic intersections or a road robot helping humans transport goods to a car. 
Note that our techniques focus on predicting a few seconds into the future, and should not be useful for non-routine activities.

To this end, we propose a multi-task learning model called~\emph{\textit{Next}} which has prediction modules for learning future paths and future activities simultaneously. 
As predicting future activity is challenging, we introduce two new techniques to address the issue. 
First, unlike most of the existing work~\cite{kitani2012activity,alahi2016social,gupta2018social,sadeghian2018sophie,manh2018scene,xie2018learning} which oversimplifies a person as a point in space, we encode a person through rich semantic features about visual appearance, body movement and interaction with the surroundings,  motivated by the fact that humans derive such predictions by relying on similar visual cues. 
Second, to facilitate the training, we introduce an auxiliary task for future activity prediction, i.e., activity location prediction.
In the auxiliary task, we design a discretized grid which we call the Manhattan Grid as location prediction target for the system.

To the best of our knowledge, our work is the first on joint future path and activity prediction in streaming videos, and more importantly the first to demonstrate such joint modeling can considerably improve the future path prediction. 
We empirically validate our model on two benchmarks: ETH \& UCY~\cite{pellegrini2010improving,lerner2007crowds}, and ActEV/VIRAT~\cite{oh2011large,2018trecvidawad}. 
Experimental results show that our method outperforms state-of-the-art baselines, achieving the best-published result on two common benchmarks and producing additional prediction about the future activity.
To summarize, the contributions of this work are threefold:
\textbf{(i)} We conduct a pilot study on joint future path and activity prediction in videos. We are the first to empirically demonstrate the benefit of such joint learning.
\textbf{(ii)} We propose a multi-task learning framework with new techniques to tackle the challenge of joint future path and activity prediction.
\textbf{(iii)} Our model achieves the best-published performance on two public benchmarks. Ablation studies are conducted to verify the contribution of the proposed sub-modules.

\section{The \textit{Next} Model}
Humans navigate through spaces often with specific purposes in mind. Such purposes may considerably orient the future trajectory/path. This motivates us to study the future path prediction jointly with the intention. In this work, we model the intention in terms of a predefined set of future activities such as ``walk'', ``open\_door'', ``talk'', etc.

\begin{figure}[!t]
	\centering
		\includegraphics[width=0.9\textwidth]{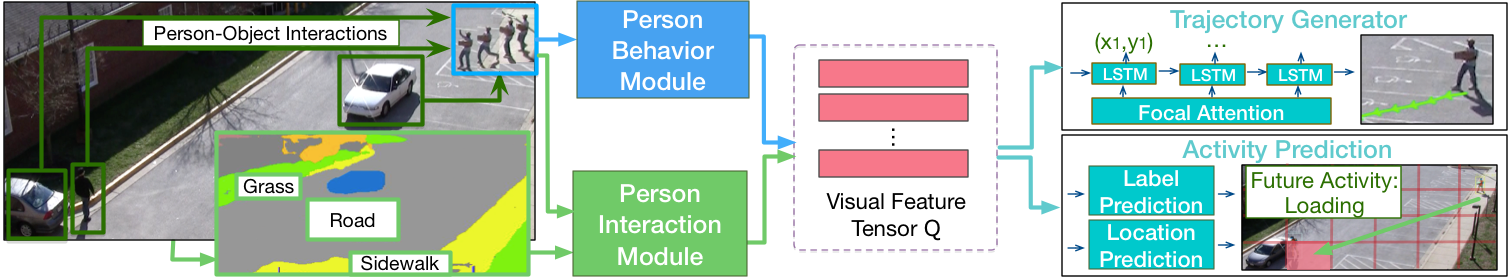}
	\caption{Overview of our model. Given a sequence of frames containing the person for prediction, our model utilizes person behavior module and person interaction module to encode rich visual semantics into a feature tensor. 
	}  
	\label{fig:next_overall}
\end{figure}

\noindent\textbf{Problem Formulation}: Following~\cite{alahi2016social,gupta2018social,sadeghian2018sophie}, we assume each scene is first processed to obtain the spatial coordinates of all people at different time instants. Based on the coordinates, we can automatically extract their bounding boxes. Our system observes the bounding box of all the people from time 1 to $T_{obs}$, and objects if there are any, and predicts their positions (in terms of $xy$-coordinates) for time $T_{obs+1}$ to $T_{pred}$, meanwhile estimating the possibilities of future activity labels at time $T_{pred}$.

\subsection{Network Architecture} \label{network}
Fig.~\ref{fig:next_overall} shows the overall network architecture of our \emph{\textit{Next}} model. 
Unlike most of the existing work~\cite{kitani2012activity,alahi2016social,gupta2018social,sadeghian2018sophie,manh2018scene,xie2018learning} which oversimplifies a person as a point in space, our model employs two modules to encode rich visual information about each person's behavior and interaction with the surroundings. \emph{\textit{Next}}  has the following key components:

\noindent\textbf{Person behavior module} extracts visual information from the behavioral sequence of the person. 

\noindent\textbf{Person interaction module} looks at the interaction between a person and their surroundings. 

\noindent\textbf{Trajectory generator} summarizes the encoded visual features and predicts the future trajectory by the LSTM decoder with focal attention~\cite{liang2018focal,liang2019focal}. 

\noindent\textbf{Activity prediction} utilizes rich visual semantics to predict the future activity label for the person.
In addition, we divide the scene into a discretized grid of multiple scales, which we call the Manhattan Grid, to compute classification and regression for robust activity location prediction. 

In the rest of this section, we will introduce the above modules and the learning objective in details.

\subsection{Person Behavior Module} \label{sec:person-behavior}
\begin{figure}[t]
	\centering
		\includegraphics[width=0.7\textwidth]{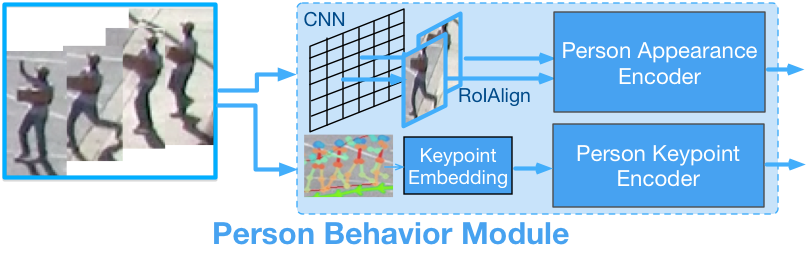}
	\caption{Person behavior module given a sequence of person frames.}
	\label{fig:next_behavior}
\end{figure}

This module encodes the visual information about every individual in a scene. As opposed to oversimplifying a person as a point in space, we model the person's the appearance and body movement. To model appearance changes of a person, we utilize a pre-trained object detection model with ``RoIAlign''~\cite{he2017mask} to extract fixed size CNN features for each person bounding box. See Fig.~\ref{fig:next_behavior}. We average the features along the spatial dimensions for each person and feed them into an LSTM encoder. Finally, we obtain a feature representation of ${T_{obs} \times d}$, where $d$ is the hidden size of the LSTM. 
To capture the body movement, we utilize a person keypoint detection model trained on MSCOCO dataset~\cite{fang2017rmpe} to extract person keypoint information. We apply the linear transformation to embed the keypoint coordinates before feeding into the LSTM encoder. The shape of the encoded feature has the shape of ${T_{obs} \times d}$.
These appearance and movement features are commonly used in a wide variety of studies and thus do not introduce new concern on machine learning fairness.

\subsection{Person Interaction Module} \label{sec:person-scene}
This module looks at the interaction between a person and their surroundings, i.e. person-scene and person-objects interactions.

\noindent\textbf{Person-scene.} 
To encode the nearby scene of a person, we first use a pre-trained scene segmentation model~\cite{deeplabv3plus2018} to extract pixel-level scene semantic classes for each frame. We use totally $N_s=10$ common scene classes, such as roads, sidewalks, etc. 
The scene semantic features are integers (class indexes) of the size $T_{obs} \times h \times w$, where $h,w$ are the spatial resolution. We first transform the integer tensor into $N_s$ binary masks (one mask for each class), and average along the temporal dimension. This results in $N_s$ real-valued masks, each of the size of $h \times w$. We apply two convolutional layers on the mask feature with a stride of 2 to get the \emph{scene CNN features} in two scales.

\begin{figure}[t]
	\centering
		\includegraphics[width=0.7\textwidth]{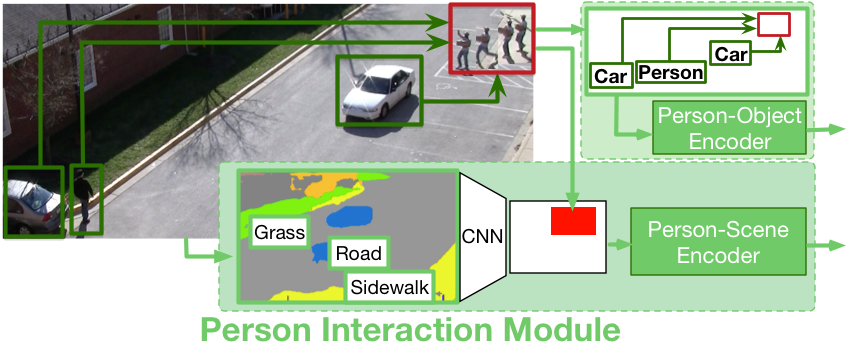}
	\caption{The person interaction module includes person-scene and person-objects modeling. 
	}
	\label{fig:next_person-scene}
\end{figure}

Given a person's $xy$-coordinate, we pool the scene features at the person's current location from the convolution feature map. As the example shown at the bottom of Fig.~\ref{fig:next_person-scene}, the red part of the convolution feature is the discretized location of the person at the current time instant.
The receptive field of the feature at each time instant, i.e. the size of the spatial window around the person which the model looks at, depends on which scale is being pooled from and the convolution kernel size. 
In our experiments, we set the scale to 1 and the kernel size to 3, which means our model looks at the 3-by-3 
surrounding area of the person at each time instant. 
The person-scene representation for a person is in $\mathbb{R}^{T_{obs} \times C}$, where $C$ is the number of channels in the convolution layer. We feed this into a LSTM encoder in order to capture the temporal information and get the final person-scene features in $\mathbb{R}^{T_{obs} \times d}$.

\noindent\textbf{Person-objects.} Unlike previous work~\cite{alahi2016social,gupta2018social} which relies on LSTM hidden states to model nearby people, our module explicitly models the \textit{geometric relation} and the \textit{object type} of all the objects/persons in the scene. At any time instant, given the observed box of a person  $(x_{b}, y_{b}, w_{b}, h_{b})$ and $K$ other objects/persons in the scene $\{(x_{k}, y_{k}, w_{k}, h_{k}) | k \in [1, K] \}$, we encode the geometric relation into $\mathcal{G} \in \mathbb{R}^{K \times 4}$, the $k$-th row of which equals to:
\begin{equation}
\small
\mathcal{G}_{k} =  [\log(\frac{|x_{b} - x_{k}|}{w_{b}}), \log(\frac{|y_{b} - y_{k}|}{h_{b}}), \log(\frac{w_{k}}{w_{b}}), \log(\frac{h_{k}}{h_{b}})]
\end{equation}
This encoding computes the geometric relation in terms of the geometric distance and the fraction box size. We use a logarithmic function to reflect our observation that human trajectories are more likely to be affected by close-by objects or people. This encoding has been proven effective in object detection~\cite{hu2018relation}.
For the object type, we simply use one-hot encoding to get the feature in $\mathbb{R}^{K \times N_o}$, where $N_o$ is the total number of object classes. We then embed the geometric features and the object type features at the current time into $d_e$-dimensional vectors and feed the embedded features into an LSTM encoder to obtain the final feature in $\mathbb{R}^{T_{obs} \times d}$. 


As shown in the example from Fig.~\ref{fig:next_person-scene}, the person-objects feature can capture how far away the person is to the other person and the cars. 
The person-scene feature can capture whether the person is near the sidewalk or grass.
We design this information to the model with the hope of learning things like a person walks more often on the sidewalk than the grass and tends to avoid bumping into cars.

\subsection{Trajectory Generation with Focal Attention}\label{sec:focal_attention}
As discussed, the above four types of visual features, i.e. appearance, body movement, person-scene, and person-objects, are encoded by separate LSTM encoders into the same dimension. Besides, given a person's trajectory output from the last time instant, we extract the trajectory embedding by
\begin{equation}
    e_{t-1} = \tanh\{W_{e} [x_{t-1}, y_{t-1}]\} + b_e \in \mathbb{R}^{d},
\end{equation}
where $[x_{t-1}, y_{t-1}]$ is the trajectory prediction of time $t-1$ and $W_e, b_e$ are learnable parameters. We then feed the embedding $e_{t-1}$ into another LSTM encoder for the trajectory. The hidden states of all encoders are packed into a tensor named $Q \in \mathbb{R}^{M \times T_{obs} \times d}$, where $M=5$ denotes the total number of features and $d$ is the hidden size of the LSTM.

Following~\cite{gupta2018social}, we use an LSTM decoder to directly predict the future trajectory in the $xy$-coordinate. The hidden state of this decoder is initialized using the last state of the person's trajectory LSTM encoder. At each time instant, the $xy$-coordinate will be computed from the decoder state $h_t = \text{LSTM}(h_{t-1}, [e_{t-1}, \tilde{q}_t])$ and by a fully connected layer. $\tilde{q}_t$ is an important attended feature vector which summarizes salient cues in the input features $Q$. We employ an effective focal attention~\cite{liang2018focal} to this end. It was originally proposed to carry out multimodal inference over a sequence of images for visual question answering. The key idea is to project multiple features into a space of correlation, where discriminative features can be easier to capture by the attention mechanism. 
To do so, we compute a correlation matrix $S^{t} \in \mathbb{R}^{M \times T_{obs}}$ at every time instant $t$, where each entry $S^t_{ij} = h_{t-1}^{\top} \cdot Q_{ij:}$ is measured using the dot product similarity and $:$ is a slicing operator that extracts all elements from that dimension. Then we compute two attention matrices: 
\begin{align}
\small
    A^t &= \text{softmax}(\max_{i=1}^{M} S^t_{i:}) \in \mathbb{R}^{M} \\
    B^t &= [\text{softmax}(S^t_{1:}), \cdots, \text{softmax}(S^t_{M:})] \in \mathbb{R}^{M \times T_{obs}} 
\end{align}
Then the attended feature vector is given by:
\begin{equation} \label{eq:next_focal}
\tilde{q}_t = \sum_{j=1}^M A^t_{j} \sum_{k=1}^{T_{obs}} B^t_{jk} Q_{jk:} \in \mathbb{R}^{d}
\end{equation}
As shown, the focal attention models the correlation among different features and summarizes them into a low-dimensional attended vector.

\subsection{Activity Prediction}
Since the trajectory generation module outputs one location at a time, errors may accumulate across time and the final destination would deviate from the actual location. 
Using the wrong location for activity prediction may lead to bad accuracy.
To counter this disadvantage, we introduce an auxiliary task, i.e. activity location prediction, in addition to predicting the future activity label of the person. We describe the two prediction modules in the following.

\begin{figure}[t]
	\centering
		\includegraphics[width=0.7\textwidth]{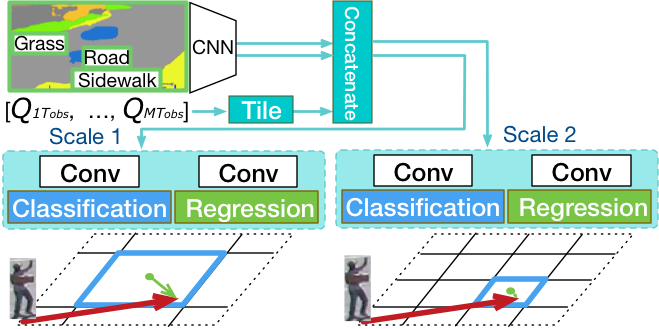}
	\caption{Activity location prediction with classification and regression on the multi-scale Manhattan Grid.}
	\label{fig:next_grid_loss}
\end{figure}

\noindent\textbf{Activity location prediction with the Manhattan Grid.}
To bridge the gap between trajectory generation and activity label prediction, we propose an activity location prediction module to predict the final location of where the person will engage in the future activity. 
The activity location prediction includes two tasks, \emph{location classification} and \emph{location regression}.
As illustrated in Fig.~\ref{fig:next_grid_loss}, we first divide a video frame into a discretized $h \times w$ grid, namely \emph{Manhattan Grid}, and learn to classify the correct grid block and at the same time to regress from the center of that grid block to the actual location. 
Specifically, the aim for the classification task is to predict the correct grid block in which the final location coordinates reside. 
After classifying the grid block, the aim for the regression task is to predict the deviation of the grid block center (green dots in the figure) to the final location coordinate (the end of green arrows).
The reason for adding the regression task are: (i) it will provide more precise locations than just a grid block area; (ii) it is complementary to the trajectory prediction which requires $xy$-coordinates localization.
We repeat this process on the Manhattan Grid of different scales and use separate prediction heads to model them. These prediction heads are trained end-to-end with the rest of the model. 
Our idea is partially inspired by the region proposal network~\cite{ren2015faster} and our intuition is that similar to object detection problem, we need accurate localization using multi-scale features in a cost-efficient way.

As shown in Fig.~\ref{fig:next_grid_loss}, we first concatenate the scene CNN features (see Section~\ref{sec:person-scene}) with the last hidden state of the encoders (see Section~\ref{sec:focal_attention}). For compatibility, we tile the hidden state $Q_{:T_{obs}:}$ along the height and width dimension resulting in a tensor of the size $M \times d \times w \cdot h$, where $w \cdot h$ is the total number of the grid blocks. The hidden state contains rich information from all encoders and allow gradients flow smoothly through from prediction to feature encoders.

The concatenated features are fed into two separate convolution layers for classification and regression. The convolution output for grid classification $\mathsf{cls}_{grid} \in \mathbb{R}^{w \cdot h \times 1}$ indicates the probability of each grid block being the correct destination. In comparison, the convolution output for grid regression $\mathsf{rg}_{grid} \in \mathbb{R}^{w \cdot h \times 2}$ denotes the deviation, in the $xy$-coordinates, between the final destination and every grid block center. A row of $\mathsf{rg}_{grid}$ represents the difference to a grid block, calculated from $[x_t-x_{ci}, y_t-y_{ci}]$ where $(x_t, y_t)$ denotes the predicted location and $(x_{ci},y_{ci} )$ is the center of the $i$-th grid block. The ground truth for the grid regression can be computed in a similar way. During training, only the correct grid block receives gradients for regression.
Recent work~\cite{manh2018scene} also incorporates the grid for location prediction. Our model differs in that we link grid locations to scene semantics, and use a classification layer and a regression layer together to make more robust predictions.

\noindent\textbf{Activity label prediction.}
Given the encoded visual observation sequence, the activity label prediction module predicts the future activity at time instant $T_{pred}$. We compute the future $N_{a}$ activity probabilities using the concatenated last hidden states of the encoders:
\begin{equation}
    \mathsf{cls}_{act} = \text{softmax}(W_{a} \cdot [Q_{1T_{obs}:}, \cdots, Q_{MT_{obs}:}])
\end{equation}
where $W_a$ is a learnable weight. The future activity of a person could be multi-class, e.g. a person could be ``walking'' and ``carrying'' at the same time.

\subsection{Training}
The entire network is trained end-to-end by minimizing a multi-task objective. The primary loss is the common $L_2$ loss between the predicted future trajectories and the ground-truth trajectories~\cite{manh2018scene,gupta2018social,sadeghian2018sophie}. The loss is summed into $L_{xy}$ over all persons from $T_{obs+1}$ to $T_{pred}$.

The second category of loss is the activity location classification and regression loss. We have $L_{grid\_cls} \!=\! \!\sum_{i=1}^N \! \text{ce}(\mathsf{cls}^i_{grid},\mathsf{cls}^{\ast i}_{grid})$, where $\mathsf{cls}^{\ast i}_{grid}$ is the ground-truth final location grid block ID for the $i^{th}$ training trajectory. 
Likewise $L_{grid\_reg} \!=\! \!\sum_{i=1}^N \text{smooth}_{L_1}(\mathsf{rg}^i_{grid},\mathsf{rg}^{\ast i}_{grid})$ and $\mathsf{rg}^{\ast i}_{grid}$ is the ground-truth difference to the correct grid block center. This loss is designed to bridge the gap between the trajectory generation task and activity label prediction task.

The third loss is for activity label prediction.
We employ the cross-entropy loss: $L_{act} = \sum_{i=1}^N \text{ce}(\mathsf{cls}^i_{act},\mathsf{cls}^{\ast i}_{act})$.
The final loss is then calculated from: 
\begin{equation}
    L = L_{xy} + \lambda(L_{grid\_cls} + L_{grid\_reg}) + L_{act}
\end{equation}
We use a balance controller $\lambda = 0.1$ for location destination prediction to offset their higher loss values during training.

\section{Experimental Results}
We evaluate the proposed~\emph{\textit{Next}}~model on two common benchmarks for future path prediction: ETH~\cite{pellegrini2010improving} and UCY~\cite{lerner2007crowds}, and ActEV/VIRAT~\cite{2018trecvidawad,oh2011large}. 

\subsection{ActEV/VIRAT}\label{sec:exp-virat}

\noindent\textbf{Dataset \& Setups.} ActEV/VIRAT~\cite{2018trecvidawad} is a public dataset released by NIST in 2018 for activity detection research in streaming video ({\footnotesize \url{https://actev.nist.gov/}}). This dataset is an improved version of VIRAT~\cite{oh2011large}, with more videos and annotations. It includes 455 videos at 30 fps from 12 scenes, more than 12 hours of recordings. Most of the videos have a high resolution of 1920x1080. 
We use the official training set for training and the official validation set for testing.
Following~\cite{alahi2016social,gupta2018social,sadeghian2018sophie}, the models observe 3.2 seconds (8 frames) of every person and predict the future 4.8 seconds (12 frames) of person trajectory. We downsample the videos to 2.5 fps and extract person trajectories using the code released in~\cite{gupta2018social}. Since we do not have the homographic matrix, we use the pixel values for the trajectory coordinates as it is done in~\cite{yagi2018future}. 

\noindent\textbf{Evaluation Metrics.} Following prior work~\cite{alahi2016social,gupta2018social,sadeghian2018sophie}, we use two error metrics for person trajectory prediction: 

\noindent i) \textit{Average Displacement Error} (ADE): The average Euclidean distance between the ground truth coordinates and the prediction coordinates over all time instants,
\begin{equation}
    \text{ADE} = \frac{ \sum^{N}_{i=1} \sum^{T_{pred}}_{t=1} \lVert \tilde{Y}^i_t - Y^i_t \rVert_{2}}{N*T_{pred}}
\end{equation}

\noindent ii) \textit{Final Displacement Error} (FDE): The euclidean distance between the predicted points and the ground truth point at the final prediction time instant $T_{pred}$,
\begin{equation}
    \text{FDE} = \frac{ \sum^{N}_{i=1} \lVert \tilde{Y}^i_{T_{pred}} - Y^i_{T_{pred}} \rVert_{2}}{N}
\end{equation}
The errors are measured in the pixel space on ActEV/VIRAT whereas in meters on ETH and UCY. 
For future activity prediction, we use mean average precision (mAP).

\noindent\textbf{Baseline methods.} We compare our method with the two simple baselines and two recent methods: \textbf{\textit{Linear}} is a single layer model that predicts the next coordinates using a linear regressor based on the previous input point. \textbf{\textit{LSTM}} is a simple LSTM encoder-decoder model with coordinates input only. \textbf{\textit{Social LSTM}}~\cite{alahi2016social}: We train the social LSTM model to directly predict trajectory coordinates instead of Gaussian parameters. \textbf{\textit{SGAN}}~\cite{gupta2018social}: We train two model variants (PV \& V) detailed in this work using the released code from Social-GAN~\cite{gupta2018social} {\footnotesize (\url{https://github.com/agrimgupta92/sgan/})}.

Aside from using a single model at test time, Gupta et al.~\cite{gupta2018social} also used 20 model outputs per frame and selected the best prediction to count towards the final performance. Following the practice, we train 20 identical models using random initializations and report the same evaluation results, which are marked ``20 outputs'' in Table~\ref{tab:next_exp1}.

\begin{table}[t]
\centering
\small
\begin{tabular}{l|l||c|c|c|c}
\hline
&Method & ADE          & FDE          & {\scriptsize move\_ADE}     & {\scriptsize move\_FDE}      \\ \hline \hline
\multirow{6}{*}{\rotatebox[origin=c]{90}{\footnotesize{Single Model}}}&Linear                     & 32.19        & 60.92        & 42.82        & 80.18         \\ 
&LSTM & 23.98        & 44.97        & 30.55        & 56.25        \\ 
&Social LSTM  & 23.10        & 44.27        & 28.59        & 53.75       \\ 
&SGAN-PV  & 30.51        & 60.90        & 37.65        & 73.01         \\ 
&SGAN-V   & 30.48        & 62.17        & 35.41        & 68.77         \\ 
&Ours     & \textbf{17.99} & \textbf{37.24} & \textbf{20.34} & \textbf{42.54} \\ 
&Ours-Noisy     & 34.32 &  57.04  & 40.33 &	66.73 \\ 
\hline
\multirow{3}{*}{\rotatebox[origin=c]{90}{\scriptsize{20 Outputs}}}&SGAN-PV-20 & 23.11        & 41.81        & 29.80        & 53.04         \\ 
&SGAN-V-20  & 21.16        & 38.05        & 26.97        & 47.57       \\ 
&Ours-20                            & \textbf{16.00}        & \textbf{32.99}        & \textbf{17.97}        & \textbf{37.28}             \\ \hline
\end{tabular}
\caption{Comparison to baseline methods on the ActEV/VIRAT validation set. Top uses the single model output. Bottom uses 20 outputs. Numbers denote errors thus lower are better.}
\label{tab:next_exp1}
\end{table}

\noindent\textbf{Implementation Details.}
We use LSTM cell for both the encoder and decoder. 
The embedding size $d_e$ is set to 128, and the hidden sizes $d$ of encoder and decoder are both 256. 
Ground truth bounding boxes of persons and objects are used during the observation period (from time 1 to $T_{obs}$).
For person keypoint features, we utilize the pre-trained pose estimator from~\cite{fang2017rmpe} to extract 17 joints for each ground truth person box. 
For person appearance feature, we utilize the pre-trained object detection model FPN~\cite{lin2017feature} to extract appearance features from person bounding boxes. 
The scene semantic segmentation features are resized to (64, 36) and the scene convolution layers are set to have a kernel size of 3, a stride of 2 and the channel dimension is 64. 
We resize all videos to 1920x1080 and utilize two grid scales, 32x18 and 16x9. 
The activation function is $tanh$ if not stated otherwise and we do not use any normalization. 
For training, we use Adadelta optimizer~\cite{zeiler2012adadelta} with an initial learning rate of 0.1 and the dropout value is 0.3. 
We use gradient clipping of 10 and weight decay of 0.0001. 
For Social LSTM, the neighbor is set to 256 pixels as in~\cite{yagi2018future}. 
All baselines use the same embedding size and hidden size as our model, therefore all encoder-decoder models have about the same numbers of parameters. Other hyper-parameters we use for the baselines follow the ones in ~\cite{gupta2018social}.

\begin{figure}[!t]
	\subfigure{\label{fig:quala}}
    \subfigure{\label{fig:qualb}}
    \subfigure{\label{fig:qualc}}
    \subfigure{\label{fig:quald}}
    \subfigure{\label{fig:quale}}
    \subfigure{\label{fig:qualf}}
    \subfigure{\label{fig:qualg}}
    \subfigure{\label{fig:qualh}}
	\centering
		\includegraphics[width=1.0\textwidth]{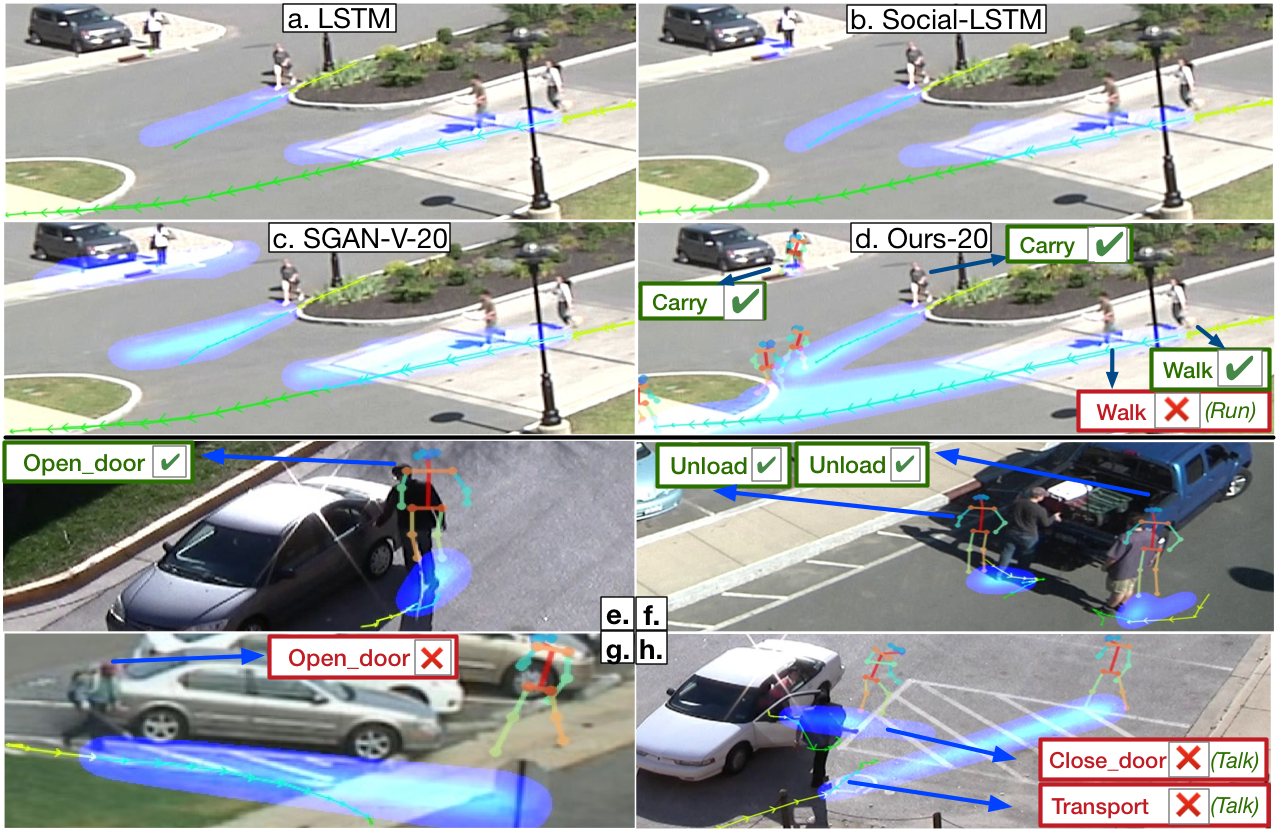} 
	\caption{(Better viewed in color.) Qualitative comparison between our method and the baselines. Yellow path is the observable trajectory and green path is the ground truth trajectory during the prediction period. Predictions are shown as blue heatmaps. Our model also predicts the future activity, which is shown in the text and with the person pose template.}  
	\label{fig:qualitative-compare}
\end{figure}

\noindent\textbf{Main Results.} Table~\ref{tab:next_exp1} lists the testing error, where the top part is the error of a single model output and the bottom shows the best result of 20 model outputs. The ``ADE'' and ``FDE'' columns summarize the error over all trajectories, and the last two columns further detail the subset trajectories of moving activities (``walk'', ``run'', and ``ride\_bike''). 
We report the mean performance of 20 runs of our single model at Row 7. The standard deviation on ``ADE'' metric is 0.043. Full numbers can be found in supplemental material.
As we see, our method performs favorably against other methods, especially in predicting the trajectories of moving activities. For example, our model outperforms Social-LSTM and Social-GAN by a large margin of 10 points in terms of the ``move\_FDE'' metric. 
The results demonstrate the efficacy of the proposed model and its state-of-the-art performance on future trajectory prediction.
Additionally, as a step towards real-world application, we train our model with noisy outputs from object detection and tracking during the observation period. For evaluation, following common practise in tracking~\cite{wu2015object}, for each trajectory, we assume the person bounding box location at time $1$ is close to the ground truth location, and we evaluate the model prediction using tracking inputs and other visual features from time $1$ to $T_{obs}$ as shown in Table~\ref{tab:next_exp1} ``Ours-Noisy''.

\noindent\textbf{Qualitative analysis.} 
We visualize and compare our model outputs and the baselines in Fig.~\ref{fig:qualitative-compare}.
As we see, our method outputs more accurate trajectories for each person, especially for the two persons on the right that were about to accelerate their movement. Our method is also able to predict most of the activities correct except one (walk versus run).
Our model successfully predicts the activity ``carry'' and the static trajectory of the person near the car, while in Fig~\ref{fig:qualc}, SGAN predicts several moving trajectories in different directions. 
We further provide a qualitative analysis of our model predictions. (i) Successful cases: In Fig~\ref{fig:quale} and~\ref{fig:qualf}, both the trajectory prediction and future activity prediction are correct. 
(ii) Imperfect case: In Fig~\ref{fig:qualg}, although the trajectory prediction is mostly correct, our model predicts that the person is going to open the door of the car, given the observation that he is walking towards the side of the car. 
(iii) Failed case: In Fig~\ref{fig:qualh}, our model fails to capture the subtle interactions between the two persons and predicts that they will go separate ways, while in fact they are going to stop and talk to each other. 

\begin{table}
\centering
\small
\begin{tabular}{l||c|c|c}
\hline
Method                             & ADE $\downarrow$  & FDE $\downarrow$  & Act mAP $\uparrow$\\ \hline
Our full model      & 17.91 & 37.11 & 0.192           \\ \hline
No p-behavior           & 18.99 & 39.82 & 0.139            \\ 
No p-interaction          & 18.83 & 39.35    & 0.163            \\ 
No focal attention                 & 19.93 & 42.08 & 0.144           \\ 
No act label loss                        & 19.48 & 41.45 & -                 \\
No act location loss              &19.07 & 39.91 &   0.152          \\
No multi-task               & 20.37 & 42.79     & -                        \\ \hline
\end{tabular}
\caption{Multi-task performance \& ablation experiments.}
\label{tab:next_multi-task}
\end{table}

\begin{table}
\centering
\small
\begin{tabular}{l||c|c|c}
\hline
Method                             & ADE $\downarrow$  & FDE $\downarrow$  & Act mAP $\uparrow$\\ \hline
Our full model      & 17.91 & 37.11 & 0.192           \\ \hline
MobileNet-ADE20K                 & 18.37 & 38.41 & 0.185           \\ 
Xception-CityScapes                     & 18.54 & 38.37 & 0.173          \\
MobileNet-CityScapes                     & 18.77 & 38.99 & 0.178          \\ \hline
No p-interaction          & 18.83 & 39.35    & 0.163            \\ \hline
\end{tabular}
\caption{Scene semantic segmentation ablation experiments.}
\label{tab:next_scene_ablation}
\end{table}

\subsection{Ablation Model}

In Table~\ref{tab:next_multi-task}, we systematically evaluate our method through a series of ablation experiments, where ``ADE'' and ``FDE'' denotes the errors thus lower are better. ``Act'' is the mean Average Precision (mAP) of the activity label prediction over 29 activities and higher are better.

\noindent\textbf{Efficacy of rich visual features.} We investigate the feature contribution of person behavior and person interactions. 
As shown in the first three rows in Table~\ref{tab:next_multi-task}, both features are important to trajectory prediction while person behavior features are more essential for activity prediction. Individual feature ablations are in the supplementary material.

\noindent\textbf{Effect of focal attention.} In the fourth row of Table~\ref{tab:next_multi-task}, we replace focal attention in Eq.~\eqref{eq:next_focal} with a simple average of the last hidden states from all encoders. Both trajectory and activity prediction hurt as a result.

\noindent\textbf{Impact of multi-task learning.} In the last three rows of Table~\ref{tab:next_multi-task}, we remove the additional tasks of predicting the activity label or the activity location or both to see the impact of multi-task learning. Results show the benefit of our multi-task learning method.

\noindent\textbf{Effect of scene semantic segmentation models.}
Since we use a pre-trained scene semantic segmentation model to extract features for our model, we want to see how different level of accuracies for the segmentation model would affect our prediction performance. In Table~\ref{tab:next_scene_ablation}, we compare the segmentation model that we use in the full model (Xception trained on ADE20K) with MobileNet, which is a light-weight segmentation model that is about 10\% worse on the scene segmentation benchmark~\cite{deeplabv3plus2018}. We also compare models trained on a different dataset (ADE20K vs. CityScapes).
As we see, a stronger scene segmentation model leads to a slightly better performance and ADE20K models are better in general.

\subsection{ETH \& UCY}\label{exp2-dataset}
\noindent\textbf{Dataset.} ETH~\cite{pellegrini2010improving} and UCY~\cite{lerner2007crowds} are common datasets for person trajectory prediction benchmark~\cite{alahi2016social,gupta2018social,manh2018scene,sadeghian2018sophie}.
Same as previous works, we report performance by averaging over both datasets.
We use the same data processing method and settings detailed in~\cite{gupta2018social}. 
This benchmark includes videos from five scenes: ETH, HOTEL, UNIV, ZARA1 and ZARA2. Leave-one-scene-out data split is used and we evaluate our model on 5 sets of data. 
We follow the same testing scenario and baselines as in the previous section. We have also cited the latest state-of-the-art results from~\cite{sadeghian2018sophie}. Due to 1 video cannot be downloaded, we use a smaller test set for UNIV and a smaller training set across all splits. The other 4 test sub-datasets are the same as in~\cite{gupta2018social} so the numbers are comparable.

\begin{table}[]
\centering
\small
\begin{tabular}{l|l||c|c||c|c|c||c}
\hline
                   & Method& ETH         & HOTEL       & UNIV  *      & ZARA1       & ZARA2       & AVG         \\ \hline \hline
\multirow{4}{*}{\rotatebox[origin=c]{90}{\footnotesize{Single Model}}}&Linear & 1.33 / 2.94 & 0.39 / 0.72 & 0.82 / 1.59 & 0.62 / 1.21 & 0.77 / 1.48 & 0.79 / 1.59 \\ 
&LSTM                & 1.09 / 2.41 & 0.86 / 1.91 & \textbf{0.61} / \textbf{1.31} & \textbf{0.41} / \textbf{0.88} & 0.52 / 1.11 & 0.70 / 1.52 \\ 
&Alahi et al.~\cite{alahi2016social}         & 1.09 / 2.35 & 0.79 / 1.76 & 0.67 / 1.40 & 0.47 / 1.00 & 0.56 / 1.17 & 0.72 / 1.54 \\ 
&Ours-single   & \textbf{0.88} / \textbf{1.98} & \textbf{0.36} / \textbf{0.74} & 0.62 / 1.32 & 0.42 / 0.90 & \textbf{0.34} / \textbf{0.75} & \textbf{0.52} / \textbf{1.14} \\ \hline

\multirow{4}{*}{\rotatebox[origin=c]{90}{20 Outputs}}&~\cite{gupta2018social}(V)         & 0.81 / 1.52 & 0.72 / 1.61 & 0.60 / 1.26 & 0.34 / 0.69 & 0.42 / 0.84 & 0.58 / 1.18 \\ 
&~\cite{gupta2018social}(PV)        & 0.87 / 1.62 & 0.67 / 1.37 & 0.76 / 1.52 & 0.35 / 0.68 & 0.42 / 0.84 & 0.61 / 1.21 \\ 
&~\cite{sadeghian2018sophie}             & \textbf{0.70} / \textbf{1.43} & 0.76 / 1.67 & \textbf{0.54} / \textbf{1.24} & \textbf{0.30} / \textbf{0.63} & 0.38 / 0.78 & 0.54 / 1.15 \\ 
&Ours-20               & 0.73 / 1.65 & \textbf{0.30} / \textbf{0.59} & 0.60 / 1.27 & 0.38 / 0.81 & \textbf{0.31} / \textbf{0.68} & \textbf{0.46} / \textbf{1.00} \\ \hline
\end{tabular}
\caption{Comparison of different methods on ETH (Column 3 and 4) and UCY datasets (Column 5-7). 
* We use a smaller test set on UNIV since 1 video is unable to download.}
\label{tab:next_exp2}
\end{table}

Since there is no activity annotation, we do not use activity label prediction module in our model. Since the annotation is only a point for each person and the human scale in each video doesn't change much, we apply a fixed size expansion from points for each video to get the person bounding box annotation for feature pooling. We do not use any other bounding box. We don't use any additional annotation compared to baselines to ensure a fair comparison.

\noindent\textbf{Implementation Details.} We do not use person keypoint feature. Final location loss and trajectory L2 loss are used. 
Unlike ~\cite{sadeghian2018sophie}, we don't utilize any data augmentation. We train our model for 40 epochs with the adadelta optimizer. Other hyper-parameters are the same as in Section~\ref{sec:exp-virat}.

\noindent\textbf{Results \& Analysis.} Experiments are shown in Table~\ref{tab:next_exp2}. Our model outperforms other methods in both evaluations, where we obtain the best-published single model on ETH and best average performance on the ETH \& UCY benchmark. 
As shown in the table, our model performs much better on HOTEL and ZARA2. The average movement at each time-instant in these two scenes are 0.18 and 0.22, respectively, much lower than others: 0.389 (ZARA1), 0.460 (ETH), 0.258 (UNIV).
Recall that the leave-one-scene-out data split is used in training.
The results suggest other methods are more likely to overfit to the trajectories of large movements, e.g. Social-GAN~\cite{gupta2018social} often "over-shoot" when predicting the future trajectories. In comparison, our method uses attention to find the "right" visual signal and show better performance for trajectories of small movements on HOTEL and ZARA2 while still being competitive for trajectories of large movements.

\section{Related Work}
This work falls under the category of sequential models that utilize both static and dynamic environmental cues in the human motion prediction literature~\cite{rudenko2020human}. 
We utilize multiple contextual visual cues from both the target agent and the environment.
We are the first work to jointly optimize activity and trajectory prediction with a unified framework.
In the following, we review a few relevant recent approaches based on their use of these contextual cues. 
Then we also review some activity prediction works.

\noindent\textbf{Person-person models for trajectory prediction.} 
Person trajectory prediction models try to predict the future path of people, mostly pedestrians. 
A large body of work learns to predict person path by considering human social interactions and behaviors in crowded scene~\cite{xu2018encoding,yi2016pedestrian}.
Zou et al. in~\cite{zou2018understanding} learned human behaviors in crowds by imitating a decision-making process. 
Social-LSTM~\cite{alahi2016social} added social pooling to model nearby pedestrian trajectory patterns. 
Social-GAN~\cite{gupta2018social} added adversarial training on Social-LSTM to improve performance. 
Different from these previous work, we represent a person by rich visual features instead of simply considering a person as points in the scene. 
Meanwhile we use \emph{geometric relation} to explicitly model the person-person relations in the scene,
which has not been used in previous work.

\noindent\textbf{Person-scene models for trajectory prediction.} 
A number of works focused on learning the effects of the physical scene, e.g., people tend to walk on the sidewalk instead of grass. 
Kitani et al. in ~\cite{kitani2012activity} used Inverse Reinforcement Learning to forecast human trajectory. 
Xie et al. in ~\cite{xie2018learning} considered pedestrian as ``particles'' whose motion dynamics are modeled within the framework of Lagrangian Mechanics. Scene-LSTM~\cite{manh2018scene} divided the static scene into Manhattan Grid and predict pedestrian's location using LSTM. 
CAR-Net~\cite{jaipuria2018transferable} proposed an attention network on top of scene semantic CNN to predict person trajectory.
SoPhie~\cite{sadeghian2018sophie} combined deep neural network features from scene semantic segmentation model and generative adversarial network (GAN) using attention to model person trajectory.
A disparity to ~\cite{sadeghian2018sophie} is that we explicitly pool scene semantic features around each person at each time instant so that the model can directly learn from such interactions.

\noindent\textbf{Person visual features for trajectory prediction.} 
Some recent works have attempted to predict person path by utilizing individual's visual features instead of considering them as points in the scene. Kooij et al. in~\cite{kooij2014context} looked at pedestrian's faces to model their awareness to predict whether they will cross the road using a Dynamic Bayesian Network in dash-cam videos. Yagi et al. in ~\cite{yagi2018future} used person keypoint features with a convolutional neural network to predict future path in first-person videos. 
Different from these works, we consider rich visual semantics for future prediction that includes both the person behavior and their interactions with soundings .

\noindent\textbf{Activity prediction/early recognition \& Tracking.}
Many works have been proposed to anticipate future human actions using Recurrent Neural Network (RNN). \cite{ma2016learning} and \cite{aliakbarian2017encouraging} proposed different losses to encourage LSTM to recognize actions early in internet videos. Srivastava et al. in~\cite{srivastava2015unsupervised} utilized unsupervised learning with LSTM to reconstruct and predict video representations.
Another line of works is anticipating human activities in robotic vision ~\cite{koppula2016anticipating,jain2015car}.
There are previous works that take into account multiple cues in videos for tracking~\cite{jain2016structural, sadeghian2017tracking} and group activity recognition~\cite{choi2014understanding, shu2015joint, shu2017cern}. 
Our work differs in that rich visual features and focal attention are used for joint person path and activity prediction. 
Meanwhile, our work utilizes novel activity location prediction to bridge the two tasks.

\section{Summary}
In this chapter, we have presented a new neural network model for predicting human trajectory and future activity simultaneously. We first encode a person through rich visual features capturing human behaviors and interactions with their surroundings. 
Then we add an auxiliary task of predicting the activity locations to facilitate the joint training process.
We refer to the resulting model as \textit{Next}. 
We showed the efficacy of our model on both popular and recent large-scale video benchmarks on person trajectory prediction.
In addition, we quantitatively and qualitatively demonstrated that our \textit{Next} model successfully predicts meaningful future activities.

However, since short-term future prediction is not enough to ensure safe operations in autonomous driving applications, in the following chapters, we introduce a new, human-annotated long-term trajectory and action prediction benchmark and new models to tackle this challenge.
\chapter{
A Multi-view Long-term Trajectory Prediction Benchmark
}  \label{chap:0302_longterm_data}

In the following chapters, we explore the problem of long-term trajectory prediction, which aims to expand the prediction horizon two-fold to three-fold compared to most previous works so that the models can provide better traffic safety.
In this chapter, we first describe our new human-annotated long-term trajectory prediction dataset with multi-view camera data in urban traffic scenes.

\section{Motivation}

\begin{figure}[ht]
	\centering
		\includegraphics[width=0.95\textwidth]{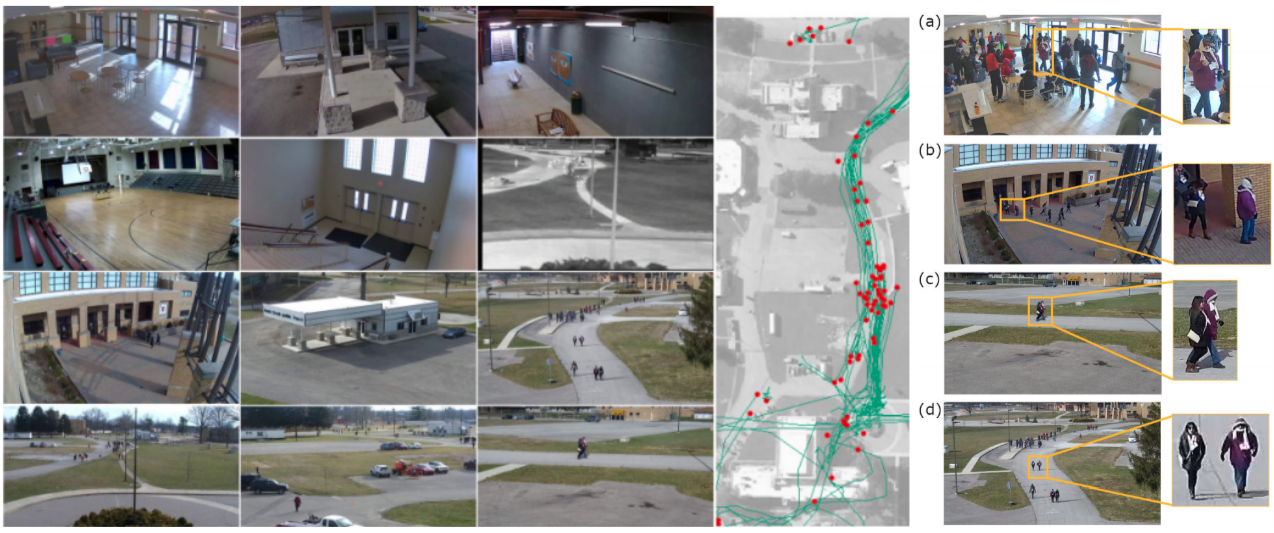} 
	\caption{Overview of the MEVA dataset. Image taken from ~\cite{corona2021meva}.}
	\label{fig:meva_dataset}
\end{figure}

Most previous works~\cite{gupta2018social,liang2019peeking} on pedestrian trajectory prediction only focus on predicting the near future (within 5 seconds). In applications like autonomous systems, safe operation and collision avoidance are crucial for the systems to be co-mingling with humans.
This requires predicting human motion beyond a short time-horizon. 
While near-future prediction can be informed by recent observed trajectory history, long-term prediction depends on inferring human intents or goals, which requires the model to make use of scene constraints, social interactions, person-vehicle interactions and even common sense reasoning.

Long-term prediction has the following challenges. 
First, it is highly uncertain compared to short-term forecasting. Using only the trajectory history is not enough to achieve accurate long-term prediction, and semantic reasoning with scene constraints and agent behaviors is crucial, especially in traffic scenes.
Second, the target person might be traveling out-of-frame and hence require the models to infer the final intended destination without actually seeing it.
Third, long-term prediction may require more observation inputs that result in greater computation demand while it is critical to have an efficient system that could output prediction in a timely fashion.

To tackle the aforementioned challenges, we need to construct a new dataset as existing datasets like ETH/UCY~\cite{pellegrini2010improving,lerner2007crowds}, Stanford Drone~\cite{robicquet2016learning}, only have short-term trajectories and no rich behavioral annotations about the agents like activities. 
Another important aspect of these previous datasets is that they lack person-vehicle interactions, which is important for trajectory prediction in urban traffic scenes.
We turn to the recently release the Multiview Extended Video with Activities (MEVA) dataset~\cite{corona2021meva}, for its characteristics that could help us solve the problems mentioned before. 
There are three main reason to use the MEVA dataset. 
First, the MEVA dataset contains long-term tracks that last up to a few minutes that are captured by at lease one camera. See Figure~\ref{fig:meva_dataset} on the right, where two women with jackets traveled from the cafe in a building to the bus station two hundred meters away.
Second, the MEVA dataset contains human-annotated activities in a urban traffic setting, which includes person-only activities and person-vehicle activities. It could be utilized for enhanced modeling of human behaviors for trajectory prediction in traffic scenes.
Last but not least, as shown in Figure~\ref{fig:meva_dataset}, there are multiple overlapping camera views, which could help us study how to solve the second challenge where the pedestrians traveled out-of-frame to other cameras.
Since the MEVA dataset does not contain object tracking annotations (only individual activity instance annotations are provided), in the next section, we describe how we collect and annotate the MEVA trajectory prediction dataset (MEVA-Trajectory) in the next section.

\noindent\textbf{Existing datasets.} There are several popular pedestrian trajectory prediction benchmarks, such as SDD~\cite{robicquet2016learning}, ETH/UCY~\cite{pellegrini2010improving,lerner2007crowds}, KITTI~\cite{geiger2013vision} and nuScenes~\cite{caesar2019nuscenes}, all of which are for short-term (1-5 seconds in the future) trajectory prediction. These datasets do not include human action/activity annotations.
The VIRAT/ActEV~\cite{oh2011large} dataset has longer tracks with activity annotations but it does not have the multi-view property as discussed before.
There are other datasets from the object tracking community like Duke MTMC~\cite{ristani2016performance}, which is also multi-view, but it is no longer available due to privacy concerns and there is no activity annotation.  

\begin{table}[]
\centering
\begin{tabular}{c||c}
\hline
                     & Classes \\ \hline \hline
Object &   \makecell{ Bike, Person, Bag, Vehicle }     \\ \hline
Activity  &  \makecell{
person\_opens\_vehicle\_door, person\_enters\_vehicle,\\
person\_talks\_to\_person,
person\_opens\_facility\_door, \\
person\_closes\_vehicle\_door, person\_texts\_on\_phone,\\
person\_enters\_scene\_through\_structure,person\_opens\_trunk,\\
person\_unloads\_vehicle,person\_closes\_trunk,\\
person\_picks\_up\_object,person\_loads\_vehicle,\\
person\_embraces\_person
}      \\ \hline
\end{tabular}
\caption{MEVA Object \& Activity Classes.}
\label{tab:meva_act_class}
\end{table}

\noindent\textbf{Statistics of the MEVA dataset.}
The MEVA dataset is a large multi-view activity dataset with 28 cameras of 1920x1080 resolution, recorded on a large military training facility with scripted and unscripted actions. 
All video data are formatted into 5-minutes video clips.
The type of activities include in our data collection (pedestrian-related) are listed in Table~\ref{tab:meva_act_class}.
We download the evaluation split with 266 videos and the training split (drop-09) with 1277 videos from the MEVA data repository~\footnote{\url{https://gitlab.kitware.com/meva/meva-data-repo/-/tree/master/annotation/DIVA-phase-2/MEVA}}.

\section{The MEVA-Trajectory Dataset}
\subsection{Data Collection}
As mentioned before, we need to get object tracks from the MEVA dataset to form trajectories for our task.
The MEVA-Trajectory dataset is released at
\url{https://github.com/JunweiLiang/MEVA-Trajectory}.
Code to reproduce the results is released at \url{https://github.com/JunweiLiang/Object_Detection_Tracking}.
The data collection process includes the following steps.

\begin{figure}[ht]
	\centering
		\includegraphics[width=0.95\textwidth]{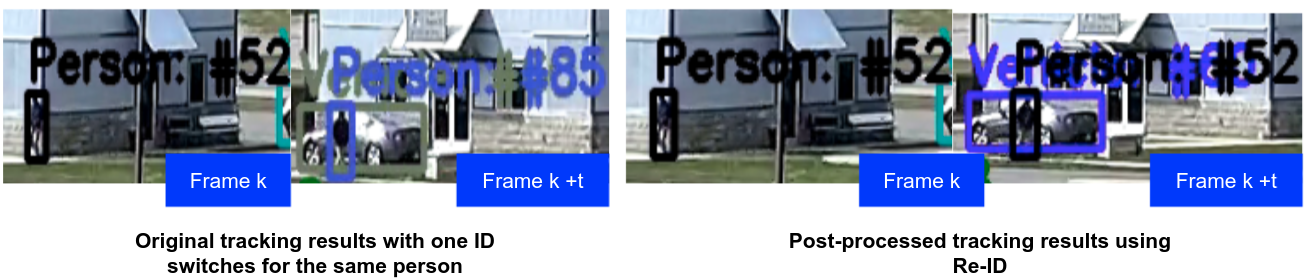} 
	\caption{Single video object detection and tracking with re-identification. On the left is the original tracking results and on the right is after re-identification. See text for details.}
	\label{fig:meva_single_reid}
\end{figure}

\noindent\textbf{Single video object detection and tracking.} First, we use our efficient and accurate object detection and tracking system proposed in \autoref{chap:0101_object} to get individual object tracks, including persons and vehicles.
In addition, in order to have more accurate long-term tracks, we utilize re-identification techniques to refine the person and vehicle tracklets.
Specifically, we compare each detected tracklets that could potentially be considered the same (based on spatial and temporal constraints) using re-identification models, and the tracklets are refined if the distance score is below a threshold.
For person re-identification, we utilize the OSNet~\cite{zhou2019omni} model trained on Market1501 dataset~\cite{zheng2015scalable}.
For vehicle re-identification, we utilize the winning entry~\cite{qian2020electricity} of vehicle tracking competition from the AI City Challenge 2020~\cite{naphade20192019}.
In Figure~\ref{fig:meva_single_reid}, we show an example of before and after using the person re-identification model. As we see, this technique is able to correct the ID switch in the original results. Empirically, we find that the re-identification model can reduce about 10\% ID switch errors from the original tracking results.

\begin{figure}[ht]
	\centering
		\includegraphics[width=0.95\textwidth]{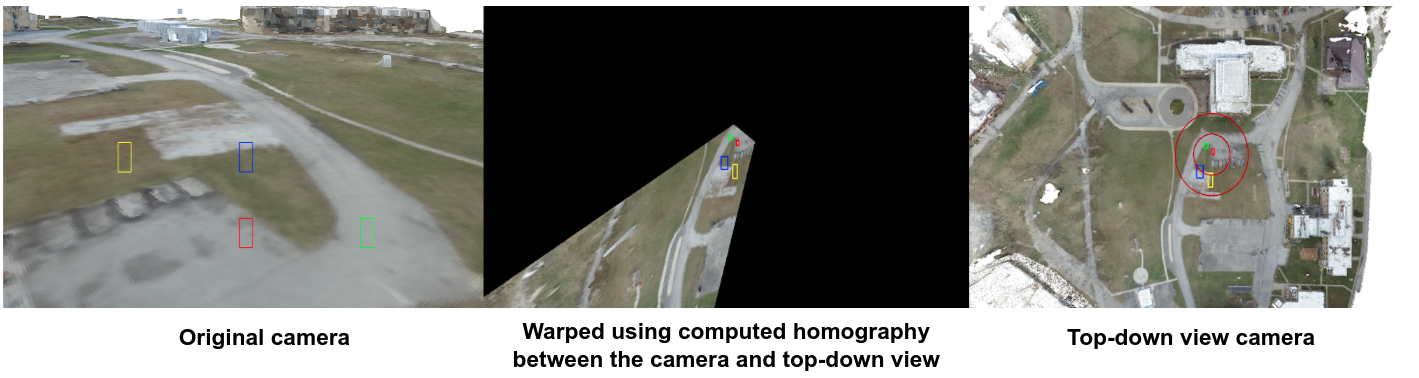}
	\caption{Visualization of homography transformation between original camera and top-down-view camera.}
	\label{fig:meva_homography}
\end{figure}

\begin{figure}[ht]
	\centering
		\includegraphics[width=0.95\textwidth]{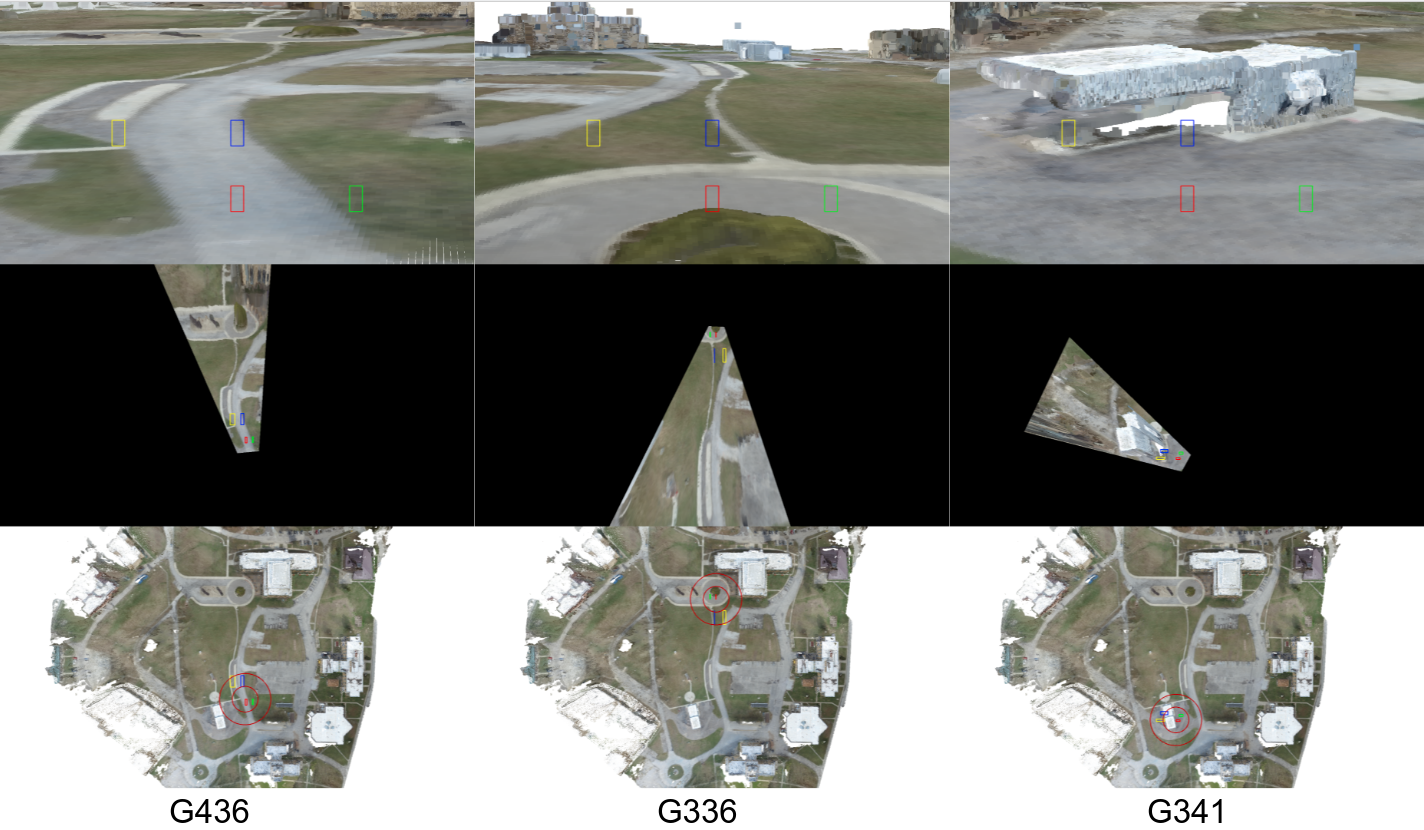}
	\caption{Overview of the MEVA dataset.}
	\label{fig:meva_more_cameras}
\end{figure}

\begin{figure}[ht]
	\centering
		\includegraphics[width=0.95\textwidth]{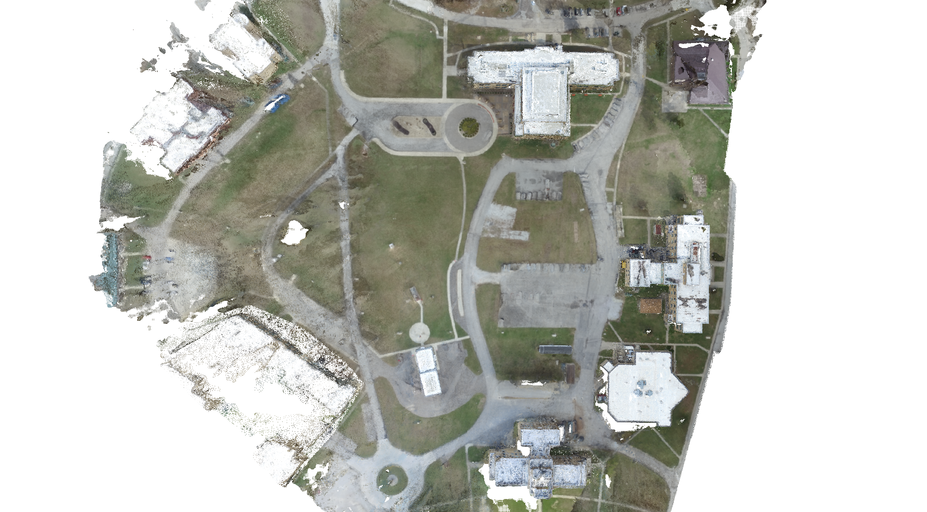}
	\caption{Visualization of top-down-view camera.}
	\label{fig:meva_topdown}
\end{figure}

\noindent\textbf{Cross-view tracklet association.} 
After we have object tracklets from individual videos, we need to associate the same object across multiple videos. 
Therefore we compute a homography matrix between each camera to the ground plane as a spatial constraint for the tracklet association process.
The MEVA dataset provides camera models (intrinsic and extrinsic parameters) and 3D point clouds of the whole facility.
We utilize the 3D point cloud and get a top-down-view camera model as shown in Figure~\ref{fig:meva_topdown}.
We then compute the homography matrix between each original camera and the top-down-view camera by:
\begin{equation}
    H_{ab} = R_a R_b^T - \frac{(-R_a * R_b^T * t_b + t_a) n^T}{d}
\end{equation}
where $R_a,R_b$ and $t_a,t_b$ are the rotation and translations of the two cameras. $n$ is the normal vector of the plane and $d$ is the distance of the camera b to the plane.
To check the accuracy of the homography transformation, we use the computed homography matrix and warp the original camera view to the top-down view, as shown in Figure~\ref{fig:meva_homography} and Figure~\ref{fig:meva_more_cameras}. The colored bounding boxes are for reference. The red circles on the top-down-view image, which are used for range reference, have a radius of 50 and 100, respectively, and they use the center of the red bounding box as their centers.
As we can see, the homography matrices have reasonable accuracy.
To get the final tracklets across different cameras, we first transform the center of each bounding box of the tracklets to the ground-plane and then compare different tracklet's (within a certain spatial and temporal tolerance) similarity using the aforementioned re-identification models.
The tracklet association is done with the Hungarian algorithm.

\begin{figure}[ht]
	\centering
		\includegraphics[width=0.95\textwidth]{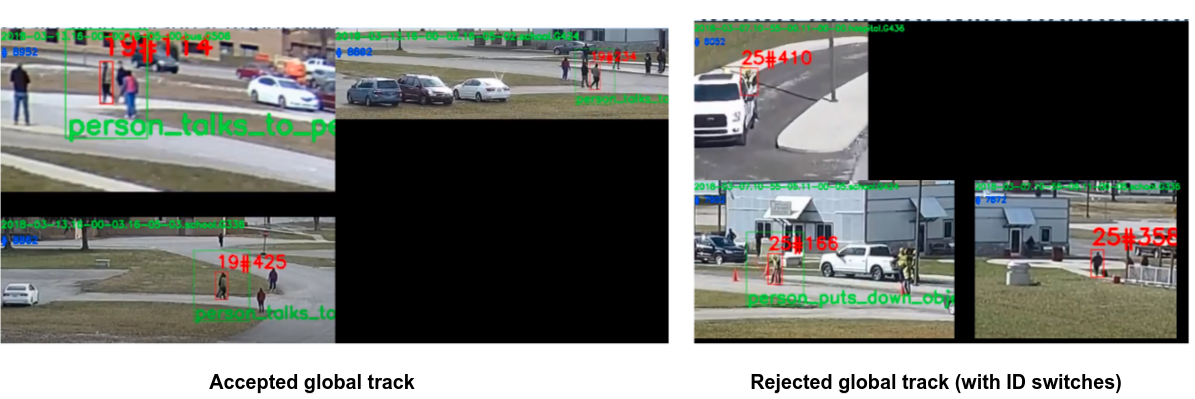}
	\caption{Visualization of accepted and rejected global tracks during the human annotation process. See text for details.}
	\label{fig:meva_annotation}
\end{figure}

\noindent\textbf{Manual annotation of global tracklets.} 
After we get the global tracklets, we conduct a manual annotation process to reject global tracks that are not correct. In Figure~\ref{fig:meva_annotation}, we show an example of the accepted global tracks and an rejected one. The numbers on top of the red bounding boxes are global track ID as well as local track ID. 
The original MEVA activity annotations are also visualized in the figure. They are assigned to the tracks based on the bounding box overlap between the tracklets and the original activity annotations.

\begin{figure}[ht]
	\centering
		\includegraphics[width=0.95\textwidth]{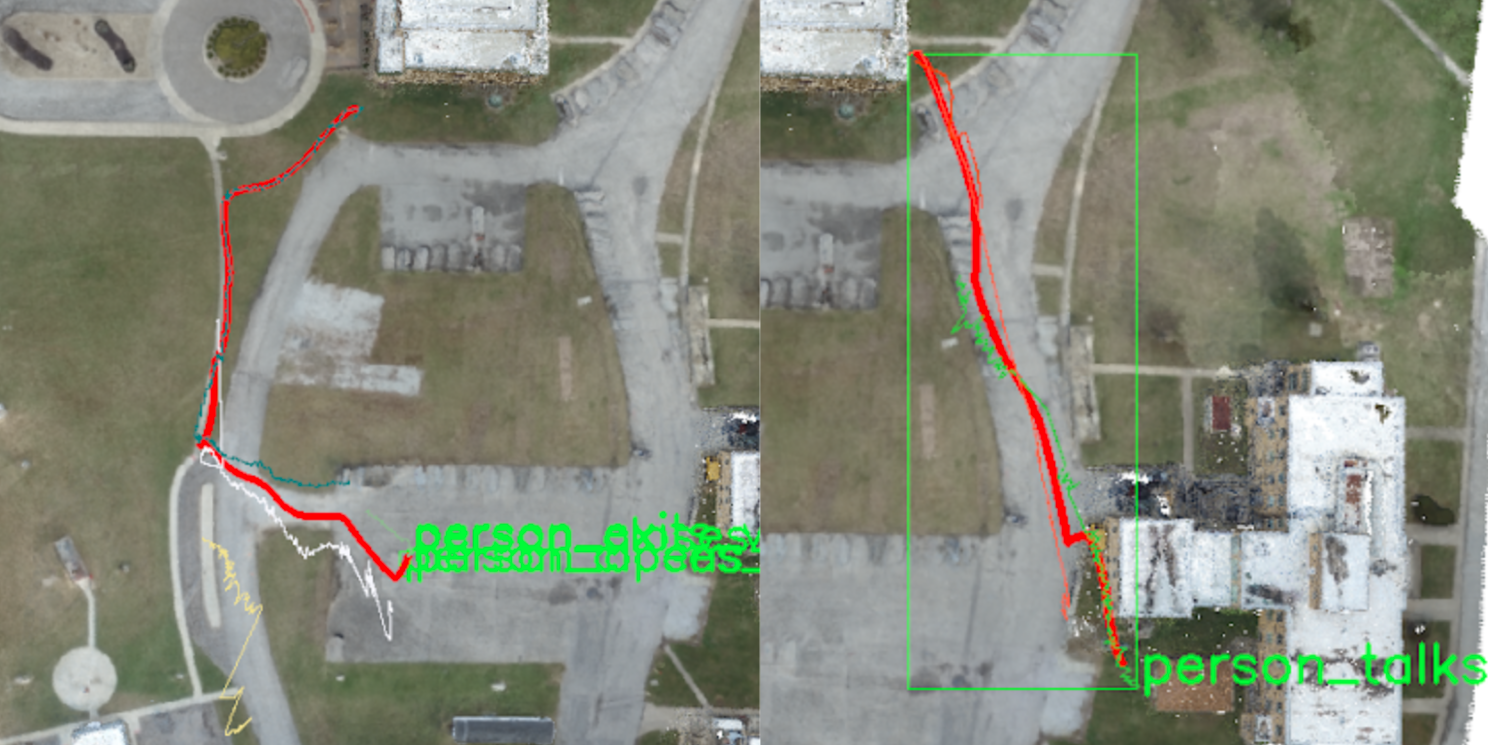}
	\caption{Example of global trajectory smoothing. The red one is the smoothed trajectory and the others are local trajectories. See text for details.}
	\label{fig:meva_smoothing}
\end{figure}

\noindent\textbf{Trajectory smoothing.}
Finally, since the bounding boxes of the tracklets might be unstable due to occlusions and miss-detections, we conduct trajectory smoothing to get a global trajectory for each global track that is suitable for the trajectory prediction task.
For each global tracklet, multiple local trajectories (computed using the bottom center of the bounding boxes) from each camera are smoothed using moving average with a window size of 200 frames (the fps is 30).
We show two examples of the smoothing results in Figure~\ref{fig:meva_smoothing}.
The red trajectories are the final trajectory after smoothing and the rest are local trajectories. The activity annotations are also shown in the figure.
As we see, the left example shows a reasonable trajectory of a person walking from the parking lot to the building along the path.

\subsection{Data Characteristics}

Our MEVA-Trajectory dataset is the first publicly available dataset composed of multi-view long-term pedestrian trajectories with rich activity annotations.
Compared to previous common pedstrian trajectory benchmarks like ETH, UCY, Stanford Drone and KITTI, our new dataset contains a few characteristics.
With the focus of long-term trajectories and enhanced visual understanding, our dataset consists of extended length high resolution videos.
The pedestrians in our dataset are often associated with a clear intent, whether it is going to the bus station to get on a bus or entering a building for a class, thanks to the scripted events from the original MEVA dataset.
Compared to the ActEV~\cite{oh2011large} dataset used in previous \autoref{chap:0301_joint}, MEVA-Trajectory has the advantage of longer tracks with synchronized multi-view data.
Full data characteristics comparison is summarized in Table~\ref{tab:meva_general_comp}.

\begin{table}[]
\centering
\begin{tabular}{l||c|c|c|c|c}
\hline
Datasets      & \small{ETH,UCY~\cite{pellegrini2010improving,lerner2007crowds}} & SDD~\cite{robicquet2016learning} & KITTI~\cite{geiger2013vision} & ActEV~\cite{oh2011large} & Ours \\ \hline \hline
HD Resolution &     -    &   -  &    \checkmark   &   partial    &    \checkmark  \\ \hline
Multi-View    &    -     &  -   &    -   &    -   &    \checkmark  \\ \hline
Extended Length   &    -     &  -   &    \checkmark   &    \checkmark   &    \checkmark  \\ \hline
Event/Goal-Driven   &    -     &  -   &    -   &    partial   &    \checkmark  \\ \hline
Traffic Scene &    -     &  partial   &   \checkmark   &   \checkmark   &    \checkmark  \\ \hline
Activity Annotation   &    -     &  -   &    -   &    \checkmark   &    \checkmark  \\ \hline
\end{tabular}
\caption{Comparison of representative trajectory prediction datasets and our new dataset called MEVA-Trajectory. The pedestrians in our dataset are often associated with a clear intent, whether it is going to the bus station to get on a bus or entering a building for a class, thanks to the scripted events from the original MEVA dataset.See text for details.}
\label{tab:meva_general_comp}
\end{table}

In addition, we compare trajectory statistics between our dataset with the ETH/UCY and the ActEV dataset in Table~\ref{tab:meva_stat_comp}.
As we see, our dataset has more number of trajectories.
Notably, in terms of median trajectory length, our MEVA-Trajectory dataset is 48.3 seconds, which is more than five times longer compared to the 8.8 seconds of the ETH/UCY dataset. This shows that our dataset is suitable for the long-term trajectory prediction task.

\begin{table}[]
\centering
\begin{tabular}{l||c|c|c}
\hline
               & ETH, UCY       & ActEV  & Ours       \\ \hline 
\#Cameras        & 4                  & 5    &     10       \\ \hline
Resolutions    & 640x480,720x576 & 1920x1080, 1280x720 & 1920x1080 \\ \hline
FPS            & 25                  & 30 & 30               \\\hline
Annotation FPS          & 2.5                  & 30 & 30               \\\hline
Total Traj. Length & 4:59:05 & 12:14:44 &   15:36:17    \\ \hline
\#Traj.  & 1535 & 1073 &   2060 / 864*    \\ \hline
Median Traj. Length & 8.8 & 28.8 & 48.3      \\ \hline
Median \#Camera & 1 & 1 &  2     \\ \hline
Annotations         &      Person coordinates        &  \footnotesize{\makecell{Person+object \\bounding boxes,activities}}    &  \footnotesize{\makecell{Person+object \\bounding boxes,activities}}    \\ \hline
\end{tabular}
\caption{Statistical comparison to commonly used person trajectory benchmark datasets. ``*'' means the number of global tracks as opposed to local tracks in a video. The unit of the trajectory length is seconds. See text for details.}
\label{tab:meva_stat_comp}
\end{table}

\begin{figure}[ht]
	\centering
		\includegraphics[width=0.95\textwidth]{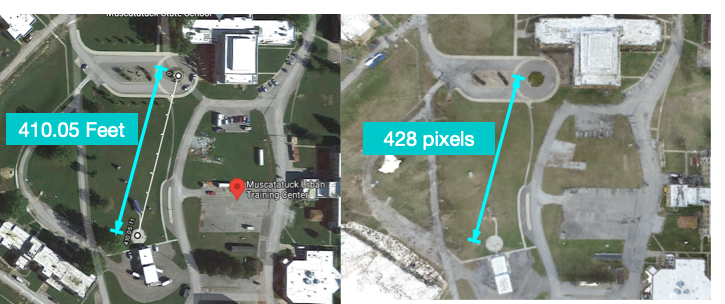}
	\caption{MEVA-Trajectory dataset's unit is in feet. On the left is an screenshot of the satellite view of the dataset location, Muscatatuck Urban Training Center, and on the right is the top-down view of the 3D point cloud.}
	\label{fig:meva_ruler}
\end{figure}

\noindent\textbf{Quality of the Dataset.}
Here we briefly discuss the quality of our collected dataset. 
For single video object detection and tracking, we utilize a well-tuned system for surveillance-type videos that achieves 0.83 average precision for persons and 0.98 for vehicles on VIRAT dataset~\cite{oh2011large}. See Table~\ref{tab:0101_virat_obj} in \autoref{chap:0101_object}.
We then utilize state-of-the-art vehicle and person re-identification model for cross-view tracklet association.
After the manual annotation effort, we have reduced the automatic global person tracklets from 2549 to 864, with a rejection rate of 66\%.
In this way, we have ensured the precision of the global person tracklets so that they could be reliably utilized for long-term trajectory prediction. 
Meanwhile, with high-quality single-video object tracks, we could extract object features for each global track from the videos accurately.

\section{Summary}
In this chapter, we describe our new dataset, called MEVA-Trajectory, for long-term trajectory prediction in urban traffic scenes.
We show in details our process of collecting and annotating the dataset. In the next chapter, we discuss the experiments we have conducted on this dataset.
\chapter{ 
Long-term Trajectory Prediction with Scene and Action Understanding
}  \label{chap:0303_longterm_model}

In previous chapters, we have laid the foundation of standalone action recognition models and future prediction models using scene semantics. 
In this chapter, we propose a new model, called Next-GAT, which utilizes graph attention with scene and action understanding module, for long-term trajectory prediction.
We conduct comprehensive experiments on two datasets, ActEV and our proposed MEVA-Trajectory, and show the efficacy of our Next-GAT model in urban traffic scenes.

\section{Motivation}
As discussed in \autoref{chap:0302_longterm_data}, most previous works~\cite{gupta2018social,liang2019peeking,huang2019stgat,mohamed2020social} on pedestrian trajectory prediction only focus on predicting trajectories of the short-term future (within 5 seconds/12 timesteps). 
To better improve safety in applications like self-driving cars and socially-aware robots, long-term predictions are needed.
In this work, we aim to increase the prediction horizon two folds compared to previous work to predicting 12 seconds (30 timesteps) into the future.
The term long-term is consistent with recent published works~\cite{karasev2016intent,tran2021goal}.

Long-term prediction is challenging compared to short-term forecasting.
In applications like self-driving cars, long-term prediction can better ensure safety.
Using only the trajectory history is not enough to achieve accurate long-term prediction, and semantic reasoning with scene constraints and agent behaviors is crucial. 
To tackle the this challenge, we build on top of the models proposed in previous chapters and propose the Next-Graph-Attention-Network (Next-GAT), which takes both scene semantic understanding and behavioral analysis into account for trajectory prediction.
We add graph attention network (GAT), inspired by a recent successful work that used GAT for trajectory prediction~\cite{huang2019stgat}, to model agent interactions.
In terms of evaluation benchmarks, we choose both the ActEV dataset~\cite{oh2011large} and our MEVA-Trajectory dataset (\autoref{chap:0302_longterm_data}) for a comprehensive analysis in urban traffic scenarios.

\begin{figure}[ht]
	\centering
		\includegraphics[width=0.96\textwidth]{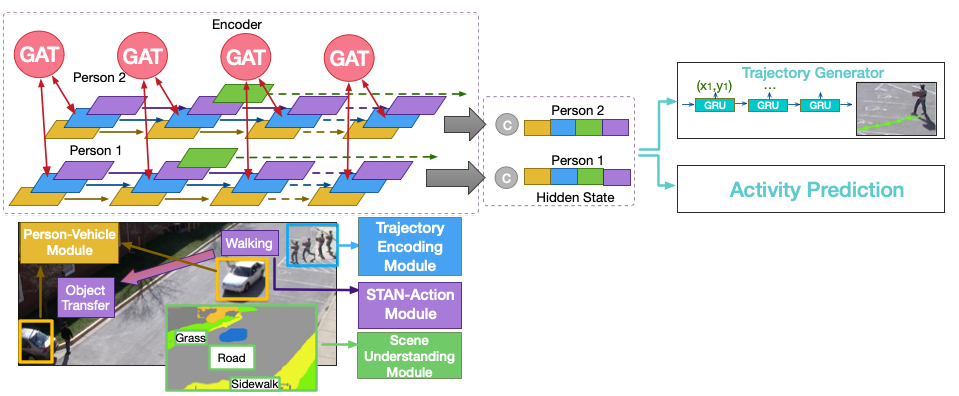} 
	\caption{Overview of the Next-GAT model.}
	\label{fig:next_gat}
\end{figure}

\section{The Next-GAT Model}
Similar to the Next model proposed in \autoref{chap:0301_joint}, we utilize multi-task learning with trajectory prediction and activity prediction.
Following the same problem formulation proposed in \autoref{chap:0301_joint}, we simplify the Next model's person behavior module by removing the pose analysis part and adding the STAN module proposed in \autoref{chap:0103_viewpoint}, to allow the model encoder to learn viewpoint invariant representations that could help both activity and trajectory prediction.
We also simplify the person interaction module, by only using one scene segmentation frames as the static scene semantics do not change a lot during the observation period.
Finally, we add graph attention layer to the trajectory encoder to model the interactions between different agents in the scene.

\subsection{Network Architecture} \label{network}
Fig.~\ref{fig:next_gat} shows the overall network architecture of our Next-GAT model. 
Our model focuses on modeling of both the scene semantics and the agent's behavior, by using viewpoint-invariant action representations and graph attention.
Next-GAT has the following key components:

\noindent\textbf{Trajectory encoding module} takes the relative coordinates of the pedestrian's trajectories during the observation period and applies graph attention modeling~\cite{velivckovic2017graph} at each timestep between all pedestrians in the scene.

\noindent\textbf{STAN-Action module} takes the frame sequence of the target agent to model the behavioral cues for better action prediction and trajectory prediction.

\noindent\textbf{Scene understanding module} takes the scene semantic segmentation feature of the first video frame of the whole scene to encode the static scene information for both action and trajectory prediction.

\noindent\textbf{Person-vehicle interaction module} takes the person-vehicle geometric features from the frame sequence to encode the important interactions between pedestrians and vehicles in traffic scenes.

\noindent\textbf{Trajectory generator} uses the encoded visual features and predicts the future trajectory by a RNN decoder. 

\noindent\textbf{Activity prediction} utilizes the concatenated encoder features to predict the future activity label.
In the rest of this section, we will introduce the above modules and the learning objective in details.

\subsection{Trajectory Encoding Module}
Given the observation trajectory of each pedestrian, we first encode the location coordinates by
\begin{equation}
    e_{t} = RNN_1(MLP [x_{t}, y_{t}]) \in \mathbb{R}^{d},
\end{equation}
where $[x_{t}, y_{t}]$ is the trajectory prediction of time $t-1$, and MLP is a single fully-connected layer to embed the coordinates into higher dimensions.
We therefore get a individual motion-encoded hidden state tensor of $E \in \mathbb{R}^{ T_{obs} \times d}$, where $d$ is the hidden size of the RNN. We use GRU in our implementation.
Then at each timestep, we utilize graph attention network~\cite{velivckovic2017graph} and another recurrent network, to refine each pedestrian's hidden states with all other pedestrians in the scene. It is computed by
\begin{equation}
    H_{t} = GAT(RNN_2(E_{t}, H_{t-1})) \in \mathbb{R}^{K \times d},
\end{equation}
where K is the number of pedestrians in the scene.
Let $h_i$ be the corresponding output of GAT for the $i$-th person at time $t$. It is computed using GAT by
\begin{align}
    h_i = 
    \sum_{j \in \mathcal{K}} f_e([v_i, v_j])v_j
\end{align}
where $\mathcal{K}$ represents all pedestrian in the scene and node $v_i$ is represented as 
$v_i = [\tilde{h_i}]$, the outputs of $RNN_2$.
$f_e$ is some edge function (implemented as an MLP in our experiments) that computes the attention/edge weights and then softmax.

\subsection{STAN-Action Module}
Given the frame sequence of each target agent $F \in \mathbb{R}^{H \times W \times T_{obs} \times c}$, we utilize a light-weighted version of the STAN model proposed in \autoref{chap:0103_viewpoint} to extract a vector representation of the agent's actions.
Specifically, the frame sequence is processed through a Resnet3D group and a STAN layer, before feeding into an MLP layer to get a fix-length vector representation.
Please refer to \autoref{chap:0103_viewpoint} for more details.

\subsection{Scene Understanding Module} 
To encode the static scene semantic information, we first use a pre-trained scene segmentation model~\cite{deeplabv3plus2018} to extract pixel-level scene semantic classes for the first video frame. We use totally $N_s=10$ common scene classes, such as roads, sidewalks, etc., same as in \autoref{chap:0301_joint}.
The scene semantic features are integers (class indexes) of the size $h \times w$, where $h,w$ are the spatial resolution. We first transform the integer tensor into $N_s$ binary masks (one mask for each class). This results in $N_s$ real-valued masks, each of the size of $h \times w$. We apply two convolutional layers on the mask feature with a stride of 2 and then an MLP to get a fix-length vector representation.

\subsection{Person-Vehicle Interaction Module} 
We use the same person-object interaction module from ~\autoref{chap:0301_joint} and explicitly model the \textit{geometric relation} of all the vehicles and pedestrians in the scene. 
At any time instant, given the observed box of a person  $(x_{b}, y_{b}, w_{b}, h_{b})$ and $K$ other vehicles in the scene $\{(x_{k}, y_{k}, w_{k}, h_{k}) | k \in [1, K] \}$, we encode the geometric relation into $\mathcal{G} \in \mathbb{R}^{K \times 4}$, the $k$-th row of which equals to:
\begin{equation}
\small
\mathcal{G}_{k} =  [\log(\frac{|x_{b} - x_{k}|}{w_{b}}), \log(\frac{|y_{b} - y_{k}|}{h_{b}}), \log(\frac{w_{k}}{w_{b}}), \log(\frac{h_{k}}{h_{b}})]
\end{equation}
This encoding computes the geometric relation in terms of the geometric distance and the fraction box size. We use a logarithmic function to reflect our observation that human trajectories are more likely to be affected by close-by objects. 
This encoding has been proven effective in object detection~\cite{hu2018relation}.
We then embed the geometric features at the current time into $d_e$-dimensional vectors and feed the embedded features into an GRU encoder to obtain the final feature in $\mathbb{R}^{T_{obs} \times d}$.

\subsection{Prediction Module} 
We concatenate the fix-length representation of scene semantics, person-vehicle interaction and agent action with the last timestep's output of the trajectory encoding module, and feed into our multi-task learning framework, which includes the trajectory generator and activity prediction module.
Similar to \autoref{chap:0301_joint}, the multi-task training objective include a cross-entropy loss the activity prediction and L2 loss for trajectory prediction.

\begin{table}[]
\centering
\begin{tabular}{l||c|c|c|c|c|c}
\hline
                  & \multicolumn{3}{c|}{Short-term Trajectory Prediction} & \multicolumn{3}{c}{Long-term Trajectory Prediction} \\ \hline
                  & Act      & Single-Future     & Multi-Future     & Act     & Single-Future     & Multi-Future     \\ \hline \hline
NN  & -             & 1.79/3.12         & -                & -            & 3.47/6.5          & -                \\
Const. Vel. & -             & 1.17/2.25         & -                & -            & 2.78/5.74         & -                \\ \hline
Human Perf.* & -             & -         & -                & -            & 2.60/5.49         & -                \\ \hline
SGAN              & -             & 1.21/2.25         & 0.88/1.63        & -            & 3.37/6.66         & 2.69/5.29        \\
STGAT             & -             & 1.43/2.75         & 0.88/1.68        & -            & 4.05/7.78         & 2.27/4.63        \\
STGCNN            & -             & 1.48/2.57         & 1.08/1.93        & -            & 3.46/6.51         & 2.78/5.46        \\\hline
Next              & 0.192         & 1.06/2.03         & 0.87/1.79        & 0.211        & 2.22/4.56         & 1.97/4.05        \\
Next-GAT          & \textbf{0.236}         & \textbf{0.84/1.57}         & \textbf{0.76/1.42}        & \textbf{0.267}        & \textbf{1.94/4.05}         & \textbf{1.63/3.36}       \\\hline
\end{tabular}
\caption{Main results on the ActEV dataset. The numbers are ADE/FDE respectively for trajectories, and mAP for activity (``Act''). ``*'': estimated performance. See evaluation metrics for details.}
\label{tab:0303_actev_exp}
\end{table}

\section{Experiments}
\label{sec:0303_exp}
\noindent\textbf{Dataset \& Setups.} We compare our model and baselines on both short-term and long-term trajectory prediction for a comprehensive analysis of model performance. We evaluate the proposed Next-GAT model on two benchmarks, ActEV and MEVA-Trajectory, since both datasets satisfy the required track length for long-term trajectory prediction (see Table~\ref{tab:meva_stat_comp} in \autoref{chap:0302_longterm_data}).
For short-term trajectory prediction task, we follow the common protocol~\cite{sadeghian2018sophie,alahi2016social,gupta2018social,huang2019stgat,mohamed2020social} of observing 8 timesteps (3.2 seconds) and predicting 12 timesteps (4.8 seconds). 
For long-term trajectory prediction, we set the criteria to observe 12 timesteps (4.8 seconds) and predict 30 timesteps (12 seconds), which is slightly further into the future compared to a recent long-term trajectory prediction work~\cite{tran2021goal}.

\noindent\textbf{Evaluation Metrics.} 
Different from previous works~\cite{sadeghian2018sophie,alahi2016social,gupta2018social,huang2019stgat,mohamed2020social} that only evaluate on the multi-modality of the results, we are also interested in how the model performs given a single best prediction, so that in real-world application, users can better optimize their system according to the computation-accuracy trade-off.
We evaluate all models with both single-future metric ($minADE_1$/$minFDE_1$) and multi-future metric ($minADE_{20}$/$minFDE_{20}$).
For detailed formulations of the metrics, please refer to Section~\ref{sec:0201_exp} in \autoref{chap:0201_multi}.
In terms of activity prediction, we use Mean Average Precision (mAP), same as in the previous chapter.

\begin{figure}[ht]
	\centering
		\includegraphics[width=0.95\textwidth]{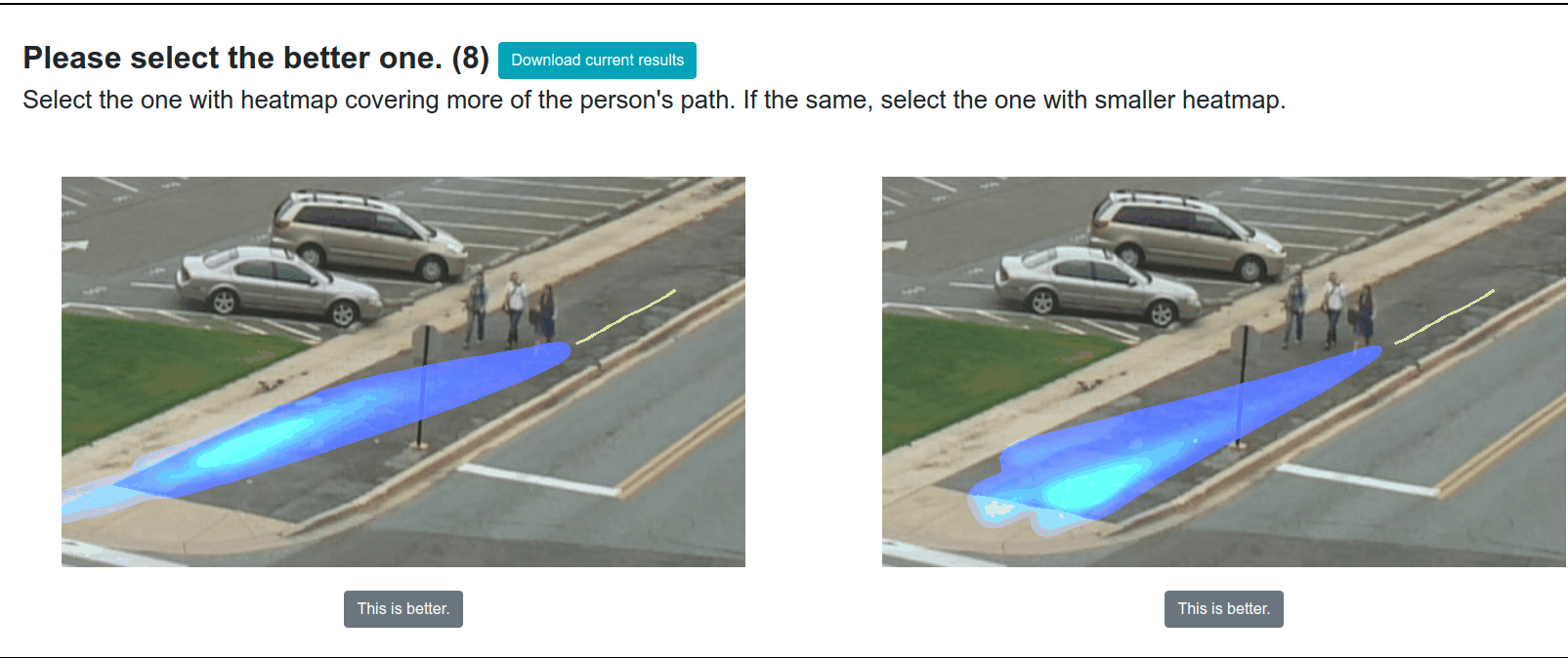}
		\includegraphics[width=0.95\textwidth]{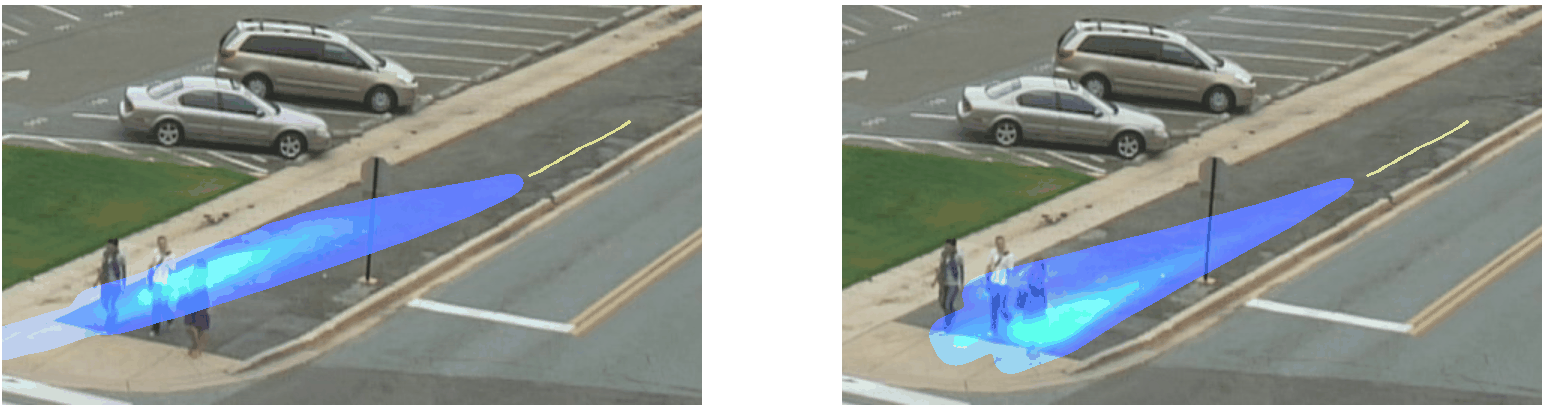}
	\caption{On the top is the user study web interface. The video frames are shown in the format of GIFs. The bottom image shows the last video frame. In this example, the women on the right is the target agent and from the two images we can see that the model of the right image is better since the predicted heatmap covers all of the woman's future path.}
	\label{fig:user_study_interface}
\end{figure}
\begin{figure}[ht]
	\centering
		\includegraphics[width=0.95\textwidth]{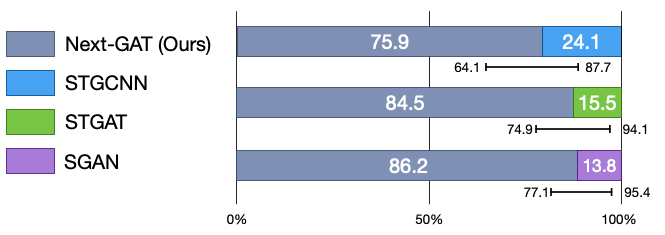}
	\caption{User study of trajectory prediction results. Both the mean and 95\% confidence interval are shown.}
	\label{fig:user_study}
\end{figure}

\begin{figure}[ht]
	\centering
		\includegraphics[width=0.95\textwidth]{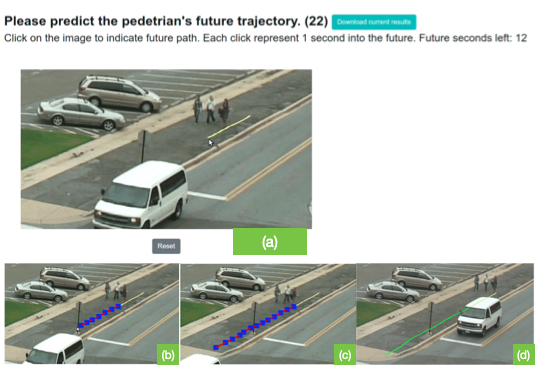}
	\caption{Human performance experiment. See text for details}
	\label{fig:human_perf}
\end{figure}

\noindent\textbf{Baseline Methods.}
We compare the proposed Next-GAT model with recent representative, open-sourced baselines as well as classic baselines, which includes:
(1) Nearest Neighbor (NN) is a simple and intuitive baseline in which we stitch the most similar trajectories observed in the training set during inference time.
(2) Constant Velocity~\cite{scholler2019simpler} (Const. Vel.) is another simple non-parametric baseline in which we use the velocity at the last observation timestep to generate future trajectories.
(3) The \textit{SGAN} \cite{gupta2018social} model consists of a generator for trajectory prediction and a discriminator which outputs the predicted trajectory with real or fake label. 
(4) The \textit{STGAT} \cite{huang2019stgat} model uses graph attention network to model agent interactions. This is a popular method that uses recurrent neural nets.
(5) The \textit{STGCNN} \cite{mohamed2020social} model substitutes the need of aggregation methods by modeling the interactions as a graph. This is a recent representative and popular method that use convolutional neural nets.
\textit{Next} model is described in \autoref{chap:0301_joint}.
\textit{Next-GAT} is our proposed method in this chapter.

\noindent\textbf{Implementation Details.}
For ActEV dataset, we transform the trajectory from pixels to the ground plane using the provided homography matrices.
For MEVA-Trajectory dataset, the global trajectory in world coordinates is used. The visual features of the longest local tracks are used as inputs to our model.
We use GRU cell for both the encoder and decoder. 
No teacher forcing is used during training of the decoder.
The embedding size $d_e$ is set to 128, and the hidden sizes $d$ of encoder and decoder are both 256. 
The scene semantic segmentation features are resized to (64, 36) and the scene convolution layers are set to have a kernel size of 3, a stride of 2 and the channel dimension is 64. 
The activation function is $tanh$ if not stated otherwise and we do not use any normalization. 
For training, we use Adam optimizer~\cite{zeiler2012adadelta} with an initial learning rate of 0.0001 and the dropout value is 0.3. 
For multi-future inference, we train 20 identical models with different random initialization.
We also tune the hyper-parameters of other baselines if the reported ones do not perform well in their papers.

\subsection{ActEV}
\label{sec:0303_actev}
\noindent\textbf{Main Results.}
The quantitative results are shown in Table~\ref{tab:0303_actev_exp}, where both short-term trajectory prediction and long-term trajectory prediction results are shown. The numbers are ADE/FDE respectively for trajectories, and mAP for activity (``Act''). 
The unit of the numbers are meters.
As reported by ~\cite{scholler2019simpler}, constant velocity can already achieve relatively good results for both short-term and long-term prediction, in terms of single-future ($minADE_1$/$minFDE_1$) metrics.
Comparing STGAT and STGCNN, we find that STGCNN is slightly better on single-future metrics, but significantly worse on multi-future metrics.
Comparing SGAN and STGAT, SGAN is slightly better for short-term trajectory prediction, but STGAT is significantly better for multi-future long-term trajectory prediction.
Our Next-GAT achieves significantly better results across all metrics and tasks.
It is able to outperform the Next model thanks to the graph attention modeling and better action representations.
Notably, Next-GAT achieves a significant 28\% improvement over the second best method that are not in this thesis (STGAT) on $minADE_{20}$ for long-term trajectory prediction task, while only 17\% improvement for short-term prediction. This validates the efficacy of our proposed method for long-term trajectory prediction.

\begin{figure}[ht]
	\centering
		\includegraphics[width=0.95\textwidth]{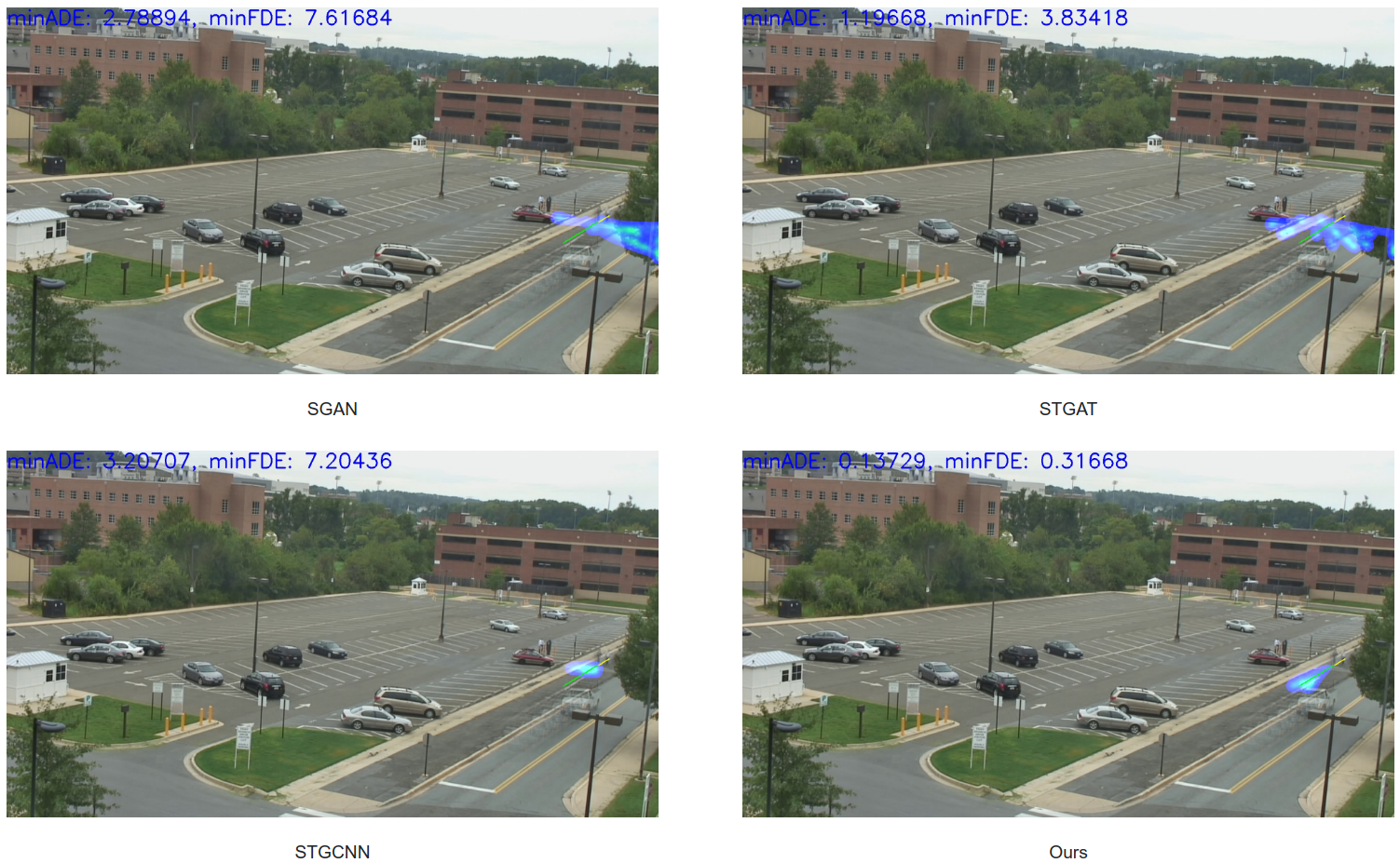}
	\caption{Qualitative analysis. The yellow trajectories are the history trajectory and the green trajectories are the ground truth future trajectories. The predicted trajectories are in green-blue heatmaps. See text for analysis.}
	\label{fig:qualitative-actev-good}
\end{figure}

\begin{figure}[ht]
	\centering
		\includegraphics[width=0.95\textwidth]{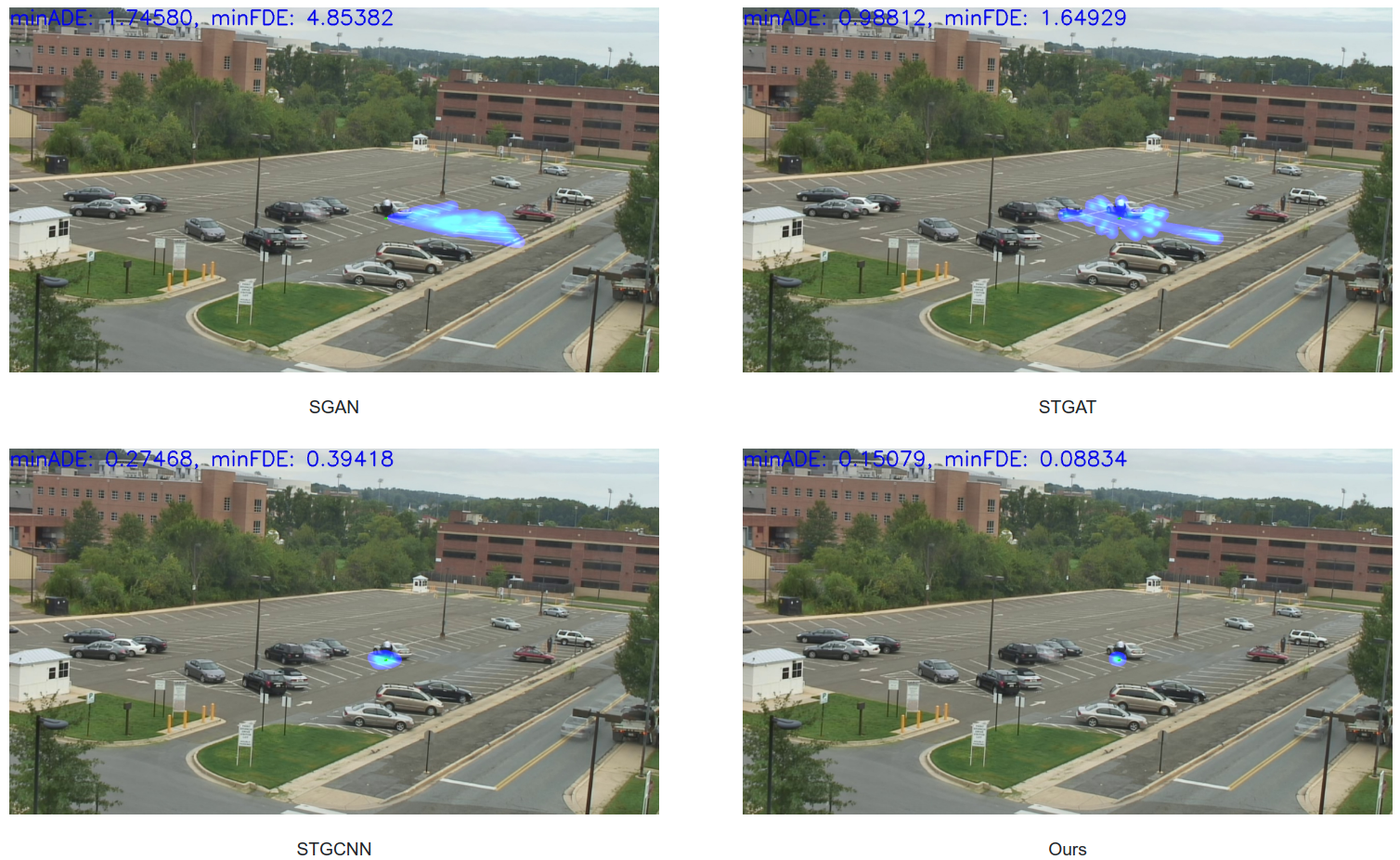}
	\caption{Qualitative analysis. The yellow trajectories are the history trajectory and the green trajectories are the ground truth future trajectories. The predicted trajectories are in green-blue heatmaps. See text for analysis.}
	\label{fig:qualitative-actev-static}
\end{figure}

\noindent\textbf{User Studies.}
We are interested to find out whether the ADE/FDE differences in Table~\ref{tab:0303_actev_exp} are conceivable by humans.
We conduct a user study to evaluate the multi-future long-term trajectory prediction performance between different baselines.
For each baseline, we show the baseline's visualization and our method side by side (the order is randomized) and ask the annotator to select the better one.
The visualization includes the yellow path, which is the observation trajectory, and the heatmaps, which are the multi-future trajectory prediction results of the corresponding method.
The video frames of the future period are shown in the format of GIFs.
Note that the ground truth future trajectory is not shown on purpose so that the human annotator would not ignore the video frames and directly compare the ground truth trajectory with the prediction results.
The annotation instructions are as follows:
Select the one with heatmap covering more of the person's path. If the same, select the one with smaller heatmap.
See Figure~\ref{fig:user_study_interface} for the user interface.
In total, results from 6 annotators with 100 pairs of samples each are recorded. 
On average, annotators have to look at each sample 3-4 times to decide. Sometimes due to the fact that two predictions are similar in human eyes, the selection could be close to a random choice between the two.
The final results are shown in Figure~\ref{fig:user_study}.
Both the mean and 95\% confidence interval are shown in the figure.
As we see, the results show that our method is significantly better than all the baselines and it is clearly perceivable by humans.
However, different from the ranking in Table~\ref{tab:0303_actev_exp}, the second best method is STGCNN rather than STGAT, suggesting a difference of 0.4 in terms of $minADE_{20}$ might not be noticeable by humans.
To the best of our knowledge, we are the first to conduct such kind of user study to look into how ADE/FDE are perceived by humans.
Note that inconceivable ADE/FDE differences may still be meaningful in many applications that rely on better trajectory prediction module.

\noindent\textbf{Human Performance.}
In this section, we conduct a human performance experiment to see how well human can predict the pedestrian future trajectories.
We randomly select 200 samples from the test set and put them into a web interface for human to annotate as shown in Figure~\ref{fig:human_perf}. Fig.~\ref{fig:human_perf} (a) shows the interface with a GIF image showing the video frames of the observation period. The annotator is asked to use the pointer to click on the image where the pedestrian is going to appear at a one-second interval. The total length they are asked to predict is 12 seconds, the same as our long-term setting.  Fig.~\ref{fig:human_perf} (b) and  Fig.~\ref{fig:human_perf} (c) show what the image looks like after the annotator has clicked multiple locations. In Fig.~\ref{fig:human_perf} (d) we show the ground truth future trajectory as a reference.
As we see, there is already about a half meter difference between the final location of the annotators' compared to the ground truth's.
Table~\ref{tab:0303_actev_exp} shows the results. As we see, the numbers are only close to the constant velocity baseline, which is expected since in our experience, it is actually quite hard for human to predict exactly where the pedestrian is going to be in a fixed future time-frame.

\begin{figure}[ht]
	\centering
		\includegraphics[width=0.95\textwidth]{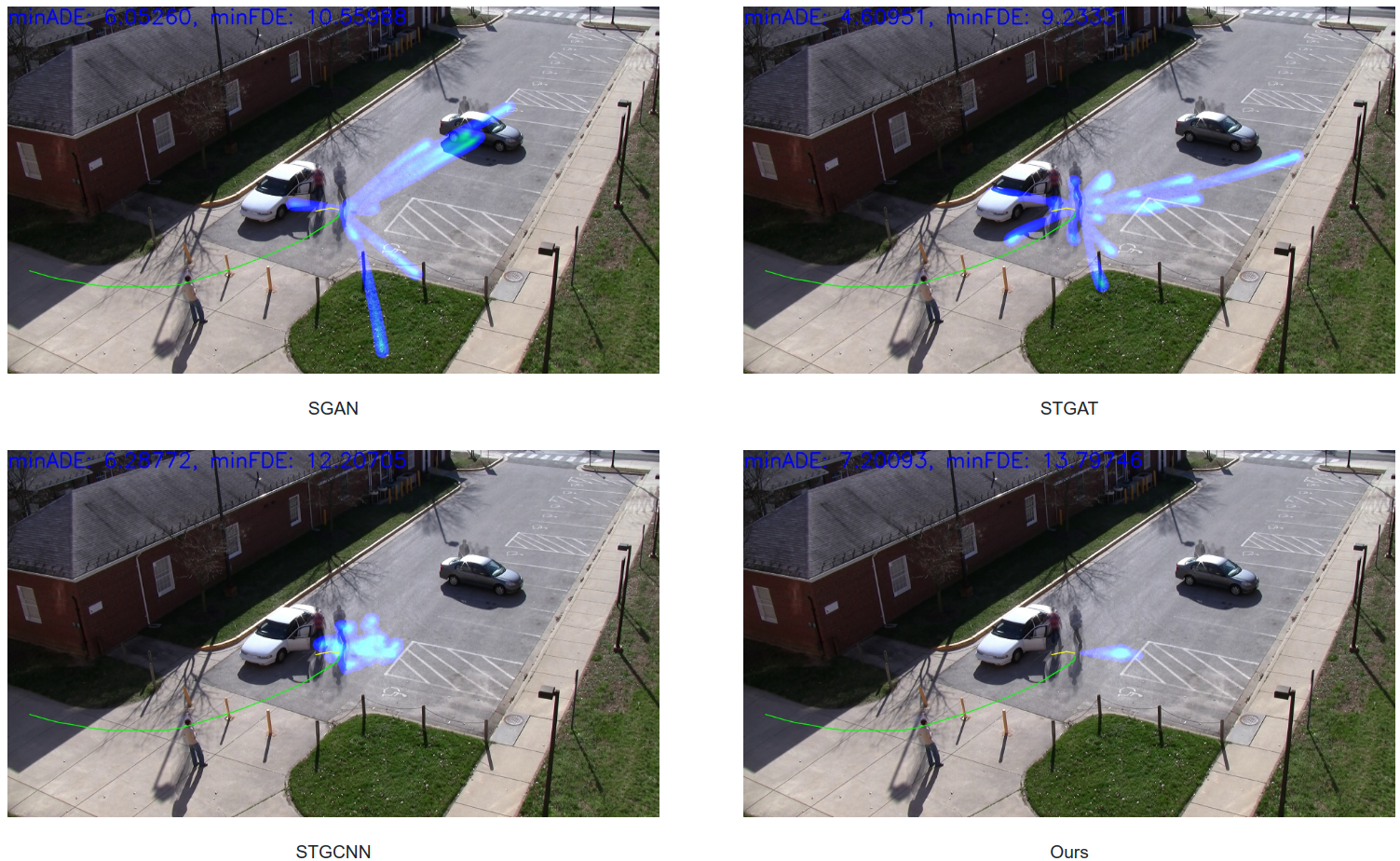}
	\caption{Error analysis. See text for details}
	\label{fig:qualitative-actev-error}
\end{figure}

\begin{figure}[ht]
	\centering
		\includegraphics[width=0.95\textwidth]{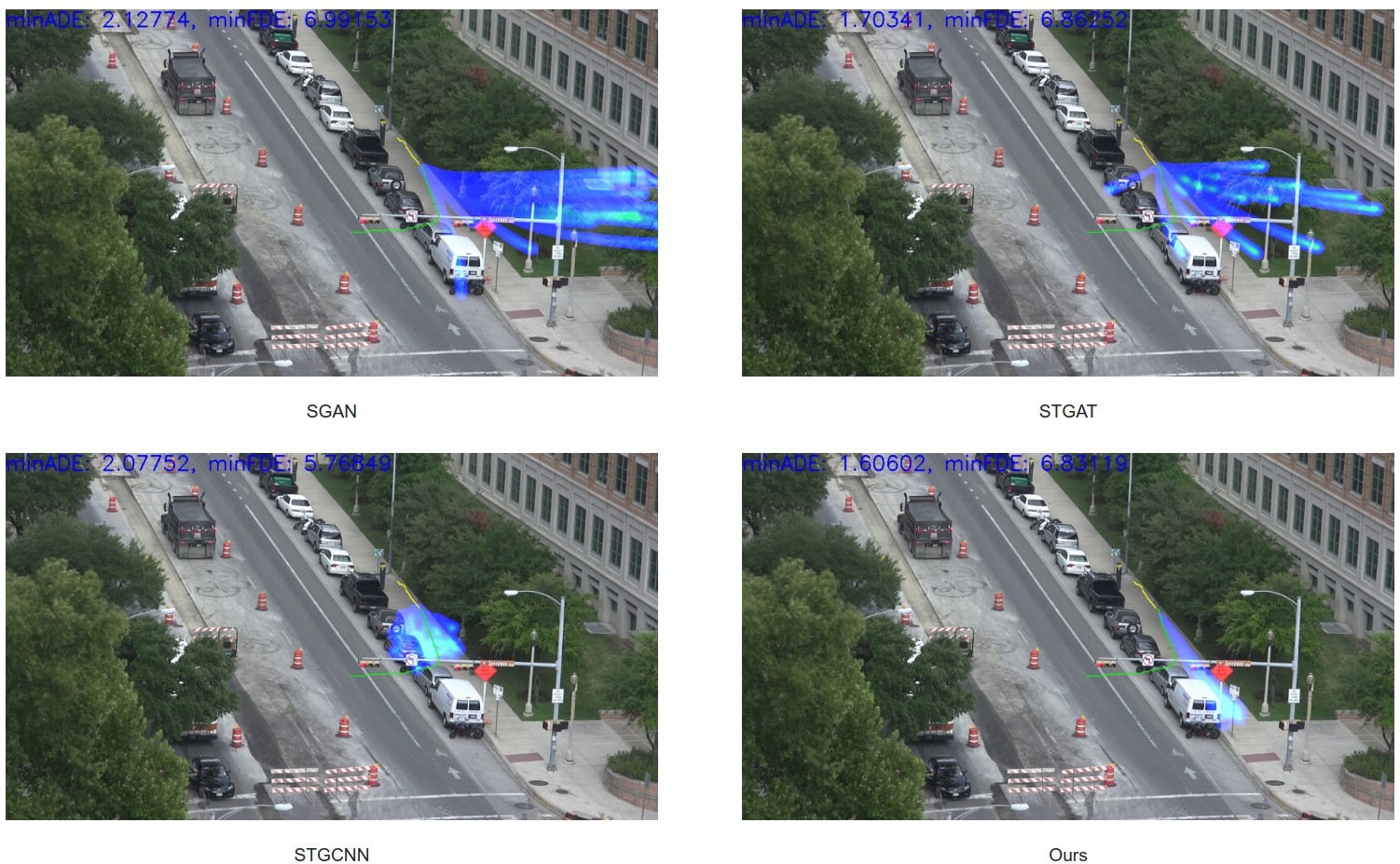}
	\caption{Error analysis. See text for details}
	\label{fig:qualitative-actev-error2}
\end{figure}

\begin{figure}[ht]
	\centering
		\includegraphics[width=0.95\textwidth]{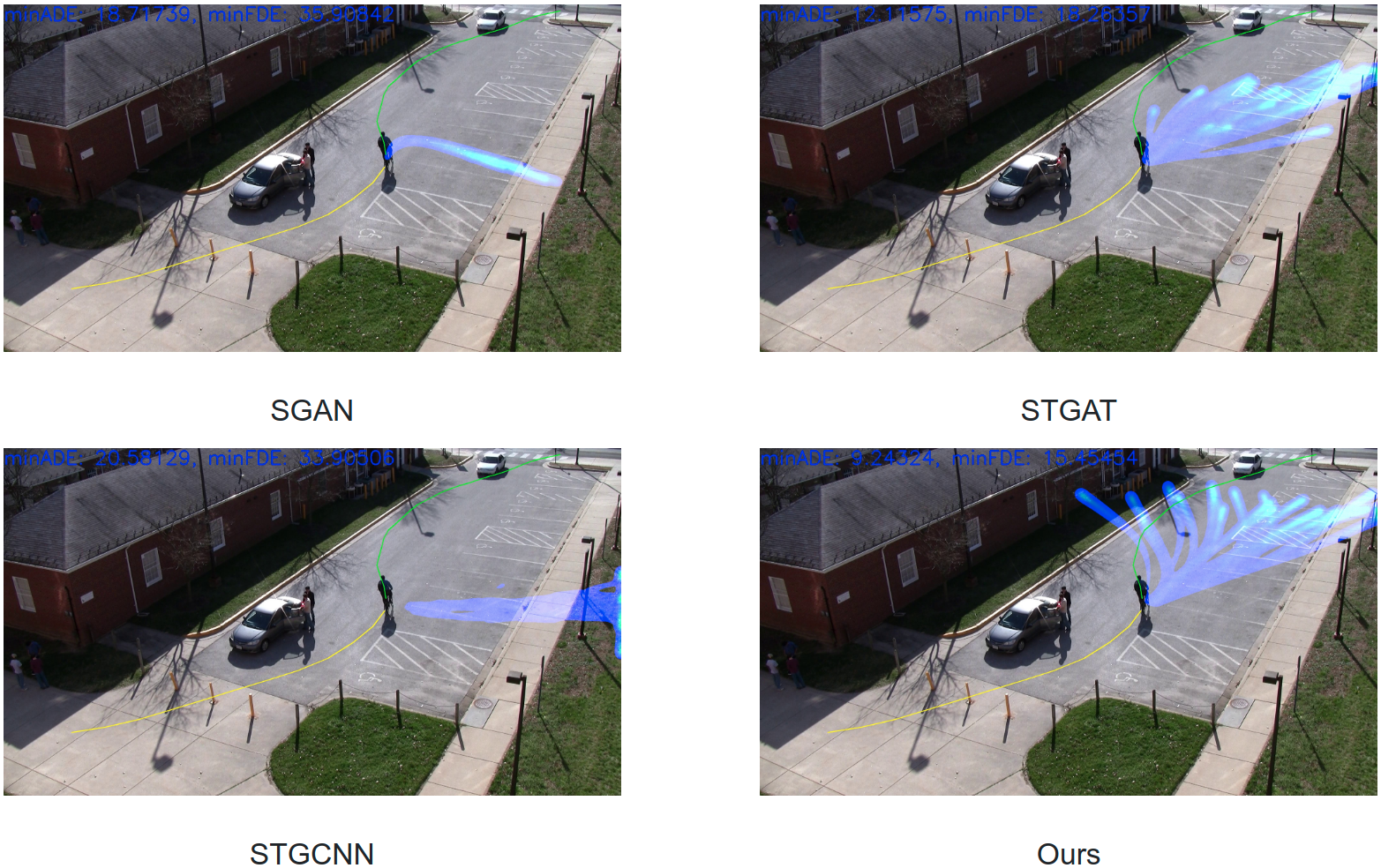}
	\caption{Error analysis. See text for details}
	\label{fig:qualitative-actev-error3}
\end{figure}

\noindent\textbf{Qualitative Analysis.}
We visualize some outputs of the 4 representative methods in Figure~\ref{fig:qualitative-actev-good} and Figure~\ref{fig:qualitative-actev-static}.
These are multi-future trajectories for long-term prediction.
In each image, the yellow trajectories are the history trajectory and the green trajectories are the ground truth future trajectories. The predicted trajectories are in green-blue heatmaps.
On the top-left corner of each image, $minADE_{20}$ and $minFDE_{20}$ of this sample are shown.
In Figure~\ref{fig:qualitative-actev-good}, an example of a pedestrian that is walking is compared.
In Figure~\ref{fig:qualitative-actev-static}, an example of a pedestrian that is static is compared.
As we see, our model correctly generally
puts probability
mass where the ground truth is, and does not ``waste''
probability mass where there is no data.
Looking at the other methods, STGAT has a more ``spred-out'' prediction pattern while STGCNN has a more ``focused'' probability mass.
SGAN is closer to STGAT but its outputs have a dominated direction.
In Figure~\ref{fig:qualitative-actev-static}, STGAT basically has a ``360'' prediction direction.

\noindent\textbf{Error Analysis.}
We analyze the most difficult test samples by ranking them based on average errors made by all four methods (three baselines and ours).
We show some typical hard cases where most methods including ours fail in Figure~\ref{fig:qualitative-actev-error}, in Figure~\ref{fig:qualitative-actev-error2} and Figure~\ref{fig:qualitative-actev-error3}.
In Figure~\ref{fig:qualitative-actev-error}, the target person slowly moves near the white car while talking to the other person that enters the car, and then during the prediction period, the person starts moving to the left side. None of the methods is able to capture that, with STGAT outputing one closest path.
This is a difficult case as there is no clear indication of the person's future direction.
After looking through the video, we think that this would only be possible for any model to predict if all previous video history can be utilized, in which it shows that the target person goes from the left side to the white car.
In Figure~\ref{fig:qualitative-actev-error2}, the target person is walking along the sidewalk but in the middle of the long-term prediction period, the target person suddenly turns and decides to go across the street.
None of the methods captures this turn.
During the observation period there is no indication that the person is trying to cross the street.
However, for a human driver seeing this pedestrian, it is common sense for them to assume that they have to be prepared for the pedestrian to cross/jay-walk.
In Figure~\ref{fig:qualitative-actev-error3}, the target person is riding a bike and a vehicle is coming towards him.
This is a hard sample since it involves person-vehicle negotiation. 
In the last observation frame, we can see that the person has decided to turn left and then later get around the incoming vehicle to exit the parking lot.
This kind of corner-cases in current dataset would need a high-level reasoning model to predict agent negotiations as well as common sense understanding (people on the sidewalk is possible to suddenly cross the road) to solve.
We leave this to future work.

\begin{table}[]
\centering
\begin{tabular}{l|c|c}
\hline
         & $minADE_{10}/minFDE_{10}$ & $minASD_{10}/minFSD_{10}$ \\\hline
SGAN     & 2.76/5.42           & 0.353/0.642         \\
STGAT    & 2.46/4.98           & 1.327/2.295         \\
STGCNN   & 2.90/5.63           & 0.761/1.120         \\
Next-GAT & 1.71/3.54           & 0.192/0.375        \\\hline
\end{tabular}
\caption{Prediction diversity measurements. See text for details}
\label{tab:0303_actev_diversity}
\end{table}

\noindent\textbf{Prediction Diversity.}
In self-driving applications, we may consider the diversity of the prediction models so that they could cover all possible future trajectories to ensure safety.
Previous metrics like $minADE_{k}$ do not measure diversity.
In \autoref{chap:0201_multi}, with the Forking Path dataset (with multi-modal ground truth trajectories) and a explicit probabilistic model, we could use the Negative-Log-Likelihood (NLL) metric to measure both diversity, precision and recall.
However in real-world video dataset like ActEV, only one ground truth trajectory is available. 
And in this chapter we are using generative model, which does not output probability for each trajectory.
So to measure prediction diversity, we follow the work of Yuan et. al.~\cite{yuan2019diverse}. 
They proposed to use minimum Average Self Distance (minASD) and minimum Final Self Distance (minFSD), which measures average L2 distance over all time steps or the final ones between a predicted sample and its closest neighbor prediction.
In this way, repeated samples will be penalized.
We hope that this metric can help users better understand the characteristics of each method.
We show the results in Table~\ref{tab:0303_actev_diversity}.
Following \cite{yuan2019diverse} we use K=10 per test sample.
Unsuprisingly, Next-GAT model has the lowest self distances as our multi-future prediction models consist of the same models trained using different random initialization.
We focus on the performance of single-future trajectory prediction for this model.
We can also see that STGAT has the best diversity according to the metric, which is consistent with what we see in the qualitative examples in Fig.~\ref{fig:qualitative-actev-good}.
In real-world deployment, such property (diversity) should be taken into account.

\subsection{MEVA-Trajectory}

\begin{table}[]
\centering
\begin{tabular}{l||c|c|c|c|c|c}
\hline
            & \multicolumn{3}{c|}{Short-term Trajectory Prediction} & \multicolumn{3}{c}{long-term Trajectory Prediction} \\\hline
            & Act      & Single-Future     & Multi-Future     & Act  & Single-Future     & Multi-Future     \\\hline\hline
NN          & -             & 7.32/13.54        & -                & -            & 15.29/30.00       & -                \\
Const. Vel. & -             & 2.76/5.76         & -                & -            & 8.35/17.89        & -                \\\hline
SGAN        & -             & 3.41/7.21         & 1.92/4.04        & -            & 8.77/18.11        & 7.24/14.98       \\
STGAT       & -             & 5.05/10.43        & 2.00/4.15        & -            & 14.75/29.51       & 7.71/15.68       \\
STGCNN      & -             & 4.79/8.56         & 3.36/6.33        & -            & 14.60/27.42       & 11.54/22.63      \\\hline
Next   &  0.257 & 2.14/5.04     & 1.95/4.55    & 0.176 & 7.62/18.20    & 6.98/16.60   \\
Next-GAT    & \textbf{0.328} & \textbf{1.91/4.33}     & \textbf{1.63/3.75}    & \textbf{0.299} & \textbf{6.51/14.67}    & \textbf{5.60/12.82}          \\\hline
\end{tabular}
\caption{Main results on the MEVA-Trajectory dataset. The unit is in feet. See text for details.}
\label{tab:0303_meva_exp}
\end{table}

\noindent\textbf{Main Results.}
We show the main results in Table~\ref{tab:0303_meva_exp}, where both short-term trajectory prediction and long-term trajectory prediction results are shown. The numbers are ADE/FDE respectively for trajectories, and mAP for activity (``Act''). 
The unit of the numbers are feet (~0.3 meters). See Figure~\ref{fig:meva_ruler}. 
From the results, we can see that our method is also able to achieve the best among all metrics.
Looking at the numbers of recent methods, we can see that STGCNN perform badly on this dataset in terms of multi-future long-term trajectory prediction, showing that this dataset is more difficult compared to ActEV.
STGAT also performs badly on single-future long-term trajectory prediction, but its multi-modality makes up for multi-future prediction.
Comparing our Next-GAT with Next, Next-GAT improves signifcantly on activity prediction and in turn it helps trajectory prediction as well.

\begin{figure}[ht]
	\centering
		\includegraphics[width=0.95\textwidth]{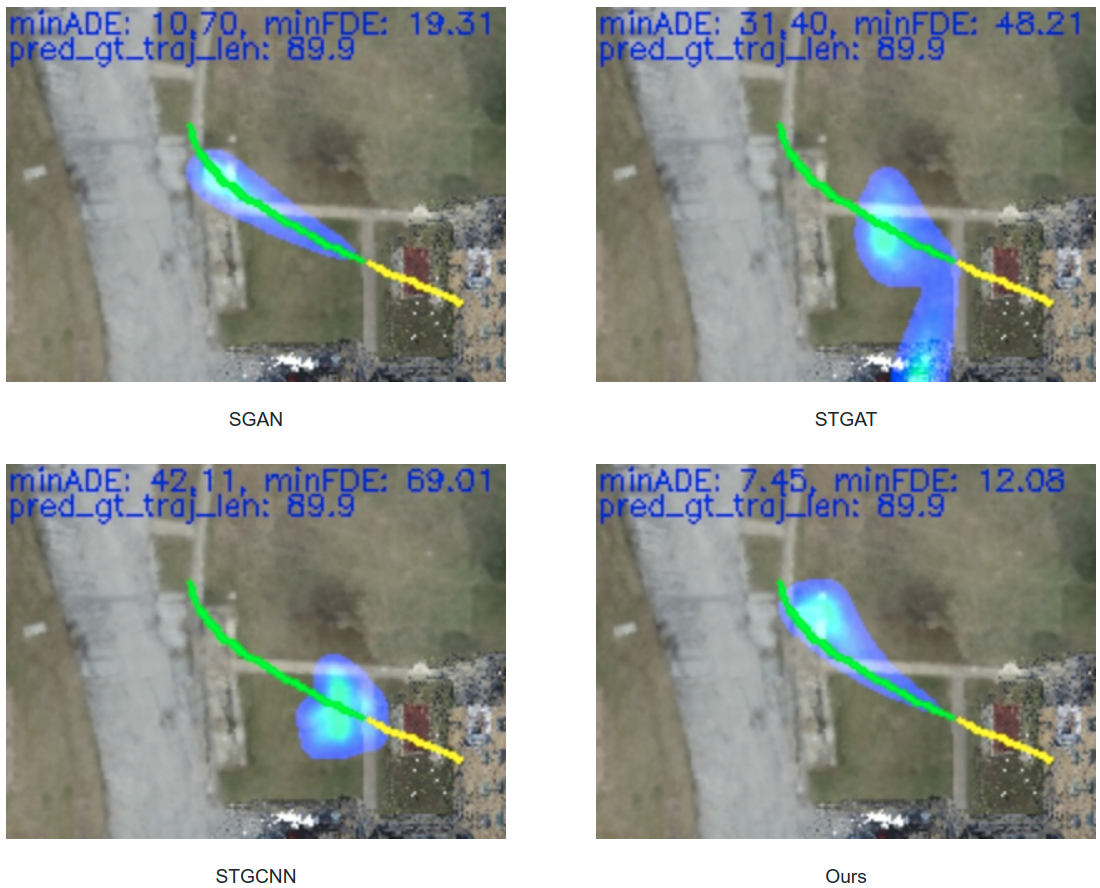}
		\includegraphics[width=0.95\textwidth]{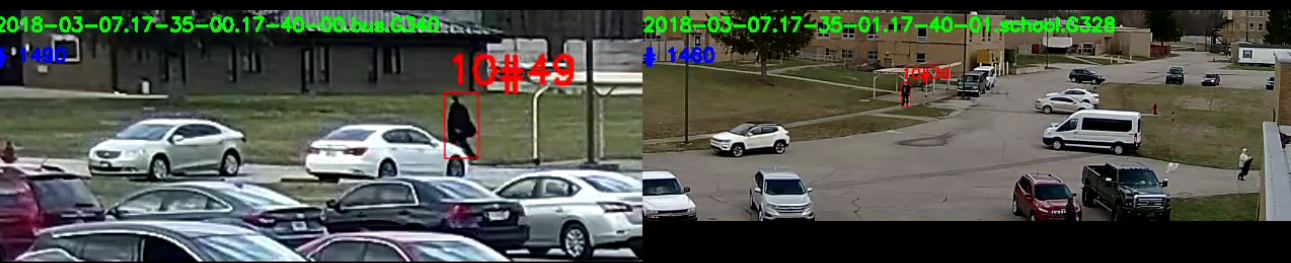}
	\caption{Qualitative analysis. The top four images show visualization of the trajectories from four methods and below shows the video frame example of the global track. See text for details}
	\label{fig:qualitative-meva-good1}
\end{figure}

\begin{figure}[ht]
	\centering
		\includegraphics[width=0.95\textwidth]{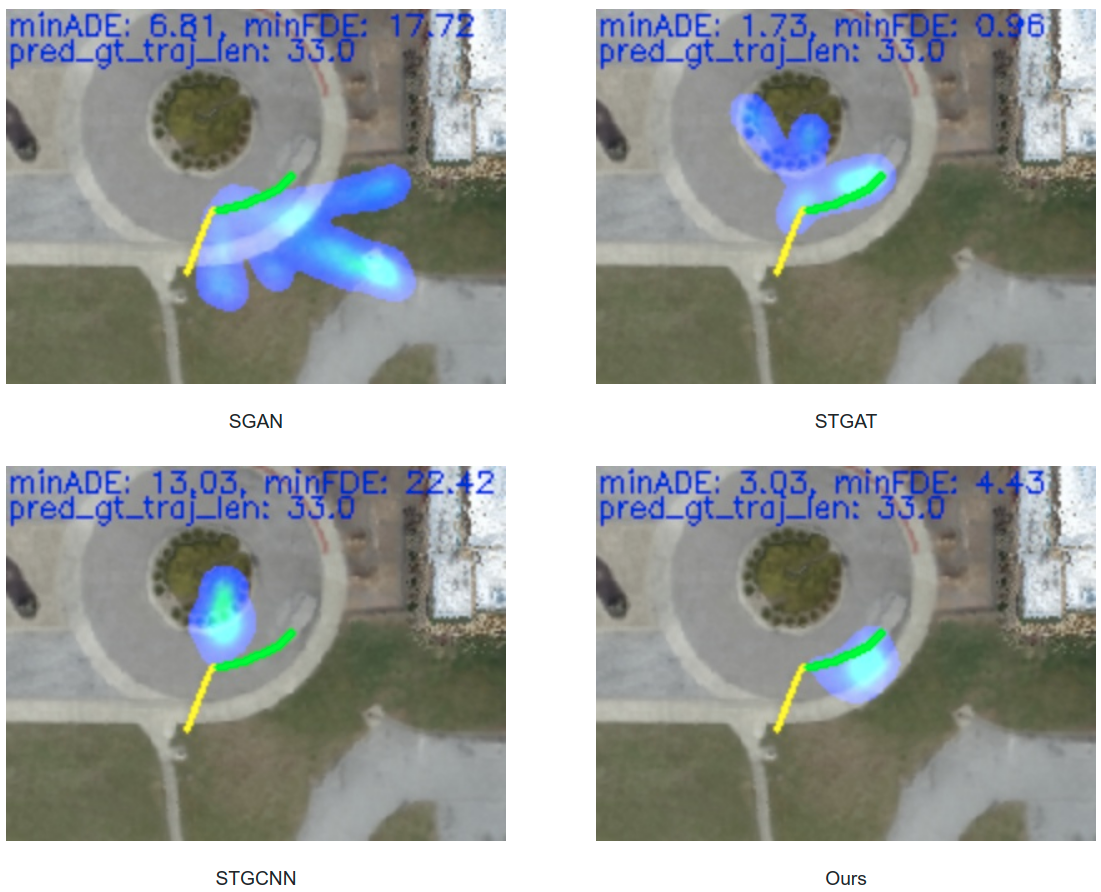}
		\includegraphics[width=0.95\textwidth]{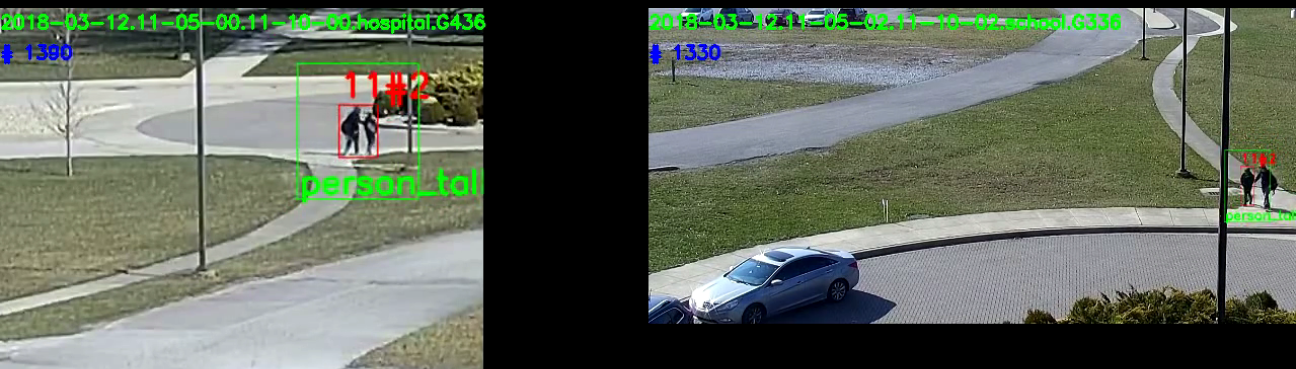}
	\caption{Qualitative analysis. The top four images show visualization of the trajectories from four methods and below shows the video frame example of the global track. See text for details}
	\label{fig:qualitative-meva-good2}
\end{figure}

\noindent\textbf{Qualitative Analysis.}
We visualize the results and show them in Figure~\ref{fig:qualitative-meva-good1} and Figure~\ref{fig:qualitative-meva-good2}.
In each figure,
the top four images show visualization of the trajectories from SGAN, STGAT, STGCNN and ours, respectively, and below shows the video frame example of this person.
We also show the $minADE_{20}$ and $minFDE_{20}$, as well as the ground truth trajectory's length at the top-left corner of each visualization image.
In Figure~\ref{fig:qualitative-meva-good1}, we can see that Next-GAT outputs the most accurate path prediction.
In Figure~\ref{fig:qualitative-meva-good2}, both Next-GAT and STGAT correctly predict that the person is going to turn.

\noindent\textbf{Ablation Experiments.}
We test various ablations of our Next-GAT model on single-future trajectory prediction to substantiate our design decisions.
Results are presented in Table~\ref{tab:meva_ablation}, where activity prediction mAP and minADE\textsubscript{1}/minFDE\textsubscript{1} are shown.
We verify three of our key designs by leaving the module out from the full model.
(1) \textit{graph attention:} Our model is built on top of graph attention to model the agents' interaction. We compare Next-GAT with Next, which does not use graph attention.
As we see, the performance drops on trajectory prediction.
(2) \textit{STAN-Action module:} We test our model without the STAN-Action module and replace it with a single ResNet group.
As we see, the significant performance drops on activity prediction verify the efficacy of this new module proposed in our study. 
(3) \textit{Scene modeling:} We train Next-GAT without the scene understanding module.
We see that performance is slightly worse if we do not have static scene modeling.
Finally, we compare our model with a simple GRU-based Encoder-Decoder model with only trajectory inputs. We can see that motion information alone cannot bring good performance.

\begin{table}[]
\centering
\begin{tabular}{l|c|c|c}
\hline
                 & \multicolumn{3}{c}{long-term Trajectory Prediction}  \\\hline
                 & Activity & $minADE_1$ & $minFDE_1$                     \\\hline
Next-GAT         & 0.299    & 6.51      & 14.67 \\\hline\hline
Next             & 0.176    & 7.62      & 18.2                          \\\hline
Next-GAT-ResNet  & 0.253    & 7.02      & 15.55                         \\\hline
Next-GAT-noScene & 0.280     & 6.88      & 15.78                         \\\hline
GRU-EncodeDecode & -        & 9.69      & 20.97      \\\hline                   
\end{tabular}
\caption{Ablation experiments on MEVA-Trajectory dataset. See text for details.}
\label{tab:meva_ablation}
\end{table}

\section{Summary}
In this chapter, we present our Next-GAT model for long-term trajectory prediction in urban traffic scenes.
Next-GAT utilizes graph attention with scene and action understanding module, which is crucial for long-term prediction.
We show our method's efficacy on two major benchmarks, ActEV and our MEVA-Trajectory dataset.

\part{Conclusions and Future Directions}
\label{part:conclusion}
\chapter{Conclusions} \label{chap:conclusion}

This thesis explores the problem of pedestrian future trajectory prediction.
Our goal is to promote human safety in applications such as socially-aware robotics or autonomous driving. 
Specifically, throughout this thesis, we develop novel models and new datasets for three closely related tasks:
\textit{human action analysis, pedestrian trajectory prediction with scene semantics, and joint analysis of human actions and trajectory prediction}. 
In this thesis, we have demonstrated that, with scene understanding and enhanced behavioral representations, one can achieve accurate state-of-the-art performance in the challenging long-term trajectory prediction task in urban traffic scenes. 
Below, we summarize the contributions of each part of our work (please refer to \autoref{sec:overall_impact} for our overall impact). 
We then provide our key insights based on all the works we have done.
Finally, we provide our recommendations for real-world applications  using our models.

\section{Contributions}
\subsection{\autoref{part:action_analysis}: Human Action Analysis}
\begin{itemize}
    \item We are one of the early works that study how we could better utilize weakly-supervised video data from the Internet. Our algorithm won several TRECVID challenges.
    \item We explore viewpoint-invariant representation for action recognition and detection, which is also one of the early works to tackle this problem in video.
\end{itemize}

\subsection{\autoref{part:future_prediction}: Pedestrian Trajectory Prediction with Scene Semantics}
\begin{itemize}
    \item We develop a new model for multi-modal trajectory prediction and propose the first human-annotated benchmark for multi-future trajectory prediction.
    \item We propose a novel algorithm to build robust trajectory prediction models that could be transfer to different domains, which is an important under-explored problem in this research field.
\end{itemize}

\subsection{\autoref{part:joint}: Joint Analysis of Human Actions and Trajectory Prediction}
\begin{itemize}
    \item We propose a new human-annotated long-term trajectory prediction benchmark with multi-view video data that includes person-vehicle interactions and rich human activities in urban traffic scenes.
    \item We develop the first joint action and trajectory prediction model utilizes both scene understanding and viewpoint-invariant action representation for the challenging long-term trajectory prediction problem.
\end{itemize}

\section{Key Insights}
\subsection{Limitations of datasets}
\begin{figure}[ht]
	\centering
		\includegraphics[width=0.95\textwidth]{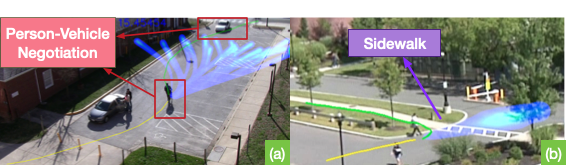}
	\caption{Important corner cases examples. See text for details.}
	\label{fig:key_take_away}
\end{figure}

Based on our experience working on datasets like ETH/UCY, ActEV, MEVA-Trajectory (\autoref{chap:0302_longterm_data}), we have noticed that these datasets based on outdoor scenes may not include enough critical corner cases (See Fig.~\ref{fig:qualitative-actev-error2}, Fig.~\ref{fig:qualitative-actev-error3}, Fig.~\ref{fig:key_take_away}) that are important for safety.
This could be supported by the fact that in many experiments (\autoref{chap:0301_joint}, \autoref{chap:0303_longterm_model}), simple constant velocity method can already achieve good results.
We also observe that after removing person pose features (from the Next model in \autoref{chap:0301_joint} to the Next-GAT model in \autoref{chap:0303_longterm_model}), the performance is not affected, which is counter-intuitive as drivers when driving in parking lot they do utilize pedestrians' poses to read their intentions, etc.
The main reason for this might be due to the fact that the datasets we use only has a very small portion of the trajectories that require the models to understand multi-agent negotiations.
For example, in Fig.~\ref{fig:key_take_away} (a), we can see that in this scenario, the target person is riding a bike and a car is coming to the person's direction.
The ground truth future trajectory suggests that the person has eventually moved to the left to get around the car's expected future path.
Such person-vehicle negotiation might be intuitive for humans but it is hard for models to learn, especially when this kind of data is rare.
See Fig.~\ref{fig:qualitative-actev-error3} in \autoref{chap:0303_longterm_model} to see how other recent baselines perform in this case.
One way to mitigate this is to find out these corner cases and use them for both training and evaluation. We have explored this by looking at samples that all baselines fail (see error analysis in \autoref{sec:0303_actev}).
However, more work needs to be done and datasets with rich multi-agent negotiations need to be collected to solve these corner/long-tail situations in traffic scenes. The PIE dataset~\cite{rasouli2019pie} is a step towards this direction but the number of scenes and viewpoints are limited.

\subsection{Limitations of training schemes}
In all of our model training, we use only the positive examples to train.
Unreasonable trajectory predictions, e.g. predicting person walking through vehicles, are not specially penalized.
We could add hand-crafted heuristic based on the semantic segmentation classes like agent cannot walk through vehicles and water.
For a more long-term vision, physical constraint in 3D simulator could be built into the training process for more general cases.
On the other hand, we have only tried training with real-world video samples, where only one ground truth trajectory for each scenario is provided. We have not tried training with the Forking Paths dataset, where multiple positive samples are provided at a time.
Using this dataset for training could allow the model to generalize better on possible predictions.

\subsection{Limitations of models using scene semantics}
During our examination of hard cases (see error analysis in \autoref{sec:0303_actev}), we have also noticed error cases where the model does not utilize the scene semantics as expected to predict the correct trajectory.
As we see in Fig.~\ref{fig:key_take_away} (b), the person walks straight towards a crosswalk but he is actually going to turn left onto the sidewalk.
Our model (and other baselines as well) predicts the trajectory solely based on the observation trajectory and does not use sidewalk semantic information.
Although we have kept the spatial dimension of scene semantic segmentation features when inputting into our model (method like SoPhie~\cite{sadeghian2018sophie} uses segmentation model's fully connected layer output, which could be even harder to learn), the model seems to not fully capture the correlations.
We think that it may be worthwhile to revisit simple Hidden Markov Model as proposed in ~\cite{kitani2012activity} to better associate directly scene classes and trajectories.

\subsection{Evaluation metrics}
Throughout this thesis, we have utilized different metrics to evaluate models capabilities for single-future prediction and multi-future prediction (\autoref{sec:0303_exp}). 
To ensure safety, diversity of multi-modal prediction model should also be measure.
Metrics like $minADE_{k}$ do not measure diversity.
In \autoref{chap:0201_multi}, with the Forking Path dataset (with multi-modal ground truth trajectories) and a explicit probabilistic model, we could use the Negative-Log-Likelihood (NLL) metric to measure both diversity, precision and recall.
However in real-world video dataset like ActEV, only one ground truth trajectory is available. 
And many generative models, which do not output probability for each trajectory, cannot be evaluated using NLL metric.
So following the work of Yuan et. al.~\cite{yuan2019diverse} to measure prediction diversity, 
we use minimum Average Self Distance (minASD) and minimum Final Self Distance (minFSD), which measures average L2 distance over all time steps or the final ones between a predicted sample and its closest neighbor prediction, in \autoref{sec:0303_actev}.
In this way, repeated samples will be penalized.
In conclusion, we believe that NLL is a better metric to use when possible, and minADE and minASD should both be considered in other situations to have a more comprehensive understanding of the model characteristics, which could be crucial during real-world deployment.

\section{Recommendations for Real-World Applications}
In this section we briefly discuss recommendations for applying our models to an unseen dataset and the possibilities of achieving real-time prediction during test time.

\noindent\textbf{Training and testing on an unseen dataset.} 
We recommend using our SimAug-trained (\autoref{chap:0202_3d}) model when trying to predict pedestrian trajectories on an unseen dataset.
Our SimAug algorithm is specially designed to optimize the domain transfer ability and our experiments show that it works well on unseen cameras like top-down view and ego-centric view.
When extra training data could be available, we recommend using at many annotations as the ActEV dataset with 12 hours of trajectories, with which models can achieve within one meter average errors for 5-second prediction and within two meters for 12-second prediction (see \autoref{sec:0303_actev}). 
The training time is estimated to be 8 hours with a middle-to-high-end GPU like GTX 1080 TI.
When human annotations are not readily available, we recommend follow our instructions when building the MEVA-Trajectory dataset (\autoref{chap:0302_longterm_data}), in which we have minimized manual effort with state-of-the-art tracking and re-identification systems. 

\noindent\textbf{Real-Time Processing.} 
As shown in \autoref{chap:0101_object}, our object detection and tracking system can achieves real-time speed if we process about 5 video frames per second (Table~\ref{tab:0101_speed_run}) given a relatively new 4-GPU machine listed in Table~\ref{tab:0101_speed_machine}.
Recall that in this work we only need 2.5 frame-per-second observation features to achieve the good results we have shown thoughout this thesis.
The prediction inference time is about 10x faster than real-time, which means that with our efficient object detection and tracking system, one can easily build a real-time end-to-end system with our prediction models. 
One of our open-source project~\footnote{\url{https://github.com/JunweiLiang/social-distancing-prediction}} already provides an example code base using the Next model (\autoref{chap:0301_joint}).


\chapter{Vision and Future Directions} \label{chap:future}
In this thesis, we have propose novel deep learning models that utilize enhanced contextual cues like scene semantics and human actions for trajectory prediction in urban traffic scenes.
Our research goal is to promote human safety in applications such as robotics or autonomous driving. 
We have explored multi-modal trajectory prediction and robust learning algorithms using simulation-augmentation. 
These are important directions as they provides better traffic safety if used in diverse environment.
To push the boundary even further, we have proposed a long-term trajectory prediction dataset and models for joint action and trajectory prediction in urban traffic scenes. 
In summary, we have answered the key research question of how to build a robust trajectory prediction system with enhanced semantic context understanding for urban traffic scenes.
Below, we outline several important future research directions that we have not studied, including some short-term directions related to applications of trajectory prediction and long-term directions that require novel model and algorithm to solve.

\vspace{4mm}
\noindent{\LARGE \textbf{Short-Term Future Directions - Applications}}
\vspace{4mm}

\noindent\textbf{First-person View.} 
We have not explored trajectory prediction in first-person view videos. 
This may be difficult and different from our problem setting as first-person view videos may often include ego-motions that the models have to take into account when extracting behavioral representations for prediction. 
In future applications like social-aware robots that could navigate among humans, first-person-view trajectory prediction is crucial.

\noindent\textbf{Long-tail Cases.} We have not specifically tackled the challenging long-tail cases exist in traffic scenes ~\cite{makansi2021exposing}. 
Rare events like traffic accidents and even terrorist attack may significantly alter the behaviors of pedestrians, causing current models to fail in such scenarios. 
This is difficult to solve as data collection may not be possible. 
It is important for prediction models to handle such rare cases to ensure traffic safety at all times.

\noindent\textbf{Computation-accuracy Trade-off.} We have not systematically studied how to balance computation cost and accuracy with our models. 
It is difficult because there are a lot of alternative models with different level of performance for object detection and tracking, scene semantic segmentation and action feature representations.
We believe this is important as in applications like self-driving cars and social robots, on-board computing resources are limited. 
A clear computation-accuracy trade-off would be beneficial in real-world model deployment.

\noindent\textbf{Trajectory Prediction in Sports.}
We have not looked into trajectory prediction in sports. For example, trajectory prediction in basketball or soccer can be used to help coaches with their decision makings.
Trajectory prediction in sports require specific in-domain prior knowledge when building an effective model. It would be interesting to explore whether complex prediction algorithms can be generalized to different sports.

\noindent\textbf{Crowd Dynamics Estimation for Public Safety Monitoring.}
Trajectory prediction can enable analysis and control of crowd flow in populated areas such as malls and airports by combining other crowd dynamics analysis tools like crowd counting~\cite{cheng2019learning} (see this news story~\footnote{\url{https://www.washingtonpost.com/investigations/interactive/2021/dc-police-records-capitol-riot/}} from the Washington Post how our crowd dynamics analysis has been used in real-world scenario).
With crowd dynamics prediction and estimation, public safety events like stampede and riot could be alerted to the authorities and actions could be taken early to save lives. 

\vspace{4mm}
\noindent{ \LARGE \textbf{Long-Term Future Directions - Model \& Algorithm}}
\vspace{4mm}

\noindent\textbf{Unifying Long-term and Short-term Trajectory Prediction.}
In \autoref{chap:0303_longterm_model} we study long-term trajectory prediction with a fixed-length horizon in the experimental setup. With the grow of datasets and better algorithm, the length of the horizon is likely to expand. Meanwhile some long-term cues could be useful for short-term prediction.
We believe that a multi-task learning framework of long-term and short-term trajectory prediction could benefit from different kinds of behavioral cues.
Models with flexible prediction horizon can also be beneficial to real-world applications as well.

\noindent\textbf{Behaviors of Different Population.} We have not looked into the difference of behaviors of different population. 
For example, traffic scenes in India may be distinctively different from the U.S, making models trained from one place difficult to transfer to another.
This would be important because as self-driving cars become more and more common, on-board safety systems like trajectory prediction should be able to maintain the same safety standard around the world.

\noindent\textbf{Unifying Vehicle and Pedestrian Trajectory Prediction.} 
Currently in most self-driving systems pedestrian trajectory and vehicle trajectory prediction are done separately with different models.
It is difficult to use one trajectory prediction for another task as vehicle and pedestrians behave differently.
For example, vehicles have lane constrains while pedestrians do not.
A unify model that takes into account these differences may be beneficial for system efficiency and better prediction accuracy as these two types of traffic actors have frequent interactions in real-world situation.

\begin{figure}[ht]
	\centering
		\includegraphics[width=0.95\textwidth]{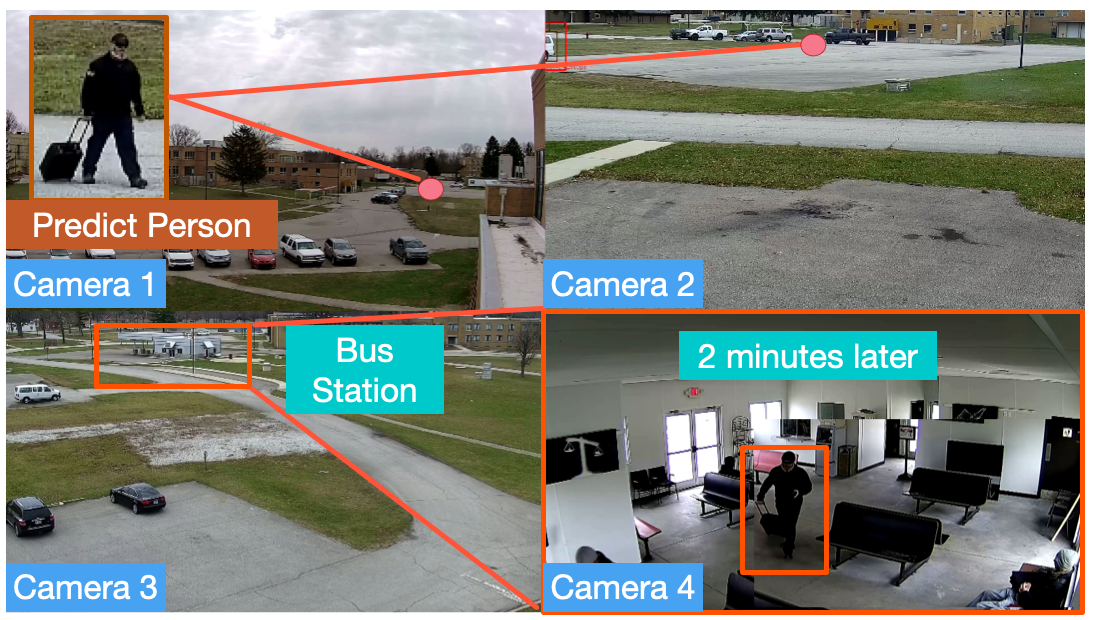}
	\caption{Example of future trajectory prediction using high-level reasoning. In two of the cameras we can see a person with a luggage, and knowing there is a bus station near the scene, the model should be equipped with a prior predicting that the person is going to appear in that location.}
	\label{fig:future_work}
\end{figure}

\noindent\textbf{Common Sense Reasoning for Long-term Future Prediction.} 
We have not explored common sense reasoning with high-level scene attributes for trajectory prediction. For example, if the scene is a school zone, we can expect the vehicles to be slow and students are present. If the scene is near a stadium and there is a game going on, we can expect people with sports jersey to be generally heading the stadium's direction. This is important information to better improve trajectory prediction as human drivers can comprehend such situation effortlessly. Ultimately, we believe it is possible to predict pedestrian trajectories minutes into the future with high-level reasoning, as shown in the Example of Figure~\ref{fig:future_work}.

\backmatter


\renewcommand{\bibsection}{\chapter{\bibname}}

\bibliographystyle{my_plainnat}
\bibliography{ref} 
\end{document}